\documentclass[3p,times]{elsarticle}
\usepackage{lineno}
\modulolinenumbers[5]
\journal{Journal of \LaTeX\ Templates}
\usepackage[utf8]{inputenc}

\usepackage{booktabs} % For formal tables
\usepackage{subcaption}
\usepackage{here}
\usepackage{wrapfig}

\usepackage{latexsym}
\usepackage{amsmath}
\usepackage{amssymb}
\usepackage{mathtools}       % many extension of amsmath
\usepackage{stmaryrd}        % \llbracket, \rrbracket
\usepackage{bm}              % bold font
\usepackage{bmpsize}
\usepackage{enumerate}

\usepackage{algorithm}
\usepackage{algpseudocode}

\usepackage{xcolor}
\usepackage{ifthen}
\usepackage{amsthm}
\usepackage[colorlinks=false,pdfborder={0 0 0}]{hyperref}
\usepackage{cleveref}

\newtheorem{theorem}{Theorem}[section]

\theoremstyle{definition}
\newtheorem{definition}[theorem]{Definition}
\newtheorem{remark}[theorem]{Remark}

\providecommand{\argmax}{\operatornamewithlimits{argmax}} % argmax
\DeclareMathOperator{\Cond}{Cond} % condition number
\providecommand{\R}{\mathbb{R}} % real field
\providecommand{\E}{\mathbb{E}} % expectation
\providecommand{\T}{\mathrm{T}} % transpose
\providecommand{\ind}[1]{\mathbb{I}\{#1\}} % indicator function
\renewcommand{\geq}{\geqslant} % geq
\renewcommand{\leq}{\leqslant} % leq
\DeclarePairedDelimiterX{\inner}[2]{\langle}{\rangle}{#1, #2}
\DeclarePairedDelimiter{\norm}{\lVert}{\rVert}
\DeclarePairedDelimiter{\abs}{\lvert}{\rvert}

\newcommand{\CC}{\mathbf{C}}

\newcommand{\X}{\mathbb{X}}

\usepackage{xcolor}
\usepackage {diagbox}

\providecommand{\X}{\mathbb{X}}
\providecommand{\pbilcma}{\texttt{AS3-CMA-ES}}

\begin{document}

\begin{frontmatter}

\title{Adaptive Scenario Subset Selection for Worst-Case Optimization \\ and its Application to Well Placement Optimization}

\author[add1,add2]{Atsuhiro Miyagi}
\ead{mygath00@pub.taisei.co.jp}

\author[add2,add3]{Kazuto Fukuchi}
\ead{fukuchi@cs.tsukuba.ac.jp}

\author[add2,add3]{Jun Sakuma}
\ead{jun@cs.tsukuba.ac.jp}

\author[add2,add3]{Youhei Akimoto}
\ead{akimoto@cs.tsukuba.ac.jp}

\address[add1]{Taisei Corporation, Infrastructure Technology Research Department, Geotechnical Research Section, Japan}
\address[add2]{University of Tsukuba, Faculty of Engineering, Information and Systems, Japan}
\address[add3]{RIKEN AIP, Japan}

\begin{abstract}
In this study, we consider simulation-based worst-case optimization problems with continuous design variables and a finite scenario set.
To reduce the number of simulations required and increase the number of restarts for better local optimum solutions, we propose a new approach referred to as \emph{ adaptive scenario subset selection } (\emph{AS3}). 
The proposed approach subsamples a scenario subset as a support to construct the worst-case function in a given neighborhood, and we introduce such a scenario subset. 
Moreover, we develop a new optimization algorithm by combining \emph{AS3} and the covariance matrix adaptation evolution strategy (CMA-ES), denoted \pbilcma{}. At each algorithmic iteration, a subset of support scenarios is selected, and CMA-ES attempts to optimize the worst-case objective computed only through a subset of the scenarios. The proposed algorithm reduces the number of simulations required by executing simulations on only a scenario subset, rather than on all scenarios.  
In numerical experiments, we verified that \pbilcma{} is more efficient in terms of the number of simulations than the brute-force approach and a surrogate-assisted approach \texttt{lq-CMA-ES} when the ratio of the number of support scenarios to the total number of scenarios is relatively small. 
In addition, the usefulness of \pbilcma{} was evaluated for well placement optimization for carbon dioxide capture and storage (CCS). In comparison with the brute-force approach and \texttt{lq-CMA-ES}, \pbilcma{} was able to find better solutions because of more frequent restarts. 
\end{abstract}


\end{frontmatter}

%\linenumbers

\section{Introduction}

% \paragraph{Background}
% \textcolor{blue}{
\subsection{Background}
% }

Simulation-based optimization is becoming popular in various industrial fields as computational performance increases. Such optimization evaluates the objective function value of a solution candidate by a computationally expensive numerical simulation. Evolutionary approaches have been successfully applied to simulation-based optimization in a variety of engineering fields. Many examples of such applications have been reported in the relevant literature \cite{urieli2011optimizing,maki2020application,schafroth2010modeling,fujii2018exploring,marsden2004optimal,hitz2017adaptive,sadeghi2014multi,bouzarkouna2012well,miyagi2019,NEURIPS2018_2de5d166,10.5555/3304415.3304617,volz2018evolving,10.1145/3449639.3459290,Nomura_Watanabe_Akimoto_Ozaki_Onishi_2021}. Evolutionary approaches are preferred for several reasons. First, because they do not require gradient information, they can be easily applied to simulation-based optimization, where the gradient is often unavailable. Second, they empirically exhibit better performance on problems with multiple local optima than approaches using local information, such as the gradient of the objective \cite{10.1145/1830761.1830790}.
Third, they can be easily accelerated by running simulations for multiple solution candidates in parallel.
% \textcolor{blue}{ [R3C2] : 
Hence, although evolutionary approaches tend to require a relatively high number of function evaluations, the execution time of the optimization process can be easily reduced \cite{miyagi@ghgt}.
% }

The objective of simulation-based optimization is typically to locate a solution whose performance is satisfactory in the real-world environment. 
For this purpose, we design a numerical model to simulate the objective function.
However, there is often a discrepancy between the numerical model and the real-world environment. 
The uncertainty in the real-world environment owing to limited information is a key reason for this divergence. 
Because of the uncertainty, there may be multiple numerical models that are consistent with the information of the real-world environment. 
A naive approach often applied in engineering optimization practice is to choose one such model, denoted as $f_\mathrm{sim}$, optimize (here, we assume minimization without loss of generality) $f_\mathrm{sim}$ and obtain a solution $x^*$, and evaluate it in the real-world environment, denoted by $f_\mathrm{real}$.
However, because of the discrepancy between $f_\mathrm{sim}$ and $f_\mathrm{real}$, the solution $x^*$ obtained in the simulation is not guaranteed to exhibit satisfactory performance in the real-world environment. 
That is, we may obtain $f_\mathrm{real}(x^*) \gg f_\mathrm{sim}(x^*)$.

A possible approach is to formulate the problem as a min--max optimization, where the objective is to locate the optimal solution to the worst-case objective function among all possible numerical models. 
Assume that we have $m$ different numerical models indexed by $s \in S = \{1, \dots, m\}$.
Then, the min--max optimization problem is formulated as
\begin{equation}
    \min_{x \in \X} \max_{s \in S} f(x, s) \label{eq:minmax} \enspace,
\end{equation}
where $x \in \mathbb{X} \subseteq \R^n$ represents the design variable, $s \in S$ is the scenario index, and $f: \mathbb{X} \times S \to \mathbb{R}$ is the objective function. 
It is recognized as the minimization of the worst-case objective function $F(x) = \max_{s \in S} f(x, s)$, which can be explicitly evaluated because $S$ is a finite set. 
Because $F$ is tractable, a simple evolutionary approach can be applied to minimize $F$.
However, each $F$ call costs (requires) $m$ simulations (i.e., $f$-calls), with a corresponding increase in the computation time with greater $m$.

The aim of this study is to develop an efficient approach to address min--max optimization \eqref{eq:minmax}. We assume the following. 
(I) The scenario set $S$ is a finite set; therefore, we can evaluate the worst-case objective $F$ analytically. 
(II) The $f$-call is computationally expensive, and the number of $f$-calls is the bottleneck of the optimization process. 
(III) $f$ is black-box and its gradient information is unavailable.  
A derivative-free approach such as an evolutionary approach is required. 
(IV) $f$ and $F$ are non-convex, and there are multiple local optima. 
To obtain a satisfactory local optimal solution, a restart strategy is required.
However, because each $F$-call requires $m$ $f$-calls and each $f$-call is computationally demanding, we may be able to perform only a few restarts within a given time budget if we naively optimize $F$. 
Based on the observation in the following motivational application, we further assume the following characteristics. 
(V) For any $s \in S$, there exists $x \in \X$ such that $F(x) = f(x, s)$. 
That is, without having all $s$, we cannot construct the true worst-case objective function.
However, (VI) if we limit our attention to the neighborhood of a local optimal solution, $F$ can be constructed using a subset $A \subset S$ with a relatively small cardinality $\abs{A} / \abs{S}$. 
Note also that characteristics (V) and (VI) are not requirements for the proposed approach to work properly and they cannot  be confirmed prior to the start of optimization.
Although characteristics (V) and (VI) are observations on a specific application, we conjecture that these characteristics appear in other simulation-based optimization problems. 
By utilizing characteristics (V) and (VI), we expect to reduce the number of $f$-calls to optimize $F$, resulting in more restarts and obtaining a better local optimum.

% \paragraph{Motivational Application}
% \textcolor{blue}{
\subsection{Motivational Application}
% }

Our motivational application considers well placement optimization for carbon dioxide capture and storage (CCS) projects \cite{CCS2005}.
CCS is a promising technique developed to reduce $\mathrm{CO_2}$ emissions by capturing $\mathrm{CO_2}$ in exhaust gases and injecting the captured $\mathrm{CO_2}$ into a reservoir deep underground through wells.
The objective of well placement optimization in CCS is to determine the well placement  that provides as much benefit (e.g., total $\mathrm{CO_2}$ injection volume) as possible with the least cost (e.g., drilling expenses) and risk (e.g., pressure build-up in the reservoir).
Simulation-based optimization is among the possible approaches for these projects.
The well placement is optimized through simulations that take a design variable, which encodes well coordinates, injection or production rate schedules, and well types (vertical / horizontal / multilateral), along with similar information, as an input and returns the abovementioned criteria.
To perform numerical simulations, experts design numerical models that describe geological conditions, such as the distribution of physical properties or boundary conditions resulting from geological surveys. 
However, in general, numerical models contain various uncertainties because geological surveys and investigations are limited. 
For instance, exploration wells are drilled to sample and investigate the physical properties of sites; however, they are insufficient to cover a vast geological formation, and the property distribution between exploration wells is highly uncertain. 
To deal with such uncertainties, multiple numerical models have been created using the same limited information. 
In previous studies, different criteria such as the average, the worst case, and the value at risk have been applied to obtain robust solutions under multiple numerical models \cite{miyagi2019, Yeten2003spe, Aitokhuehi2004, Artus2006comgeo, Alhuthali2010spe}. 
Among the different formulations, the min--max optimization \eqref{eq:minmax} is suitable to guarantee the commerciality or feasibility of the project in the worst scenario.

Here, we summarize the important characteristics of this application.
First, simulations executed to evaluate $f(x,s)$ are computationally demanding, as they require multicomponent and multiphase fluid simulators.
% \textcolor{blue}{[R3C4] : 
For example, previous studies \cite{miyagi@ghgt, yamaoto2012} have reported one simulation taking several hours to complete. 
% }
Therefore, the number of $f$-calls is limited for the optimization.
Second, the scenario set $S$ is a finite set. 
This uncertainty is represented by numerical models created by experts.
Although we cannot guarantee that the real-world environment is included in $S$, we expect the solution to the worst-case objective function under different numerical models created by experts to become reliable as we increase the number $m$ of scenarios. 
Third, $f(x, s)$ is a black box, and no gradient information is available. 
Fourth, $f$ and $F(x)$ are non-convex.  
Finally, we observe characteristics (V) and (VI) in \Cref{fig:2Dlandscape}, as discussed in \Cref{sec:rwa}.

% \paragraph{Related works}
% \textcolor{blue}{
\subsection{Related work} \label{sec:relatedworks}
% }

Studies on derivative-free worst-case optimization under finite scenarios may be categorized into two classes. 
One focuses on the smoothness of the worst-case objective function $F$.
It becomes naturally non-smooth owing to its construction. 
For some derivative-free optimization approaches, the smoothness of the objective function is important for its success.
% \textcolor{blue}{[R1C2]: 
Some studies have been conducted along these lines; for example, \cite{Liuzzi2006, Bogani2009, Warren2013}, in which $F$ was approximated by a smooth function, and a derivative-free optimization method was applied to minimize the approximate function so as not to fail to locate a local optimal solution owing to the non-smoothness of $F$.
The other research direction that has been explored involves reducing the computation cost.
The evaluation of the worst-case objective value $F(x)$ for each $x$ requires $m$ simulations to compute $f(x, s)$ for $s \in S$, and each $f$-call is computationally demanding.
Therefore, the computation cost is often a primary bottleneck in practice. 
Reducing the computation cost allows more restarts to be performed, thereby increasing the chance of obtaining better solutions if the objective is non-convex.
% \textcolor{blue}{[R1C2]: 
These two research directions are orthogonal, and these ideas may be combined. Nevertheless,
in this study, we focus on the latter topic.
% }

The following two approaches have been investigated to reduce the computation cost of the optimization. 

The first approach is to subsample scenarios from scenario set $S$ before optimization by using domain knowledge. In other words, the computation cost can be reduced by preliminarily decreasing the number of scenarios used for the optimization. 
This approach has been applied in the optimization of the designs or operations of oil fields. 
Prior to optimization, some criteria, such as the potential oil volume per scenario evaluated by numerical simulation, are prepared. 
Then, scenarios are subsampled based on the prepared criteria, and optimization is performed on the subsampled scenarios \cite{Ballin1992,Fenik2009, McLennan2005, Scheidt2009,Scheidt2009b,Scheidt2010,LI2014,Rahim2015math}.
This approach assumes that the worst-case objective function on the scenario set $S$ can be approximated on a subset $A$. 
However, this assumption is not generally satisfied. 
The approaches of this type are domain-specific and cannot be applied to other problems directly.

Surrogate-assisted approaches are the other primary alternative. They can reduce the number of $f$-calls by approximating $F(x)$ with a surrogate model trained during the optimization \cite{Zhou2010,Wang2016,Beielstein2017,Hansen2019}.
Because the quality of the surrogate model determines the effectiveness of these approaches, various methods such as a linear-quadratic model \cite{Hansen2019} or Kriging \cite{Beielstein2017} have been studied. These methods are effective when the objective function is smooth; however, the worst-case objective function $F$ becomes naturally non-smooth. 
Therefore, it may be difficult to select a proper surrogate model to approximate $F$.

% \clearpage{}

% \paragraph{Contributions}
\subsection{Contributions}

In this study, we develop and evaluate a novel approach for the min--max optimization problem \eqref{eq:minmax} with finite scenarios satisfying Assumptions (I)--(VI) described above.% 
The contributions of this study are summarized as follows.%
\footnote{
This study is an extension of the previous work in \cite{miyagi2021}.
The novel contributions of this paper are summarized as follows. (i) Our proposed approach, \pbilcma{}, is compared with a general-purpose surrogate-assisted CMA-ES, \texttt{lq-CMA-ES} \cite{Hansen2019}, on test problems in \Cref{sec:ex3}. (ii) \pbilcma{} is applied to a well placement optimization problem and its advantage over some existing approaches is demonstrated in \Cref{sec:rwa}. (iii) The sensitivities of the hyperparameters and the scalabilities of \pbilcma{} are analyzed in \ref{app:sensitivity} and \ref{app:scal}.
}

\begin{enumerate}

\item We define the notion of \emph{support scenarios} $S_\mathrm{support}(H)$ in a subset $H$ of the search space $\X$. This notion is used to describe the subset of scenarios that are sufficient to compute the worst-case objective function value for a solution candidate generated from the current search distribution with high probability. Assumptions (V) and (VI) above mean that the number of support scenarios is equivalent to the number of all scenarios at the beginning of the search, where the search distribution is widely spread, whereas it is significantly smaller if the search distribution is concentrated around a local optimal solution. We utilize this notion to develop the proposed approach and develop test problems. 

\item
We propose an \emph{adaptive scenario subset selection (AS3)} mechanism. \emph{AS3} attempts to reduce the number of $f$-calls required to compute the worst-case objective function values during the optimization by approximately sampling the support scenarios corresponding to the search distribution at each iteration. 
In contrast to general-purpose surrogate-assisted approaches, \emph{AS3} is specialized for worst-case optimization. 
Further, compared to domain-specific approaches that subsample scenarios prior to optimization based on some prior knowledge, \emph{AS3} does not require such prior knowledge.
\emph{AS3} mechanism was integrated into the CMA-ES. The resulting approach is called \pbilcma. Numerical experiments showed that \emph{AS3} mechanism follows the change in the support scenarios in the area $H_\gamma^t$, where the solution candidates are generated with probability $\gamma$ at iteration $t$. That is, \emph{AS3} successfully reduced the number of $f$-calls on problems where Assumption (VI) holds.

\item We compared \pbilcma{} with the brute-force approach optimizing $F$ directly using the CMA-ES and a surrogate-assisted approach, \texttt{lq-CMA-ES} \cite{Hansen2019}, on test problems. 
We confirmed that \pbilcma{} outperforms the brute-force approach in terms of the number of $f$-calls in most cases. Moreover, \pbilcma{} was more efficient than \texttt{lq-CMA-ES} for problems satisfying Assumption (VI), where \texttt{lq-CMA-ES} is more efficient if the number of support scenarios around the local optimal solution is close to the number of scenarios.

\item  
The effectiveness of \pbilcma{} was evaluated in a real-world application (well placement optimization for CCS) and comparison with \texttt{lq-CMA-ES} and the brute-force approach. 
The experimental results show that \pbilcma{} achieves a better solution within a given $f$-call budget than the compared approaches because of more restarts owing to its faster convergence, leading to a better local optimal solution for multimodal problems.   

\end{enumerate}

The remainder of this paper is organized as follows. 
The baseline approach---the brute-force approach optimizing $F$ using the CMA-ES---is introduced in \Cref{sec:cmaes}. 
The proposed approach is explained in \Cref{sec:approach}.
The numerical experiments performed to demonstrate the efficiency of the proposed algorithm are outlined in \Cref{sec:exp}. 
The comparisons carried out with the baseline approaches are also discussed. 
The utilization of the proposed algorithm for well placement optimization is presented in \Cref{sec:rwa}.
Concluding remarks are presented in \Cref{sec:conc}.
Some experimental results are provided in the appendices to further elucidate the usefulness of the proposed approach.

\section{CMA-ES: Covariance Matrix Adaptation Evolution Strategy (baseline approach)}\label{sec:cmaes}

Our baseline approach optimizes $x \in \mathbb{X}$ on the worst-case objective function $F$. For a solution candidate $x \in \mathbb{X}$, $F(x)$ can be evaluated by 
evaluating $m$ $f$-calls, i.e., $f(x, 1), \dots, f(x, m)$, and by taking their maximum $\max_{s \in S} f(x,s)$.

We employ the covariance matrix adaptation evolution strategy (CMA-ES) \cite{akimoto2019,Hansen2014,Hansen2001} with a restart strategy as the baseline approach.
The CMA-ES is recognized as a state-of-the-art derivative-free approach for black-box continuous optimization of non-convex functions \cite{10.1145/1830761.1830790,rios2013derivative}. The CMA-ES is a quasi-parameter-free approach.\footnote{%%%
The only parameter that is advisable to modify from the default value depending on the problem is the number $\lambda_x$ of the solution candidates generated at each iteration. 
A greater $\lambda_x$ tends to converge to a better local optimal solution if the problem is a well-structured multimodal problem \cite{Hansen2004}, while requiring more $f$-calls to converge. Restart strategies that run the CMA-ES with different (incremental) $\lambda_x$ have been proposed to alleviate the tedious parameter tuning for $\lambda_x$ \cite{Auger2018cec,10.1145/1570256.1570333}.
The successful performance of such a restart strategy has been reported in benchmarking \cite{Auger2018cec,10.1145/1830761.1830790} as well as real-world applications \cite{maki2020application}.}
That is, the users of this approach need not tune its hyperparameters on their own tasks, but rather can use it out-of-the-box.
This property and its superior performance have attracted practitioners, and hence, the CMA-ES has been widely applied to real-world applications \cite{urieli2011optimizing,maki2020application,schafroth2010modeling,fujii2018exploring,marsden2004optimal,hitz2017adaptive,sadeghi2014multi,bouzarkouna2012well,miyagi2019,NEURIPS2018_2de5d166,10.5555/3304415.3304617,volz2018evolving,10.1145/3449639.3459290,Nomura_Watanabe_Akimoto_Ozaki_Onishi_2021}.

\begin{wrapfigure}{r}{0.45\textwidth}
\begin{minipage}{\hsize}
\begin{algorithm}[H]
\caption{The baseline approach \label{alg:baseline}}
\begin{algorithmic}[1]\small
\Require $m^0 \in \R^n$, $\Sigma^0 \in \R^{n \times n}$
\State $\lambda_x \leftarrow \lfloor 4 + 3\log(n) \rfloor$
\For{$t = 0, \dots, T-1$}
\For{$k = 1, \dots, \lambda_x$}
\State Sample $x^{t}_{k} \sim \mathcal{N}(m^{t}, \Sigma^{t})$.
\State Evaluate $f^{t}_{k,s} = f(x^{t}_{k}, s)$ for all $s \in S$.
\State Set $F^{t}_{k} = \max_{s \in S} f^{t}_{k,s}$.
\EndFor
\State Perform CMA-ES update using $\{(x^{t}_{k}, F^{t}_{k})\}_{k=1}^{\lambda_x}$
\If{converged}
\State Reset $m^{t+1}$ and $\Sigma^{t+1}$ (and possibly change $\lambda_x$)
\EndIf
\EndFor
%\State \Return $m^{T}$
\end{algorithmic}
\end{algorithm}
\end{minipage}
\end{wrapfigure}

Our baseline approach---the CMA-ES, optimizing the worst-case objective $F$---is outlined in \Cref{alg:baseline}.
The population size $\lambda_x$, i.e., the number of solution candidates generated at each iteration, is set based on the search space dimension $n$. 
At each iteration, the CMA-ES generates $\lambda_x$ solution candidates, $x_k^t$ ($k=1,\dots,\lambda_x$) from the multivariate normal distribution $\mathcal{N}(m^t, \Sigma^t)$ with mean $m^t$ and $\Sigma^t$. 
Each solution candidate is evaluated on $f(\cdot, s)$ for all scenarios $s \in S$. 
Then, the worst-case objective function value $F(x_k^t)$ for each $x_k^t$ is computed by $\max_{s \in S} f(x_k^t, s)$ and assigned to $F_{k}^t$. 
Using the pairs $\{(x_k^t, F_k^t)\}$ of the solution candidates and their worst-case objective function values, the CMA-ES updates the distribution parameters $m$ and $\Sigma$, and other dynamic parameters used for their updates. 
These updates are known to follow the natural gradient of the expected objective function value \cite{10.1007/s00453-011-9564-8}. 
These steps are repeated until the distribution is regarded as converged.
Once the distribution converges, the current mean vector is registered as a candidate local optimal solution.
Then, the mean vector and the covariance matrix are reset for restart.

Of note, this brute-force approach to optimizing $F$ involves an important limitation. 
Because $F$ is assumed to be non-convex and possibly multimodal, it is essential to perform as many restarts as possible. 
However, to evaluate the worst-case objective function value $F(x)$ for each solution candidate $x$, $m$ $f$-calls are required. 
That is, as the number $m$ of the scenarios increases, this approach can perform fewer restarts, possibly leading to a poorer local optimal solution.

\section{Adaptive Scenario Subset Selection (AS3) Mechanism}\label{sec:approach}

We propose a new approach, namely \emph{adaptive scenario subset selection} (\emph{AS3}), to reduce the number of $f$-calls for each restart and to enable more restarts to be performed for a better local optimal solution. 
This section presents the design principles and details of \emph{AS3}.

\subsection{Design Principles}

Our idea is to save $f$-calls when computing the worst-case objective function value $F$ at each iteration of \Cref{alg:baseline} without changing its behavior. 
For this purpose, we wish to subsample a scenario set $A^t \subseteq S$ at each iteration $t$ such that $F(x_k^t) = F(x_k^t;A^t)$ for all $k = 1,\dots,\lambda_x$, where $F(x; A^t)$ is the worst-case objective function under the scenario subset $A^t$ and is defined as $F(x; A^t) = \max_{s \in A^t} f(x, s)$.
If we can select a subset $A^t$ with $\abs{A^t} / \abs{S} < 1$, we can save $\abs{S} - \abs{A_t}$ $f$-calls for the evaluation of the worst-case objective function value $F(x_k^{t})$ of each solution candidate $x_k^{t}$ without changing the algorithmic behavior.
However, $A^t$ cannot be known without evaluating $f(x_k^t, s)$ for all $s \in S$ as $f$ is a black box. 
Therefore, we estimate $A^t$ during the optimization.

To estimate $A^t$, we utilize information about a neighborhood in which solution candidates are expected to be generated at each iteration. 
Let $x$ be $\mathcal{N}(m, \Sigma)$-distributed. 
Then, it is easy to see that $(x - m)^\mathrm{T} \Sigma^{-1} (x - m)$ is $\chi^2_n$-distributed with the degrees of freedom of $n$. 
In other words, $(x - m)^\mathrm{T} \Sigma^{-1} (x - m) \leq P_{\chi_n^2}^{-1}(\gamma)$ with probability $\gamma \in [0, 1]$, where $P_{\chi_n^2}$ is the cumulative density function of a $\chi^2$ distribution with $n$ degrees of freedom. 
Because the CMA-ES samples solution candidates from $\mathcal{N}(m^t, \Sigma^t)$, each solution candidate falls into
\begin{equation}
  H^t_{\gamma} = \left\{x \in \R^n : (x - m^t)^\T (\Sigma^t)^{-1} (x - m^t) \leq P_{\chi_n^2}^{-1}(\gamma) \right\},
  \label{eq:hgammat}
\end{equation}
with probability $\gamma$. 
Therefore, if we can select $A^t \subseteq S$ such that $F(x) = F(x; A^t)$ for all $x \in H_{\gamma}^t$, and we set $\gamma$ close to $1$, 
we need to evaluate $f(x, s)$ only for all $s \in A^t$ to compute $F(x)$, i.e., $\abs{A^t}$ $f$-calls are required. Hence,
we can omit $\abs{S \setminus A^t}$ $f$-calls for each solution candidate while mimicking the behavior of the baseline approach (\Cref{alg:baseline}),
which evaluates $f(x, s)$ for all $s \in S$ to compute $F(x)$, requiring $\abs{S}$ $f$-calls.

To formalize our idea, we define the notion of support scenarios.
\begin{definition}[Support Scenario]
A scenario $s \in S$ such that $F(x) = f(x, s)$ at $x \in \mathbb{X}$ is called a \emph{support scenario} of $x$. 
A scenario $s \in S$ is called a \emph{support scenario in neighborhood $H \subseteq \mathbb{X}$} if there exists $x \in H$ such that $F(x) = f(x, s)$.
The set of support scenarios in $H$ is denoted by $S_\mathrm{support}(H) \subseteq S$. 
If $s \in S$ is a support scenario in all neighborhoods $H$ of $x^*\in \mathbb{X}$, it is called a \emph{support scenario around $x^*$}. 
The set of support scenarios around $x^*$ is denoted by $S_\mathrm{support}(x^*) = \bigcap_{x^*} S_\mathrm{support}(H) \subseteq S$, where $\bigcap_{x^*}$ is the intersection of all neighborhoods of $x^*$.
\end{definition}

Our idea is to select $A^t = S_{\mathrm{support}}(H_\gamma^t)$ at each iteration and use $F(x; S_{\mathrm{support}}(H_\gamma^t))$ as a surrogate of $F$. 
Then, the worst-case objective function value of each solution candidate is correctly computed with probability $\gamma$, whereas $\abs{S} - \abs{S_{\mathrm{support}}(H_\gamma^t)}$ $f$-calls are omitted. 
The smaller $\abs{S_\mathrm{support}(H^t_{\gamma})} / \abs{S}$, the more efficient the subsampling becomes. 

Although the worst-case objective function values of $(1 - \gamma) \cdot \lambda_x$ solution candidates are underestimated ($F(x) \geq F(x; A^t)$), the effect of such solution candidates on the algorithmic behavior is expected to be small.
This is because the CMA-ES is a ranking-based approach (hence, the magnitude of the difference $F(x) - F(x; A^t)$ itself does not matter), and the ranking change due to the underestimation of $F(x)$ is restricted.
\begin{remark}\label{rem}
To estimate the effect of the underestimation of $F(x)$ by $G(x) = F(x; S_{\mathrm{support}}(H_\gamma^t))$, we consider the population version of Kendall's rank correlation $\tau$, defined as
\begin{align}
\tau = \Pr[(F(X) - F(Y))(G(X) - G(Y)) > 0] - \Pr[(F(X) - F(Y))(G(X) - G(Y)) < 0] \enspace, \label{eq:tau}
\end{align}
where $X$ and $Y$ are solution candidates and are independently $\mathcal{N}(m^t, \Sigma^t)$-distributed.
For technical simplicity, we assume that all the level sets of $F$ and $G$ have zero Lebesgue measure (roughly speaking, there is no constant area).
Then, because $\Pr[F(X) = F(Y)] = \Pr[G(X) = G(Y)] = 0$, we have
\begin{multline}
    \Pr[(F(X) - F(Y))(G(X) - G(Y)) > 0] 
    = \Pr[(F(X) - F(Y))(G(X) - G(Y)) \geq 0] \\
    = 1 - \Pr[(F(X) - F(Y))(G(X) - G(Y)) \leq 0]
    = 1 - \Pr[(F(X) - F(Y))(G(X) - G(Y)) < 0]. 
\end{multline}
Therefore, $\tau = 2 \Pr[(F(X) - F(Y))(G(X) - G(Y)) \geq 0] - 1$. 
If $X$ and $Y$ are both in $H_\gamma^t$, which occurs with probability $\Pr[X, Y \in H_\gamma^t]=\gamma^2$, we have $F(X) = G(X)$ and $F(Y) = G(Y)$. Then, we have 
\begin{equation}
\Pr[(F(X) - F(Y))(G(X) - G(Y)) \geq 0 \mid X, Y \in H_\gamma^t] 
 = \Pr[(F(X) - F(Y))^2 \geq 0\mid X, Y \in H_\gamma^t] = 1 \enspace.
\end{equation}
Hence, 
\begin{equation}
\begin{split}
\MoveEqLeft[2]
\Pr[(F(X) - F(Y))(G(X) - G(Y)) \geq 0]
\\
&=
\Pr[(F(X) - F(Y))(G(X) - G(Y)) \geq 0 \mid X, Y \in H_\gamma^t] \Pr[X, Y \in H_\gamma^t] \\
&\quad +
\Pr[(F(X) - F(Y))(G(X) - G(Y)) \geq 0 \mid \neg (X, Y \in H_\gamma^t)] \Pr[\neg (X, Y \in H_\gamma^t)]
\\
&\geq \Pr[(F(X) - F(Y))(G(X) - G(Y)) \geq 0 \mid X, Y \in H_\gamma^t] \Pr[X, Y \in H_\gamma^t] = \gamma^2 \enspace.
\end{split}\label{eq:positivecorrelation}
\end{equation}
Finally, we obtain $\tau \geq 2 \gamma^2 - 1$.
That is, the rank correlation between $F(x)$ and $F(x; S_{\mathrm{support}}(H_\gamma^t))$ under $x \sim \mathcal{N}(m^t, \Sigma^t)$ is lower-bounded by $2 \gamma^2 - 1$, which can be made arbitrarily close to one by setting $\gamma$ close to $1$. 
Although we do not analyze the relation between $\tau$ and the algorithmic behavior theoretically here, $\tau$ is used to measure the goodness of surrogate models in practice \cite{Hansen2019,10.1145/3321707.3321709}.
Hence, we expect that a sufficiently high $\tau$ value will lead to sufficiently close behavior.
\end{remark}

To estimate $S_\mathrm{support}(H_\gamma^t)$ during optimization, we introduce $p^t = (p_1^t, \cdots, p_m^t) \in [0, 1]^{m}$, where $p_s^t$ indicates the certainty of the algorithm whether $s \in S$ is in $S_\mathrm{support}(H_\gamma^t)$. 
Ideally, we want $p_s^t = \ind{s \in S_\mathrm{support}(H_\gamma^t)}$, where it is $1$ if $s \in S_\mathrm{support}(H^t_\gamma)$ and $0$ otherwise.
Then, by using $p_s^t$ as the probability of sampling a scenario $s \in S$ to construct a subset $A^t$, we obtain $A^t = S_\mathrm{support}(H_\gamma^t)$. 
Because the search distribution of the CMA-ES changes gradually over the series of iterations, $H_\gamma^t$ also does so. 
Then, we expect that $S_\mathrm{support}(H_\gamma^t)$ also changes gradually with iteration.
Therefore, we maintain $p^t$ over iterations.
In the next section, we describe a heuristic approach to maintain $p^t$. 

\subsection{Parameter Update}

At each iteration, we sample $A^t \subseteq S$ by using a binomial distribution with probability $p_s^t$ of sampling scenario $s$ and evaluate the solution candidates $x_1^t, \dots, x_{\lambda_x}^t$ on $f(x, s)$ for $s \in A^t$. 
Therefore, we can observe whether $F(x_k^t; A^t) = f(x_k^t; s)$ for each $s \in A^t$. 
If $S_{\mathrm{support}}(H_\gamma^t) \subseteq A_t$, $F(x_k^t; A^t) = f(x_k^t; s)$ for some $x_k^t \in H_\gamma^t$ indicates that $s \in S_{\mathrm{support}}(H_\gamma^t)$. 
If $S_{\mathrm{support}}(H_\gamma^t) \not\subseteq A_t$, $F(x_k^t; A^t) = f(x_k^t; s)$ for some $x_k^t \in H_\gamma^t$ does not necessarily mean $s \in S_{\mathrm{support}}(H_\gamma^t)$.
Because the algorithm cannot distinguish the above two situations, we increase $p_s^t$ for such scenarios.
However, we know that $F(x_k^t; A^t) > f(x_k^t; s)$ for all $x_k^t \in H_\gamma^t$ does not provide information on whether $s \in S_{\mathrm{support}}(H_\gamma^t)$ because there may exist $x \in H_{\gamma}^t\setminus\{x_k^t\}_{k=1}^{\lambda_x}$ such that $F(x) = f(x, s)$. 
However, the probability of such an event is $(1 - \Pr[F(x) = f(x, s) \wedge x \in H_{\gamma}^{t} \mid x \sim \mathcal{N}(m^t, \Sigma^t)])^{\lambda_x}$, which is sufficiently small as we set $\lambda_x$ to a large value. 
Therefore, we decrease $p_s^t$ for scenarios with $F(x_k^t; A^t) > f(x_k^t; s)$ for all $x_k^t \in H_\gamma^t$. 
For $s \not\in A^t$, we have no information on whether $s \in S_\mathrm{support}(H_\gamma^t)$. 
Hence, we keep $p_s^t$. 

To realize this idea, we update $p^t_s $ as $p^{t+1}_s = p^{t}_s + \Delta^t_s$, where
\begin{equation}
\Delta_s^t = \ind{s \in A^t} \cdot \left( c_p \cdot \sum_{i=1}^{\lambda_x} \ind{F(x^{t}_{i} ; A^t) = f(x^{t}_{i}, s) \wedge x^{t}_{i} \in H^t_\gamma} 
 - c_n \cdot \prod_{i=1}^{\lambda_x} (1 - \ind{F(x^{t}_{i} ; A^t) = f(x^{t}_{i}, s) \wedge x^{t}_{i} \in H^t_\gamma}) \right)
  \enspace. \label{eq:delta}
\end{equation}
Here, $c_p$ is the learning rate for the increase in $p_s^t$, and $c_n$ is the learning rate for the decrease in $p_s^t$. 
The first term is $c_p$ times the number of solution candidates for which $s$ is the support scenario. 
The second term is $-c_n$ if $s$ is not a support scenario for any solution candidate. 
Because we use $p_s^t$ as the sampling probability, it must be in $[0, 1]$. 
To keep $p_s^t \in [\epsilon, 1]$ for some $\epsilon > 0$, we clip $p_s^{t+1}$ into $[\epsilon, 1]$.
The minimal probability $\epsilon > 0$ is introduced because if $p_s^t = 0$, then $s$ will never be sampled and $\Delta_s^{t'} = 0$ for all $t' \geq t$. 

\subsection{Expected Behavior}\label{sec:expectation}

First, we investigate the expected behavior of \eqref{eq:delta} to determine where the probability $p_s^t$ is increased. 
To answer this question, consider the expectation of $\Delta_s^t$
\begin{equation}
  \E[\Delta_s^{t} \mid p^t_s]
  =  p_s^t \cdot \left(\lambda_x \cdot c_p \cdot \Pr[F(x ; A^t) = f(x, s) \wedge x \in H^t_\gamma] 
  - c_n (1 - \Pr[F(x; A^t) = f(x, s) \wedge x \in H^t_\gamma])^{\lambda_x} \right) \enspace.
\end{equation}
Then, it may be easily observed that $\E[\Delta_s^{t} \mid p^t_s]$ is positive if and only if 
\begin{align}
  \frac{\lambda_x \cdot \Pr[F(x) = f(x, s) \wedge x \in H^t_\gamma] }{(1 - \Pr[F(x) = f(x, s) \wedge x \in H^t_\gamma])^{\lambda_x}} > \frac{c_n}{c_p} \enspace.
  \label{eq:p_condition}
\end{align}
Note that the left-hand side (LHS) of \eqref{eq:p_condition} increases with $\Pr[F(x) = f(x, s) \wedge x \in H^t_\gamma]$. 
Therefore, the condition can be written as $\Pr[F(x) = f(x, s) \wedge x \in H^t_\gamma] > \delta(\lambda_x, c_n/c_p)$ with some function $\delta(\lambda_x, c_n/c_p)$. 
In other words, with the update formula \eqref{eq:delta}, we can increase $p_s^t$ only for scenarios with sufficiently large $\Pr[F(x) = f(x, s) \wedge x \in H^t_\gamma]$. 
Let $S_\gamma^{t}(\lambda_x, c_n/c_p)$ be the set of scenarios satisfying \eqref{eq:p_condition}. Therefore, $p_s^t$ is increased in the expectation of $s \in S_\gamma^{t}(\lambda_x, c_n/c_p)$. 

Second, we investigate how small the probability $\Pr[F(x) = f(x, s) \wedge x \in H^t_\gamma]$ is for $s \in S_\mathrm{support}(H_\gamma^t) \setminus S_\gamma^{t}(\lambda_x, c_n/c_p)$. 
For $\lambda_x \to \infty$, the LHS of \eqref{eq:p_condition} diverges to $+\infty$ unless $\Pr[F(x) = f(x, s) \wedge x \in H^t_\gamma] = 0$. Therefore, $\delta(\lambda_x, c_n/c_p) \to 0$ as $\lambda_x \to \infty$, and $S_\gamma^{t}(\lambda_x, c_n/c_p) = S_\mathrm{support}(H_\gamma^t)$ for a sufficiently large $\lambda_x$. That is, $p_s^t$ increases if and only if $s \in S_\mathrm{support}(H_\gamma^t)$, which is a promising behavior. 
For a finite $\lambda_x$, using the approximation $(1 - \Pr[F(x) = f(x, s) \wedge x \in H^t_\gamma])^{\lambda_x} \approx 1-\lambda_x \Pr[F(x) = f(x, s) \wedge x \in H^t_\gamma]$, condition \eqref{eq:p_condition} is approximated as 
\begin{align}
  \Pr[F(x) = f(x, s) \wedge x \in H^t_\gamma] > \delta(\lambda_x, c_n/c_p) \approx \frac{c_n}{\lambda_x (c_p + c_n)} \enspace.
  \label{eq:p_condition_approx}
\end{align}
In other words, the scenarios with $\Pr[F(x) = f(x, s) \wedge x \in H^t_\gamma] \lessapprox \frac{c_n}{\lambda_x (c_p + c_n)}$ may not be included in $S_\gamma^{t}(\lambda_x, c_n/c_p)$, whereas $s \in S_\mathrm{support}(H_\gamma^t)$. 

Third, we investigate the probability that a solution candidate $x$ such that $F(x) > F(x; S_\gamma^t(\lambda_x, c_n/c_p))$ may occur is generated from $\mathcal{N}(m^t, \Sigma^t)$.
% \textcolor{red}{
Because $F(x; \emptyset)$ is undefined, we define $\tilde{S}_\gamma^t$ as $\tilde{S}_\gamma^t = S_\gamma^t(\lambda_x, c_n/c_p)$ if $\abs{S_\gamma^t(\lambda_x, c_n/c_p)} \geq 1$ and $\tilde{S}_\gamma^t = \{\tilde{s}\}$ with a uniform-randomly sampled $\tilde{s} \in \{1, \dots, m\}$ if $S_\gamma^t(\lambda_x, c_n/c_p) = \emptyset$, and consider $F(x; \tilde{S}_\gamma^t)$ instead of $F(x; S_\gamma^t(\lambda_x, c_n/c_p))$.
% }
Scenarios with $\delta(\lambda_x, c_n/c_p) \geq \Pr[F(x) = f(x, s) \wedge x \in H^t_\gamma] > 0$ are included in $S_\mathrm{support}(H_\gamma^t)$ but not in $S_\gamma^t(\lambda_x, c_n/c_p)$.
Therefore, the probability of a solution candidate for which $F(x) > F(x; 
% \textcolor{red}{
\tilde{S}_\gamma^t
% }
)$ may occur is
\begin{equation}
    \sum_{s=1}^m \ind{s \in S_\mathrm{support}(H^t_{\gamma}) \setminus
    % \textcolor{red}{
    \tilde{S}_\gamma^t
    % }
    } \cdot \Pr[F(x) = f(x, s) \wedge x \in H_\gamma^t ] 
    \leq \delta(\lambda_x, c_n/c_p) \cdot \abs{S_{\mathrm{support}}(H_\gamma^t) \setminus 
    % \textcolor{red}{
    \tilde{S}_\gamma^t}
    % } 
    \enspace,
\end{equation}
where the equality holds if $\Pr[F(x) = f(x, s) \wedge x \in H^t_\gamma] = \delta(\lambda_x, c_n/c_p)$ for all $s \in S_\mathrm{support}(H_\gamma^t) \setminus 
% \textcolor{red}{
\tilde{S}_\gamma^t
% }
$.
% \textcolor{red}{
Note that $\abs{S_{\mathrm{support}}(H_\gamma^t) \setminus 
% \textcolor{red}{
\tilde{S}_\gamma^t
% }
} \leq m - 1$.
% }
Therefore, in the worst case, $F(x)$ is underestimated by $F(x; 
% \textcolor{red}{
\tilde{S}_\gamma^t
% }
)$ with a probability of at most $\delta(\lambda_x, c_n/c_p) \cdot (m - 1)$.
By the same argument as in \Cref{rem}, the population version of the Kendall rank correlation $\tau$ between $F(x)$ and $F(x; \tilde{S}_\gamma^t)$ is lower-bounded by $2 (1 - \delta(\lambda_x, c_n/c_p) \cdot (m - 1))^2 - 1$.  
To bound $\tau$ by a constant, we need to set $c_n$ sufficiently small such that $\delta(\lambda_x, c_n/c_p) \leq \eta / (m - 1)$ for some $\eta > 0$. 
\subsection{\pbilcma}

The proposed scheme, \emph{AS3}, was combined with the baseline CMA-ES (\Cref{alg:baseline}). 
The resulting algorithm, \pbilcma \footnote{Our implementation is publicly available at \url{https://gist.github.com/a2hi6/2f1989dc311e41df2181c250c941a54c}.}, is provided in \Cref{alg:cma-PBIL}. 
In Lines~3--9, a subset $A^t \subseteq S$ is constructed.
Each scenario $s$ is included with probability $p_s^t$.
To avoid $A^t$ being empty, we sample a scenario from a categorical distribution $\text{Cat}(p^t_s / \sum_{s=1}^{m} p_{s}^t)$ with probability vector $p^t_s / \sum_{s=1}^{m} p_s^t$ if $A^t = \emptyset$.
In Lines~10--14, $\lambda_x$ solution candidates are sampled, and their objective values are evaluated for $s \in A^t$.
In Line~15, the parameters of the CMA-ES are updated.
In Lines~16 and 17, $p^t$ is updated. 
In Lines~18--20, a restart with doubled population size is performed.

\begin{minipage}[t]{0.47\textwidth}
\centering
\vspace{0pt}
\begin{algorithm}[H]
  \caption{\pbilcma{}}  \label{alg:cma-PBIL}
  \begin{algorithmic}[1]\small
    \Require $m^0 \in \R^n$, $\Sigma^0 \in \R^{n \times n}$, $p^{0} \in [\epsilon, 1]^m$
    \Require $c_p \in (0, 1]$, $\eta \in (0, 1]$, $\gamma > 0$, $\epsilon > 0$
\State $\lambda_x \leftarrow \lfloor 4 + 3\log(n) \rfloor$
    \For{$t = 0, \dots, T-1$}
    \State $A^{t} = \emptyset$
    \For{$s = 1, \dots, m$}
    \State $A^{t} \leftarrow A^{t}\cup\{s\}$ with probability $p_s^t$
    \EndFor
    \If{$A^t = \emptyset$}
    \State $A^t\! \leftarrow\! \{s\}$ with $s \sim \text{Cat}(p^t_s / \sum_{s=1}^{m} p_{s}^t)$
    \EndIf
    \For{$k = 1, \dots, \lambda_x$}
    \State Sample $x^{t}_{k} \sim \mathcal{N}(m^{t}, \Sigma^{t})$.
    \State Evaluate $f^{t}_{k,s} = f(x^{t}_{k}, s)$ for all $s \in A^t$.
    \State Set $F^{t}_{k} = F(x^{t}_{k}; A^t) = \max_{s \in A^t} f^{t}_{k,s}$.
    \EndFor
    \State Perform CMA-ES update using $\{(x^{t}_{k}, F^{t}_{k})\}_{k=1}^{\lambda_x}$.
    \State Compute $c_n$ as \eqref{eq:cn} and $\Delta^t$ as \eqref{eq:delta}.
    \State Update $p^{t+1} = \texttt{clip}(p^t + \Delta^t; \epsilon, 1)$.
\If{converged}
\State Reset $m^{t+1}$, $\Sigma^{t+1}$ and $p^{t+1}$ and $\lambda_x \leftarrow 2 \cdot \lambda_x$.
\EndIf
    \EndFor
    \State \Return $m^{T}$
  \end{algorithmic}
\end{algorithm}
\vspace{0pt}
\end{minipage}%
\begin{minipage}[t]{0.01\textwidth}
\ 
\end{minipage}%
\begin{minipage}[t]{0.47\textwidth}
\centering
\vspace{0pt}
\begin{algorithm}[H]
  \caption{\pbilcma\ with fixed $\lambda_s$} \label{alg:cma-PBIL-fixed}
  \begin{algorithmic}[1]\small
    \Require $m^0 \in \R^n$, $\Sigma^0 \in \R^{n \times n}$, $p^{0} \in [\epsilon, 1]^m$
    \Require $c_p \in (0, 1]$, $\gamma > 0$, $\lambda_s > 0$, $\epsilon > 0$
\State $\lambda_x \leftarrow \lfloor 4 + 3\log(n) \rfloor$
    \For{$t = 0, \dots, T-1$}
    \State $A^{t} = \emptyset$
    \While{$\abs{A^t} < \lambda_s$}
    \State Sample $s \sim \text{Cat}(p^t_s / \sum_{s=1}^{m} p_{s}^t)$.
    \State $A^{t} \leftarrow A^{t}\cup\{s\}$ if $s \notin A^t$
    \EndWhile
    \For{$k = 1, \dots, \lambda_x$}
    \State Sample $x^{t}_{k} \sim \mathcal{N}(m^{t}, \Sigma^{t})$.
    \State Evaluate $f^{t}_{k,s} = f(x^{t}_{k}, s)$ for all $s \in A^t$.
    \State Set $F^{t}_{k} = F(x^{t}_{k}; A^t) = \max_{s \in A^t} f^{t}_{k,s}$.
    \EndFor
    \State Perform CMA-ES update using $\{(x^{t}_{k}, F^{t}_{k})\}_{k=1}^{\lambda_x}$.
    \State Compute $c_n$ as \eqref{eq:cn2} and $\Delta^t$ as \eqref{eq:delta}.
    \State Update $p^{t+1} = \texttt{clip}(p^t + \Delta^t; \epsilon, 1)$.
\If{converged}
\State Reset $m^{t+1}$, $\Sigma^{t+1}$ and $p^{t+1}$ and $\lambda_x \leftarrow 2 \cdot \lambda_x$.
\EndIf
    \EndFor
    \State \Return $m^{T}$
  \end{algorithmic}
\end{algorithm}
\vspace{0pt}
\end{minipage}%

Instead of letting $c_n$ be a user parameter, we introduce $\eta > 0$ and set $c_n$ depending on $c_p$, $\eta$, $m$, and $\lambda_x$ as
\begin{equation}
  c_n = c_p \cdot  \left( \frac{\eta \lambda_x}{\max\{ m - \eta \lambda_x - 1, \eta \lambda_x\}} \right) \enspace. \label{eq:cn}
\end{equation}
The rationale is as follows.
As we discussed in \Cref{sec:expectation}, the Kendall rank correlation $\tau$ between $F(x)$ and $F(x; \tilde{S}_\gamma^t)$ can be as small as $2 (1 - \delta(\lambda_x, c_n/c_p) (m - 1))^2 - 1$. 
Here, $p_s^t$ for all $s \in S_\gamma^t(\lambda_x, c_n/c_p)$ is expected to increase; hence, we expect that $p_s^t$ eventually approaches $\ind{s \in S_\gamma^t(\lambda_x, c_n/c_p)}$ and $A^t$ is considered as a realization of 
$
% \textcolor{red}{
\tilde{S}_\gamma^t
% }
$. 
Therefore, 
the above $\tau$ value is expected to approximate the $\tau$ between $F(x)$ and $F(x; A^t)$.
Then, we aim to keep the $\tau$ value as high as possible so that we do not change the behavior of the baseline CMA-ES significantly. For this purpose, we need to set $c_n$ such that $\delta(\lambda_x, c_n/c_p) \leq \eta / (m - 1)$ for some $\eta > 0$, as described in \Cref{sec:expectation}. 
By applying the approximation of $\delta(\lambda_x, c_n/c_p)$ given in the right-hand side of \eqref{eq:p_condition_approx}, we obtain
\begin{equation}
\frac{c_p}{c_n}
     \geq \frac{m - 1 - \lambda_x\eta}{\lambda_x\eta} 
     \enspace.
\end{equation}
That is, $c_n$ needs to be set carefully depending on $m$, $\lambda_x$, and $\eta$. 
To absorb the dependency between the user parameters, we let $\eta$ be the user parameters and $c_n$ be computed automatically from $c_p$, $\eta$, $\lambda_x$, and $m$.

\subsection{\pbilcma{} with fixed $\lambda_s$}

For comparison purposes, we propose a variant of \pbilcma{} that samples a fixed number $\lambda_s$ of scenarios in each iteration, as detailed in \Cref{alg:cma-PBIL-fixed}. 
In contrast to \Cref{alg:cma-PBIL}, we sample $\lambda_s$ scenarios from a categorical distribution $\text{Cat}(p^t_s / \sum_{s=1}^{m} p_{s}^t)$ without replacement. 
If $\lambda_s = m$, it is identical to \Cref{alg:baseline}. 
The other difference is the setting of $c_n$. We use the following formula.
\begin{equation}
  c_n = \frac{c_p \cdot \lambda_x}{ \abs{ \{ s \in A^t \mid f(x_{i}^{t}, s) < F(x_i^{t}; A^t)\ \forall i=1,\dots,\lambda_x\}} } \enspace. \label{eq:cn2}
\end{equation}
Because the sum $\sum_{s = 1}^{m} p_s^t$ is irrelevant in this variant, we force $\sum_{s=1}^{m} \Delta_s^t = 0$ by using \eqref{eq:cn2}. 

The main purpose of presenting this variant is to demonstrate the efficiency of the \emph{AS3} mechanism in \pbilcma. 
If $\lambda_s < \abs{S_\mathrm{support}(x^*)}$, where $x^*$ is the optimal solution to $F$, \pbilcma{} with fixed $\lambda_s$ cannot approximate the worst-case objective function around the optimal solution, and it may fail to converge toward $x^*$. 
Therefore, $\lambda_s$ is a sensitive user parameter, and its adequate value cannot be determined in advance. 
\Cref{alg:cma-PBIL} is advantageous over \Cref{alg:cma-PBIL-fixed} in that the number $\abs{A^t}$ of sampled scenarios is adapted during the optimization. Note that its expected value is $\sum_{s=1}^{m} p_s^t$. 
By comparing the performances of \Cref{alg:cma-PBIL} and \Cref{alg:cma-PBIL-fixed} with $\lambda_s = \abs{S_\mathrm{support}(x^*)}$, we also show the efficiency of \pbilcma{}. 
The effect of $\lambda_s$ in \Cref{alg:cma-PBIL-fixed} is investigated in \ref{sec:lambdas}.

\section{Numerical Evaluation on Test Problems}\label{sec:exp}

We compare \pbilcma{} with the baseline approaches, \texttt{CMA-ES} (\Cref{alg:baseline}), \pbilcma{} with fixed $\lambda_s$ (\Cref{alg:cma-PBIL-fixed}), and a surrogate-assisted approach \texttt{lq-CMA-ES} \cite{Hansen2019} through numerical experiments on the test problems. 
In particular, we confirm the following hypotheses.
(1) \pbilcma{} updates $p^t_s$ for each $s \in S$ to follow the indicator value $\ind{s \in S_\mathrm{support}(H^t_\gamma)}$ (\Cref{sec:ex1}).
(2) \pbilcma{} is more efficient in terms of the number of $f$-calls than \texttt{CMA-ES} if $\abs{S_\mathrm{support}(x^*)} < m$. The efficiency is particularly high for smaller $\abs{S_\mathrm{support}(x^*)} / m$ (\Cref{sec:ex2}).
(3) \pbilcma{} is competitive with \pbilcma{} with fixed $\lambda_s = \abs{S_\mathrm{support}(x^*)}$ (\Cref{alg:cma-PBIL-fixed}) (\Cref{sec:ex3}).
(4) \pbilcma{} is more efficient than \texttt{lq-CMA-ES} 
if $\abs{S_\mathrm{support}(x^*)} / m$ is relatively small.
By contrast, \texttt{lq-CMA-ES} is more efficient than \pbilcma{} if $\abs{S_\mathrm{support}(x^*)} / m \approx 1$ (\Cref{sec:ex4}).

\subsection{Test problems}\label{sec:test}

\providecommand{\figsizefunc}{0.2}
\begin{figure*}[t]
  \centering
    \includegraphics[width=\hsize]{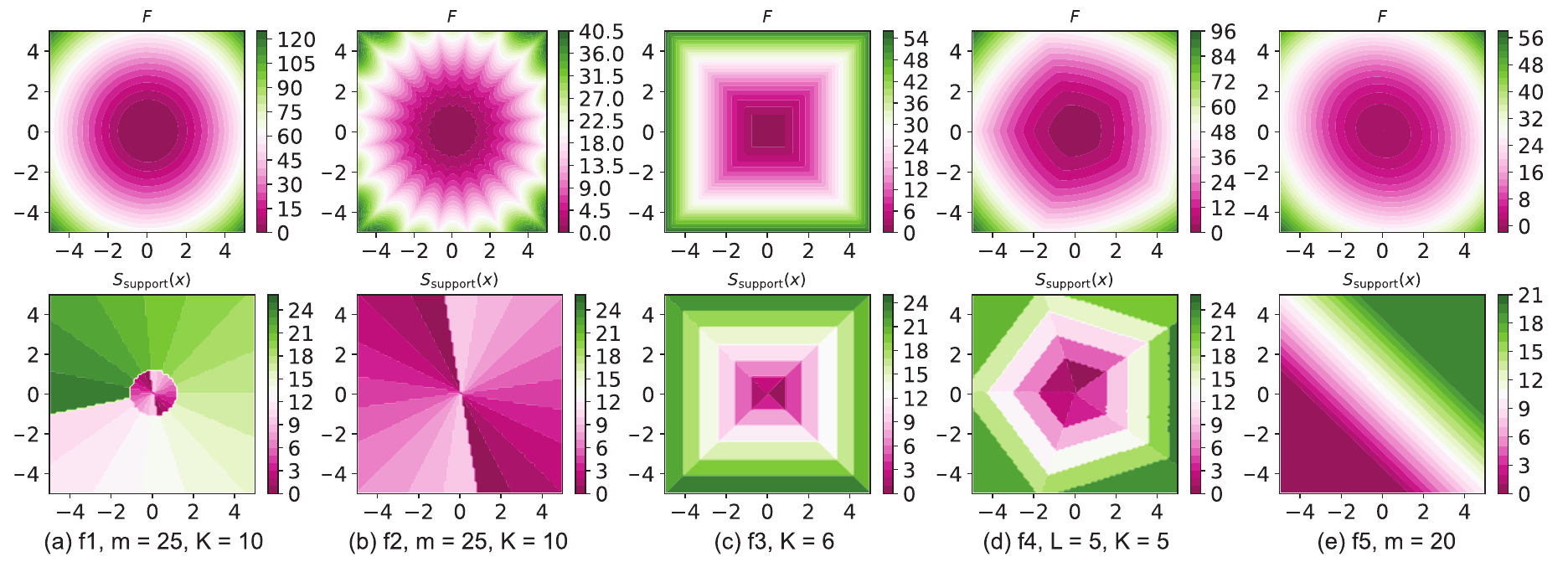}%
  \caption{
Visualization of test problems $f_1$--$f_5$ for dimension $n = 2$. The first row shows the distribution of the ground-truth worst-case objective function $F(x)$. The second row shows the distribution of the worst-case scenario index $\argmax_{s \in S} f(x, s)$ per $x$.}
  \label{fig:testproblem}
\end{figure*}

To test the hypotheses, we construct test problems P1--P5 below.
In these problems, the number $m$ of scenarios and the number $\abs{S_\mathrm{support}(x^*)}$ of the support scenarios around the optimal solution $x^*$ of the worst-case objective $F$ are controllable. 
The 2D landscape of the worst-case functions $F$ and the support scenarios at each $x$ on P1--P5 are shown in \Cref{fig:testproblem}. 
The worst-case function in all problems is a single peak function, but each problem has a different distribution of support scenarios. 
All the test problems have their optimal solutions at $x^* = \bm{0}$ and $F(x^*) = 0$. 
The test problem definitions (P1--P5) are listed as follows.

\begin{enumerate}
\item[P1] For any $n \geq 2$, $m \geq 2$, and $K \geq 2$, 
  \begin{equation*}
    f_1(x, s)
    =
    \begin{cases}
      \norm{x}^2 - \big(1 + \alpha\big)\inner{x}{v_s}^2 & s \leq K \\
      2 \norm{x-v_s}^2 - 8 & s > K
      \enspace,
    \end{cases}
  \end{equation*}
  where
  $v_s = (\cos(\omega\cdot s), \sin(\omega\cdot s), 0, \dots, 0)$,  
  $\alpha = (\tan \omega)^{-2}$, 
  and $\omega = \pi / K$ for $s \leq K$, and %. In addition,  
  $v_s = (\cos(\tilde{\omega}\cdot (s-K)), \sin(\tilde{\omega}\cdot (s-K)), 0, \dots, 0)$, $\tilde{\omega} = 2 \pi / (m-K)$ for $s > K$. 
Thus, we have $S_\mathrm{support}(\R^n) = S$ and $S_\mathrm{support}(x^*) = \{1,\dots,K\}$.

\item[P2] For any $n \geq 2$, $m \geq 2$, and $K \geq 2$, 
  \begin{equation*}
    f_2(x, s)
    =
    \begin{cases}
      \norm{x}^2 - \big(1 + \alpha\big)\inner{x}{v_s}^2 & s \leq K \\
      \norm{x-v_s} - 2 & s > K
      \enspace,
    \end{cases}
  \end{equation*}
  where
  $v_s = (\cos(\omega\cdot s), \sin(\omega\cdot s), 0, \dots, 0)$, $\alpha = (\tan \omega)^{-2}$, and $\omega = \pi / K$ for $s \leq K$, and%. In addition,  
  $v_s = (\cos(\tilde{\omega}\cdot (s-K)), \sin(\tilde{\omega}\cdot (s-K)), 0, \dots, 0)$, and $\tilde{\omega} = 2 \pi / (m-K)$ for $s > K$.
    Thus, we have that $S_\mathrm{support}(\R^n) = S_\mathrm{support}(x^*) = \{1,\dots,K\}$. 

\item[P3] For any $n \geq 1$, $m \geq 2n$, 
  \begin{equation*}
    f_3(x, s)
    =
      \inner{x - \alpha_s \cdot v_s}{v_s}^2 - \beta_s \enspace,
  \end{equation*}
  where $K = \lceil m / 2n \rceil $, $\alpha_s$, $\beta_s$, and $v_s$ for each $s \in S$ are defined as follows. 
  Let $k = \lceil s / (2 \cdot n) \rceil$ and $\ell = s - 2 \cdot n \cdot (k-1)$. 
  Then, $v_s$ is the unit vector whose $\lceil \ell / 2\rceil$-th element is $(-1)^{\ell}$.
  We define $\tilde{\alpha}_k := 5 \cdot k / K$, and $\tilde{\beta}_k := \tilde{\beta}_{k-1} + (\tilde{\alpha}_k + \tilde{\alpha}_{k-1})^2 - (2\tilde{\alpha}_{k-1})^2$ and $\tilde{\beta}_1 := \tilde{\alpha}_1^2$.
  Hence, we have $S_\mathrm{support}(\R^n) = S$ and $S_\mathrm{support}(x^*) = \{1,\dots,2n\}$.

\item[P4] For any $n \geq 1$, $m \geq L \geq 2$, 
  \begin{equation*}
    f_4(x, s)
    =
    \norm{x}^2 + 2 \inner{x}{v_s} - \norm{v_s}^2 + \frac{5}{K} \enspace,
  \end{equation*}
  where $K = m / L$, $v_s = (5 \cdot k / K) \cdot (\cos(\omega \cdot \ell), \sin(\omega \cdot \ell), 0, \dots, 0)$ for $\omega = 2 \pi / L$, $k = \lceil s / L \rceil$, and $\ell = s - L \cdot (k-1)$. 
%  The number of scenarios is $m = L \cdot K$. 
  We have $S_\mathrm{support}(\R^n) = S$, $S_\mathrm{support}(x^*) = \{1,\dots,L\}$.

\item[P5] For any $n \geq 1$, $m \geq 2$, 
 \begin{align*}
%	 & f_5(x, s) = x^{2} + x \omega_{s} - \omega_{s}^{2} + 2n  \\
    f_5(x, s) = x^{2} + x \omega_{s} - \omega_{s}^{2} \enspace. 
 \end{align*}
where $\omega_{s} = \frac{2(s-1)}{m-1} - 1$ for all $s=1,..,m$.
Thus $S_\mathrm{support}(x^*)=\{(m-1)/2 + 1\}$ if $m$ is odd, and $S_\mathrm{support}(x^*) = \{m/2, m/2+1 \}$ if $m$ is even.

\end{enumerate}

\subsection{Common Settings}\label{sec:common}

We optimized P1--P5 using \pbilcma, \pbilcma{} with fixed $\lambda_s = \abs{S_\mathrm{support}(x^*)}$, \texttt{CMA-ES} and \texttt{lq-CMA-ES} for 20 trials with different random seeds. 
% \textcolor{blue}{[R3C6] : 
The search domain was $\X = \R^n$, with $S = \llbracket 1, m\rrbracket$.
% }
We used \texttt{pycma} \cite{pycma} for the implementation of \texttt{lq-CMA-ES}.
We implemented the other approaches using the version of the CMA-ES proposed in \cite{akimoto2019} as the baseline\footnote{\url{https://gist.github.com/youheiakimoto/1180b67b5a0b1265c204cba991fa8518}}. For a fair comparison between \texttt{lq-CMA-ES} and the other approaches, we turned off the diagonal acceleration mechanism of \cite{akimoto2019}. All hyperparameters were set to their default values. The initial mean and covariance matrix of the CMA-ES was $m^0 \sim \mathcal{U}(-4, 4)^n$ and $\Sigma^0$ to $2^2 \cdot I_n$. 
% \textcolor{blue}{[R3C7]: 
We used the same initial mean vector and covariance matrix for \texttt{lq-CMA-ES}. 
The other hyperparameters for \texttt{lq-CMA-ES} were set to their default values implemented in \texttt{pycma}.
% }
In these experiments, we did not perform a restart because the test problems were all single-peak problems. The hyperparameters for \pbilcma{} were set as follows: $c_p = 0.3$, $\eta = 0.3$, $\epsilon = 1/m$, $\gamma = 0.99$, and $p_s^0 = 0.1$ for all $s \in S$. Their sensitivities are analyzed in \ref{app:sensitivity}. 
For \pbilcma{} with fixed $\lambda_s$, the hyperparameters were set as follows: $c_p = 0.1$, $\epsilon = 1/m$, $\gamma=0.99$, and $p^0_s = \lambda_s/m$.

The termination criteria were as follows. 
We regarded a run as successful if $\abs{F(m^t) - F(x^*)} < 10^{-12}$ was reached before $10^6$ $f$-calls were spent.
If $10^6$ $f$-calls were spent before reaching $\abs{F(m^t) - F(x^*)} < 10^{-12}$, we regarded a run as a failure.
Additionally, we implemented the following conditions: too small a search distribution\footnote{The covariance matrix in the CMA-ES is usually split as $\Sigma = \sigma^2 \cdot \CC$, and they are updated separately.} $\sigma^t < 10^{-12}$, and an excessively large condition number\footnote{We observed that $\Cond(\Sigma^t)$ reached $10^{14}$ when we applied \texttt{lq-CMA-ES} on P4 and optimization was interrupted although $F(x)$ was improved. In \texttt{pycma}, a functionality that avoids this situation has been implemented (\texttt{alleviate-conditioning-in-coordinates}), and we used this function for \texttt{lq-CMA-ES} on P4.} $\Cond(\Sigma^t) > 10^{14}$.
If one of them was reached, we regarded the run as a failure.

\providecommand{\figsizeexone}{0.2}
\begin{figure}[t]
  \centering
  \begin{subfigure}{\figsizeexone\hsize}%
    \includegraphics[width=\hsize]{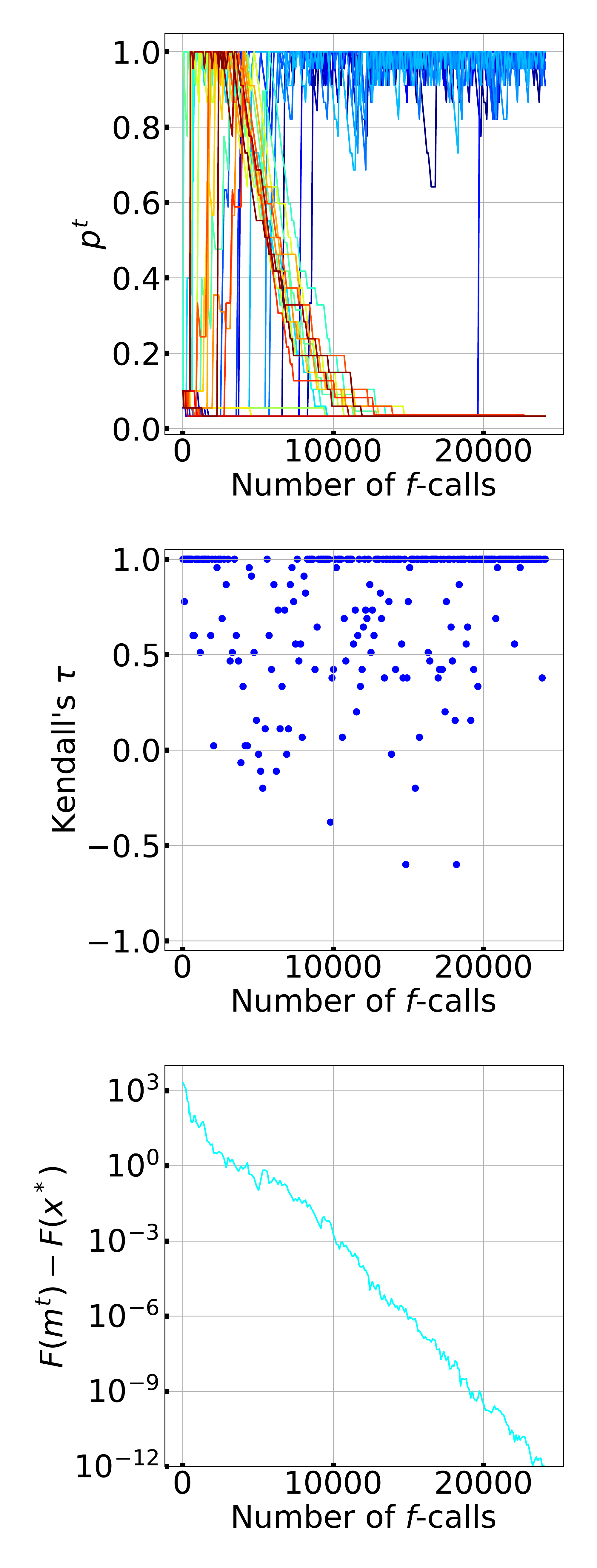}%
    \caption{P1}%
  \end{subfigure}%
  \begin{subfigure}{\figsizeexone\hsize}%
    \includegraphics[width=\hsize]{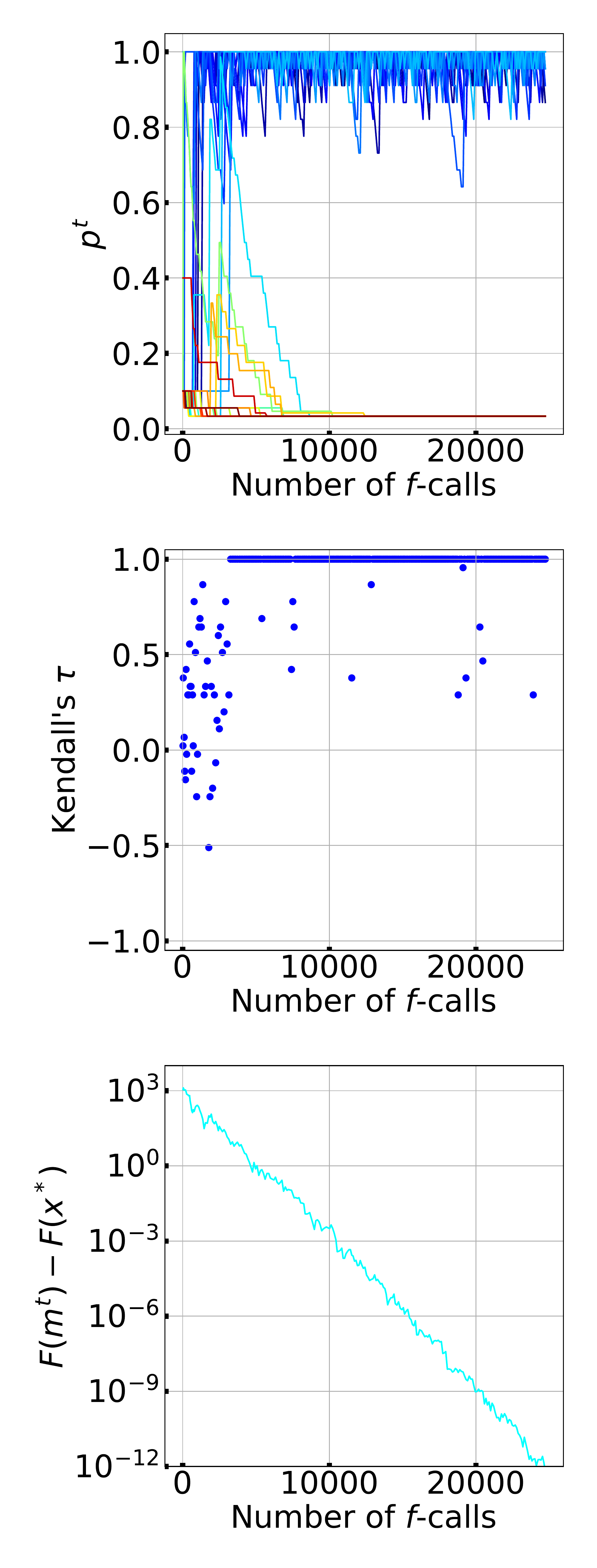}%
    \caption{P2}%   
  \end{subfigure}%
  \begin{subfigure}{\figsizeexone\hsize}%
    \includegraphics[width=\hsize]{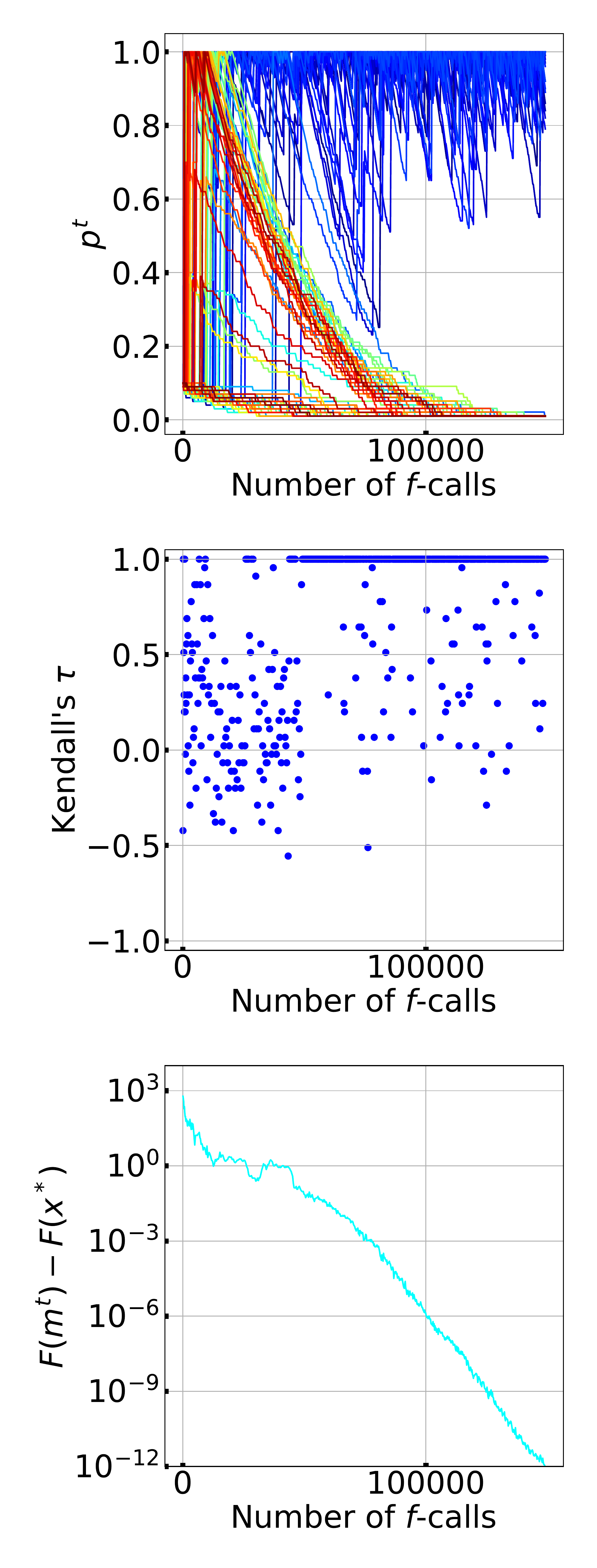}%
    \caption{P3}%   
  \end{subfigure}%
  %\\
    \begin{subfigure}{\figsizeexone\hsize}%
    \includegraphics[width=\hsize]{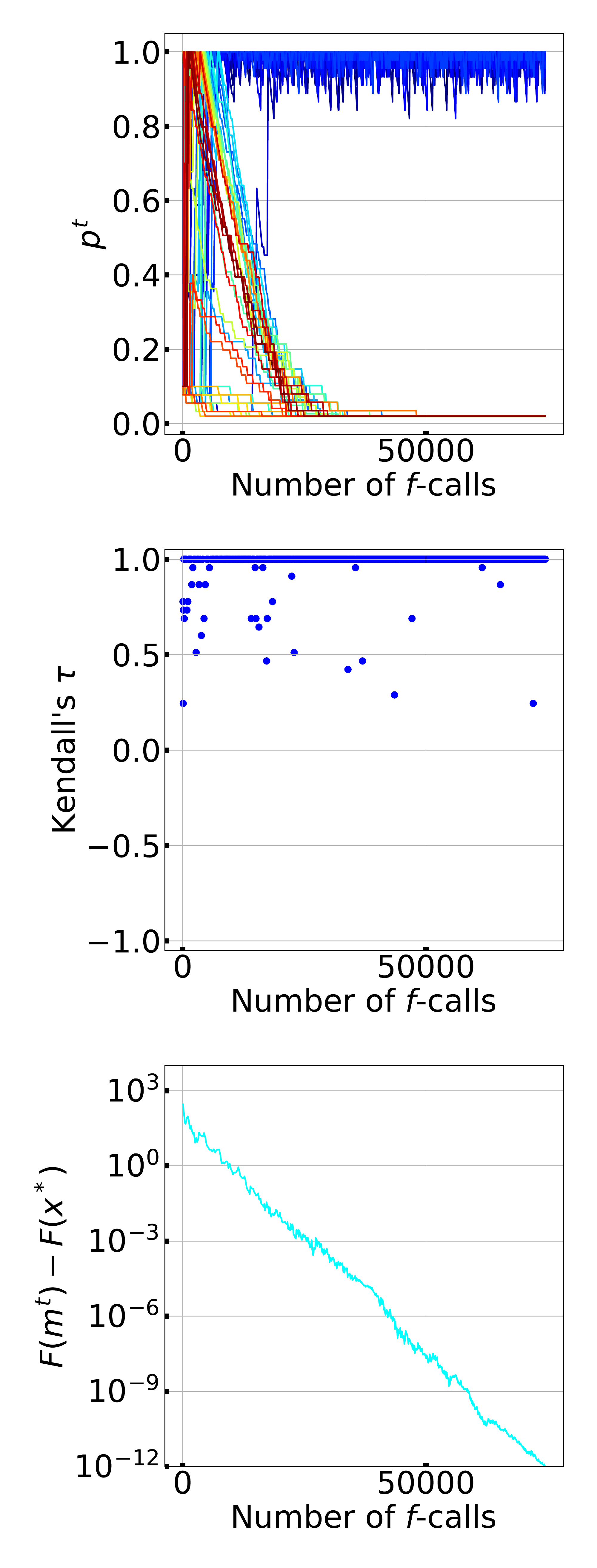}%
    \caption{P4}%   
  \end{subfigure}%
    \begin{subfigure}{\figsizeexone\hsize}%
    \includegraphics[width=\hsize]{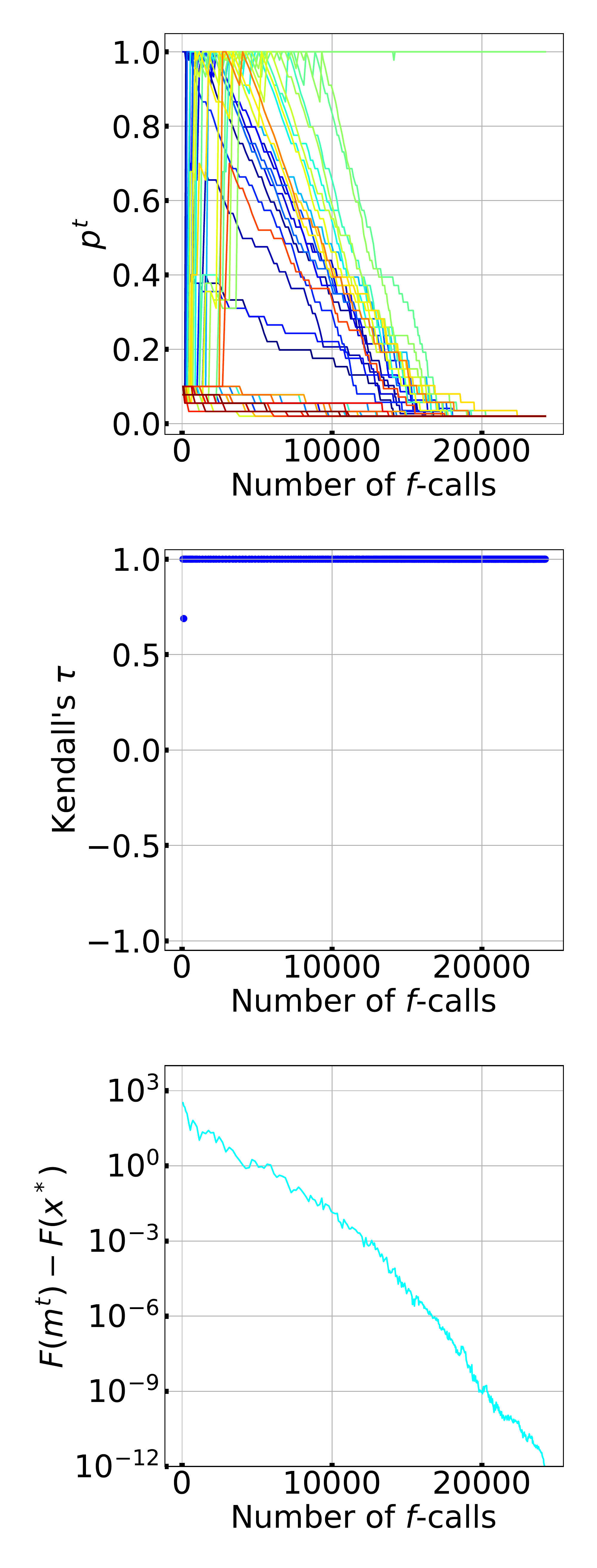}%
    \caption{P5}%   
  \end{subfigure}%
  \caption{History of probability $p^t_s$, Kendall's $\tau$, and the gap $F(m^t) - F(x^*)$ in a typical run on P1--P5. Curves for higher scenario indices are depicted in warmer colors.}
\label{fig:p_dynamics}
\end{figure}

\subsection{Adaptive Behavior of $p_s^t$ in \pbilcma{} (Hypothesis (1))}\label{sec:ex1}

To test Hypothesis (1), we applied \pbilcma{} to P1--P5 with $n = 10$.
The problem control parameters were set as follows: $m = 30$ and $K = 10$ for P1 and P2, $K = 5$ (hence, $m = 100$) for P3, $L = 10$ and $K = 5$ (hence, $m = 50$) for P4 and $m=50$ (hence, $\abs{S_\mathrm{support}(x^*)} = 2$) for P5. 

\Cref{fig:p_dynamics} shows the history of $p^t_s$ for each $s \in S$, the Kendall rank correlation coefficient $\tau$ between $F(x^t_1), \dots, F(x^t_{\lambda_x})$ and $F(x^t_1; A^t), \dots, F(x^t_{\lambda_x}; A^t)$, and 
% \textcolor{blue}{[R3C11]:
the gap $F(m^t) - F(x^*)$ 
% }
in a typical optimization run for each problem. 
All the runs were successfully terminated by reaching the target threshold for $F$. 
The results for all cases show that $p^t_s$ tended to $\ind{s \in S_\mathrm{support}(x^*)}$ at the end of the run, and Kendall's $\tau$ remained at one at a high frequency. 
P2 has the property that only $K$ scenarios can be support scenarios over the entire domain. 
Therefore, we see from \Cref{fig:p_dynamics}(b) that \pbilcma{} mistakenly increased $p_s^t$ for a few scenarios that are not in $S_\mathrm{support}(H_\gamma^t)$ at the beginning. 
However, their $p_s^t$ values started to decrease after a few iterations.
P1 has the property that it is identical to P2 around $x^*$, but outside the neighborhood of $x^*$, the support scenarios are $S \setminus S_\mathrm{support}(x^*)$. 
In \Cref{fig:p_dynamics}(a), it can be observed that $p_s^t$ were increased for $s \notin S_\mathrm{support}(x^*)$ initially, and $p_s^t$ subsequently began to decrease for $s \notin S_\mathrm{support}(x^*)$, whereas $p_s^t$ started to increase for $s \in S_\mathrm{support}(x^*)$, where we observed relatively low $\tau$ values. 
% \textcolor{blue}{[R3C11]:
When iterations of low $\tau$ values continued, the reduction rate of the gap $F(m^t) - F(x^*)$ decreased. 
% }
Similar behaviors were observed on P3--P5, where the support scenarios $S_\mathrm{support}(H_\gamma^t)$ change gradually as $H_\gamma^t$. 
From these results, we confirm that $p_s^t$ follows the change of $S_\mathrm{support}(H_\gamma^t)$. 

\providecommand{\figsizefuncr}{0.32}
\begin{figure}[t]
  \centering
  \begin{subfigure}{\figsizefuncr\hsize}%
    \includegraphics[trim=20 20 20 20,clip, width=\hsize]{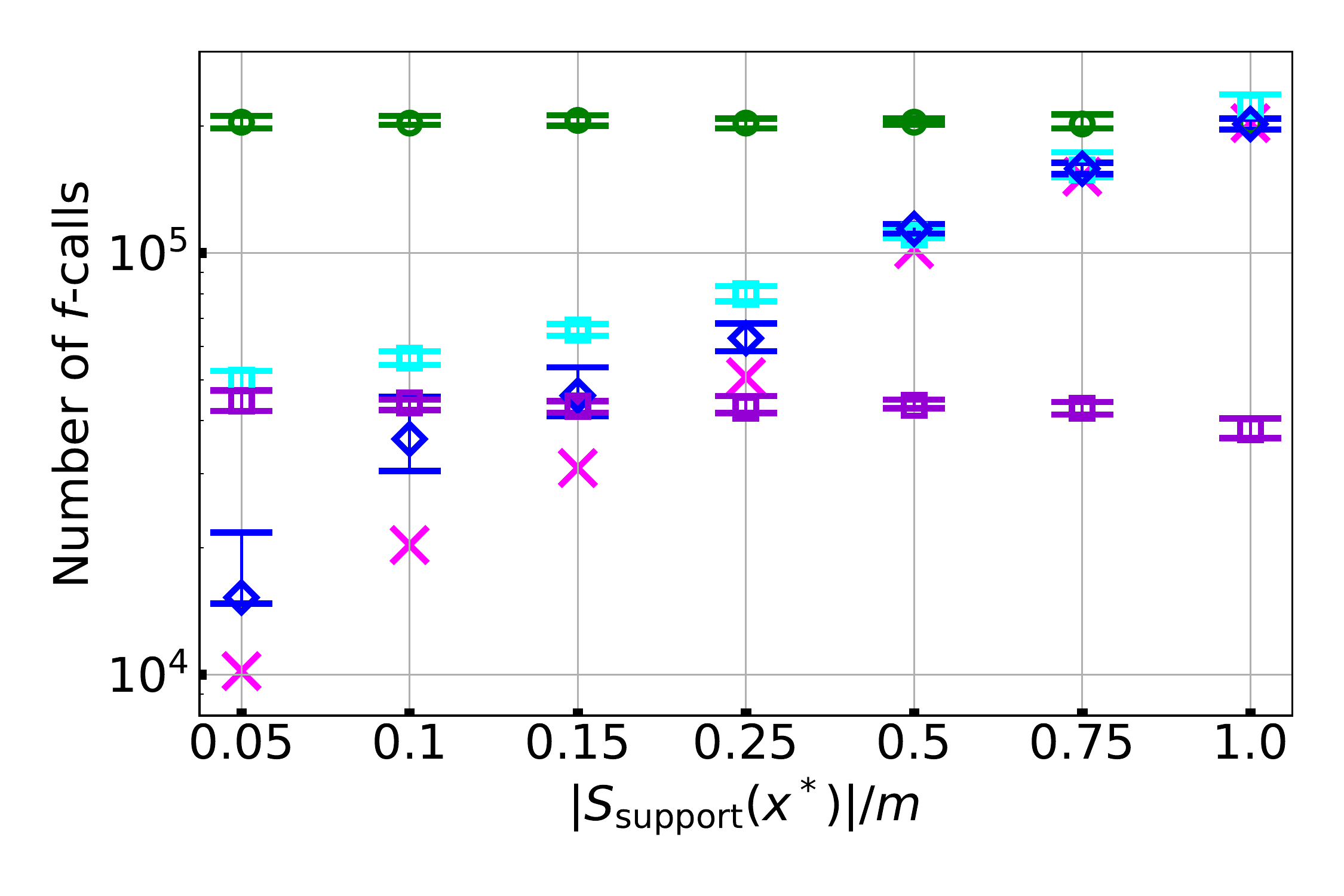}%
    \caption{P1}%    
  \end{subfigure}%
  \begin{subfigure}{\figsizefuncr\hsize}%
    \includegraphics[trim=20 20 20 20,clip, width=\hsize]{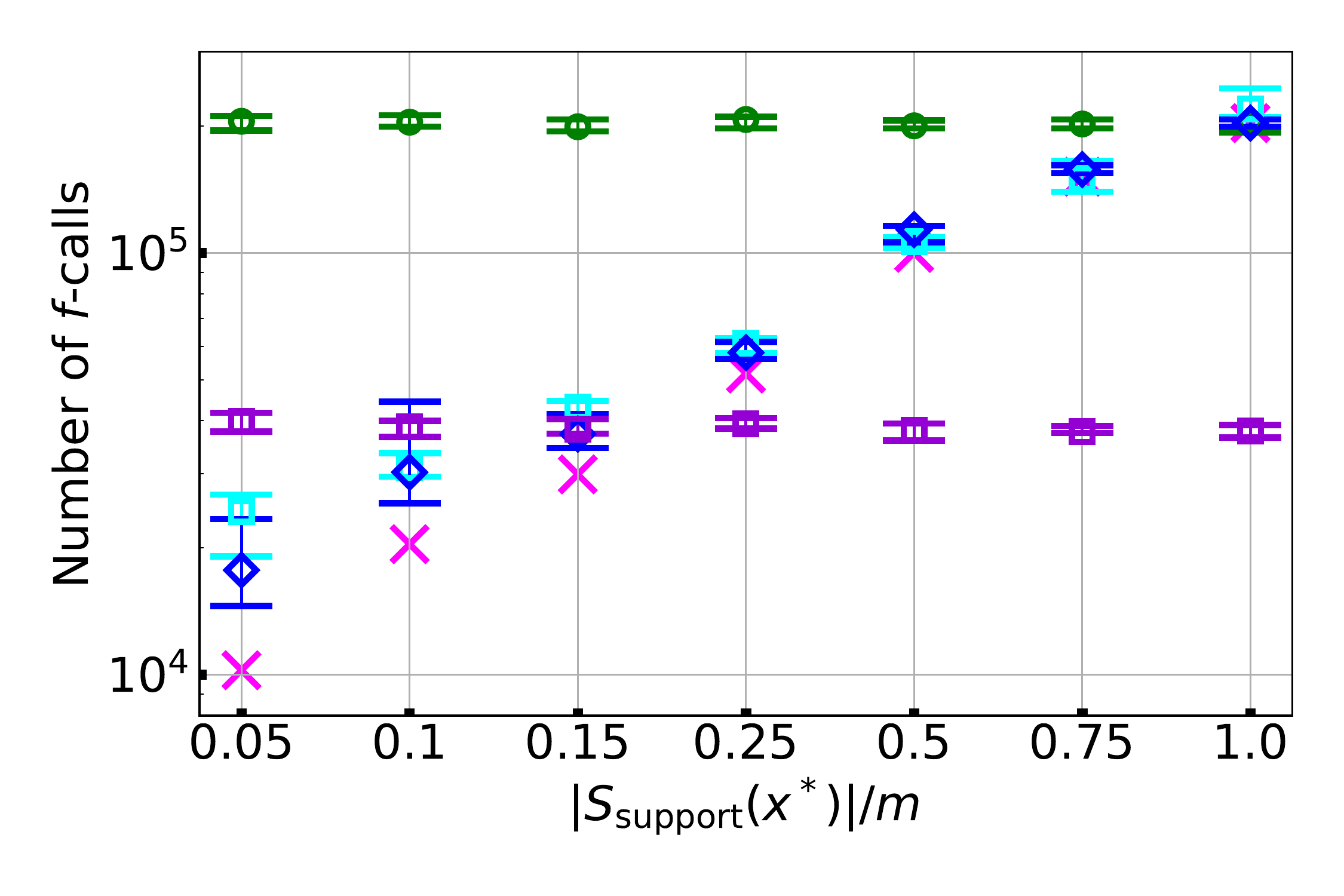}%
    \caption{P2}%    
  \end{subfigure}%
    \begin{subfigure}{\figsizefuncr\hsize}%
    \includegraphics[trim=20 20 20 20,clip, width=\hsize]{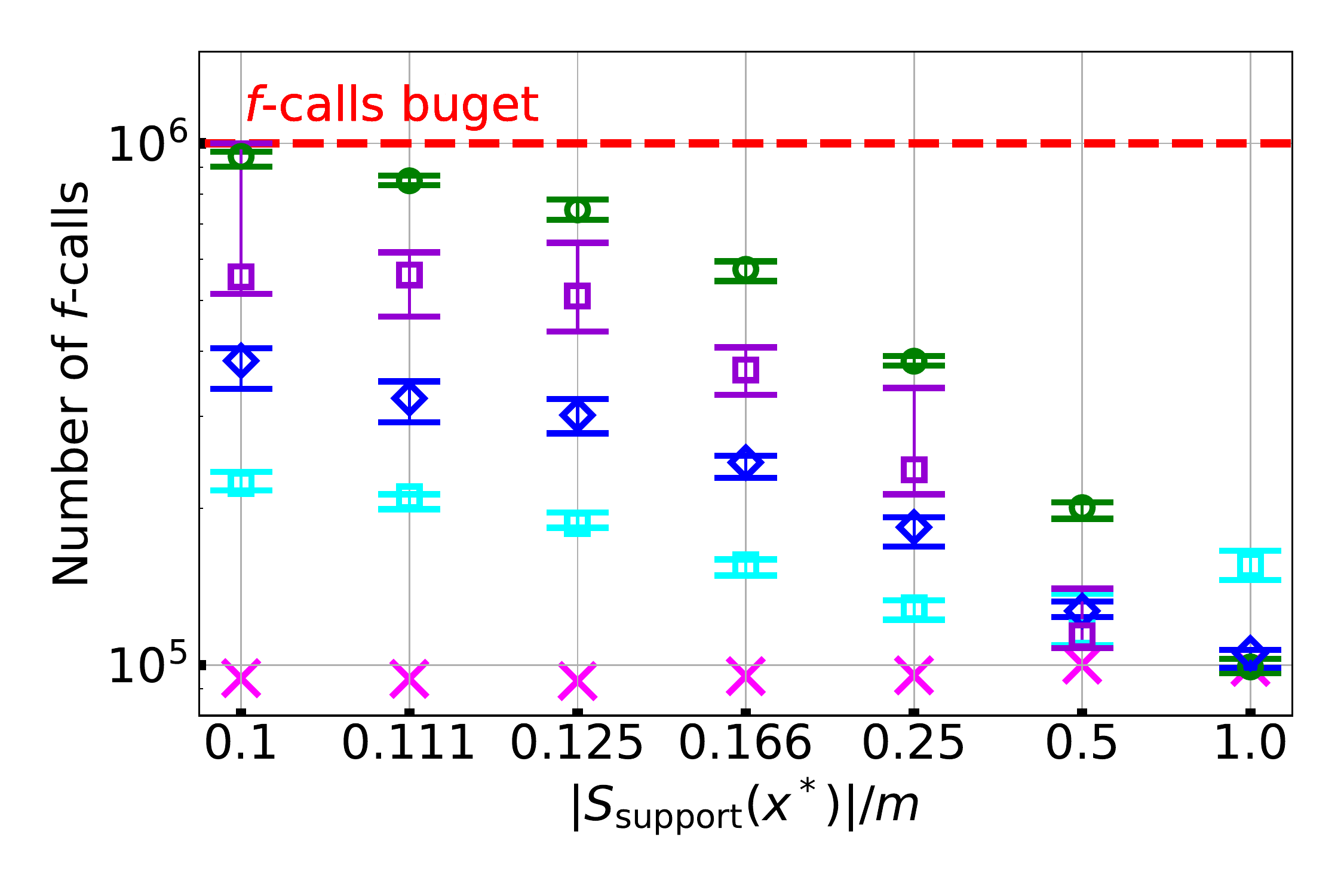}%
    \caption{P3}%    
  \end{subfigure}%
  \\%
% \begin{minipage}{0.64\textwidth}
  \begin{subfigure}{\figsizefuncr\hsize}%
    \includegraphics[trim=20 20 20 20,clip, width=\hsize]{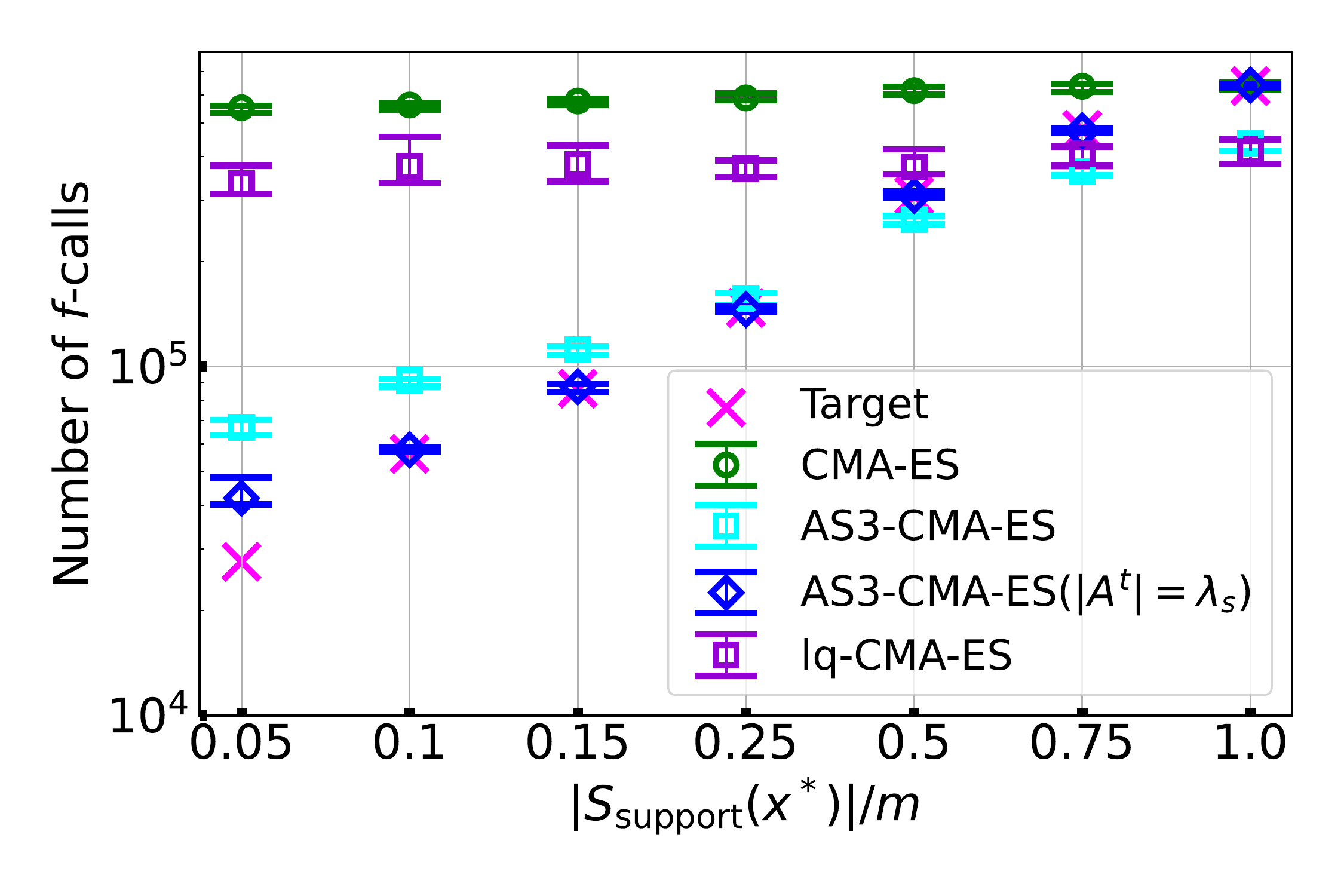}%
    \caption{P4}%    
  \end{subfigure}%  
  \begin{subfigure}{\figsizefuncr\hsize}%
    \includegraphics[trim=20 20 20 20,clip, width=\hsize]{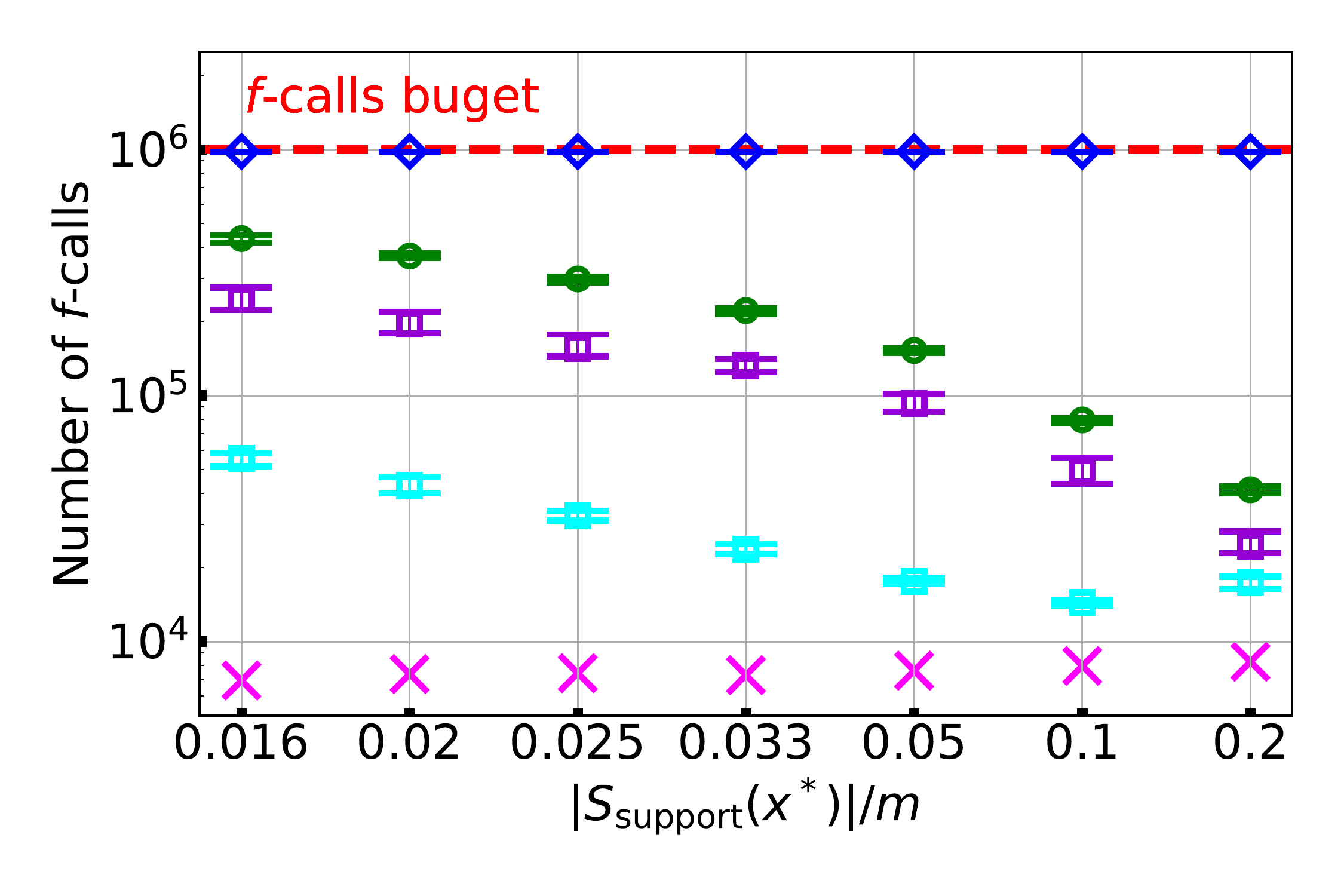}%
    \caption{P5}%    
  \end{subfigure}%    
  \caption{Comparison among \texttt{CMA-ES}, \pbilcma{}, \pbilcma{} with fixed $\lambda_s = \abs{S_\mathrm{support}(x^*)}$ (denoted as \pbilcma{} ($\abs{A^t}=\lambda_s$) at the legend) and \texttt{lq-CMA-ES} on problems P1--P5. Mean and standard deviation of the number of $f$-calls required for convergence over 20 trials. The target is the number of $f$-calls which multiplies the average number of $f$-calls from \texttt{CMA-ES} and the factor $\abs{S_\mathrm{support}(x^*)} / m$. The ratio $\abs{S_\mathrm{support}(x^*)} / m$ is $K/m$ for problems P1 and P2, $1/K$ for problems P3 and P4, and $2/m$ for P5. Cases with $f$-calls reaching $10^6$ were evaluated as optimization failures in this experiment.
%   \textcolor{blue}{[R3C9] : 
  \texttt{CMA-ES} failed at 2 trials on P3 with $\abs{S_\mathrm{support}(x^*)} / m = 0.1$. \texttt{lq-CMA-ES} failed at 8, 4, and 4 trials, respectively, on P3 with $\abs{S_\mathrm{support}(x^*)} / m = 0.1, 0.111, 0.125$. \pbilcma{} with fixed $\lambda_s = \abs{S_\mathrm{support}(x^*)}$ failed at all trials on P5. The other trials were successful.
%   }
  }
  \label{fig:ssensitivity}
% \end{minipage}
% \quad
\end{figure}
% \\

\begin{figure}[t]
\centering
\begin{subfigure}{\figsizefuncr\hsize}%
    \includegraphics[trim=20 20 20 20,clip, width=\hsize]{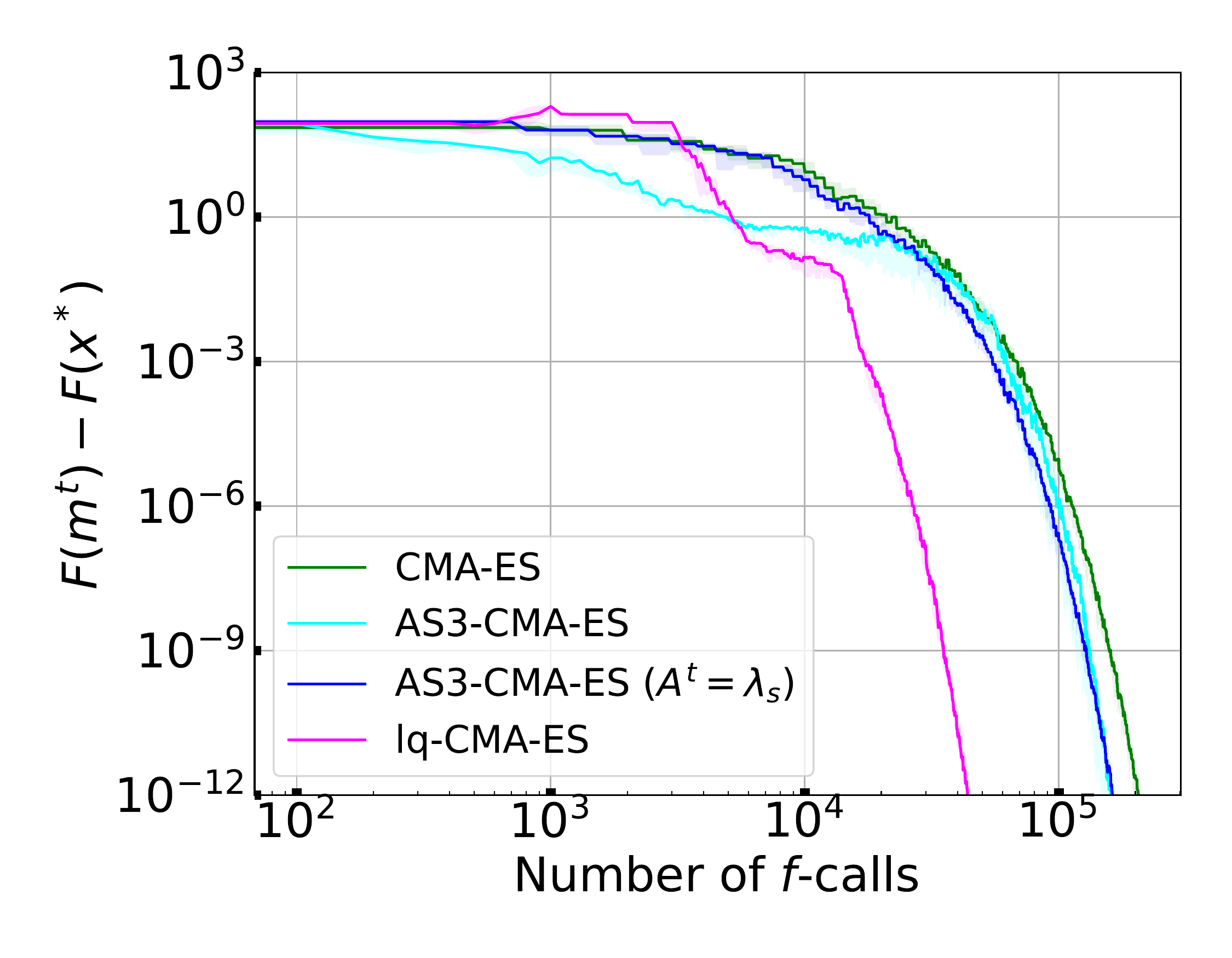}%
    \caption{$\abs{S_{\mathrm{support}}(x^*)}/m=0.75$ on P1}
\end{subfigure}
\quad
\begin{subfigure}{\figsizefuncr\hsize}%
    \includegraphics[trim=20 20 20 20,clip, width=\hsize]{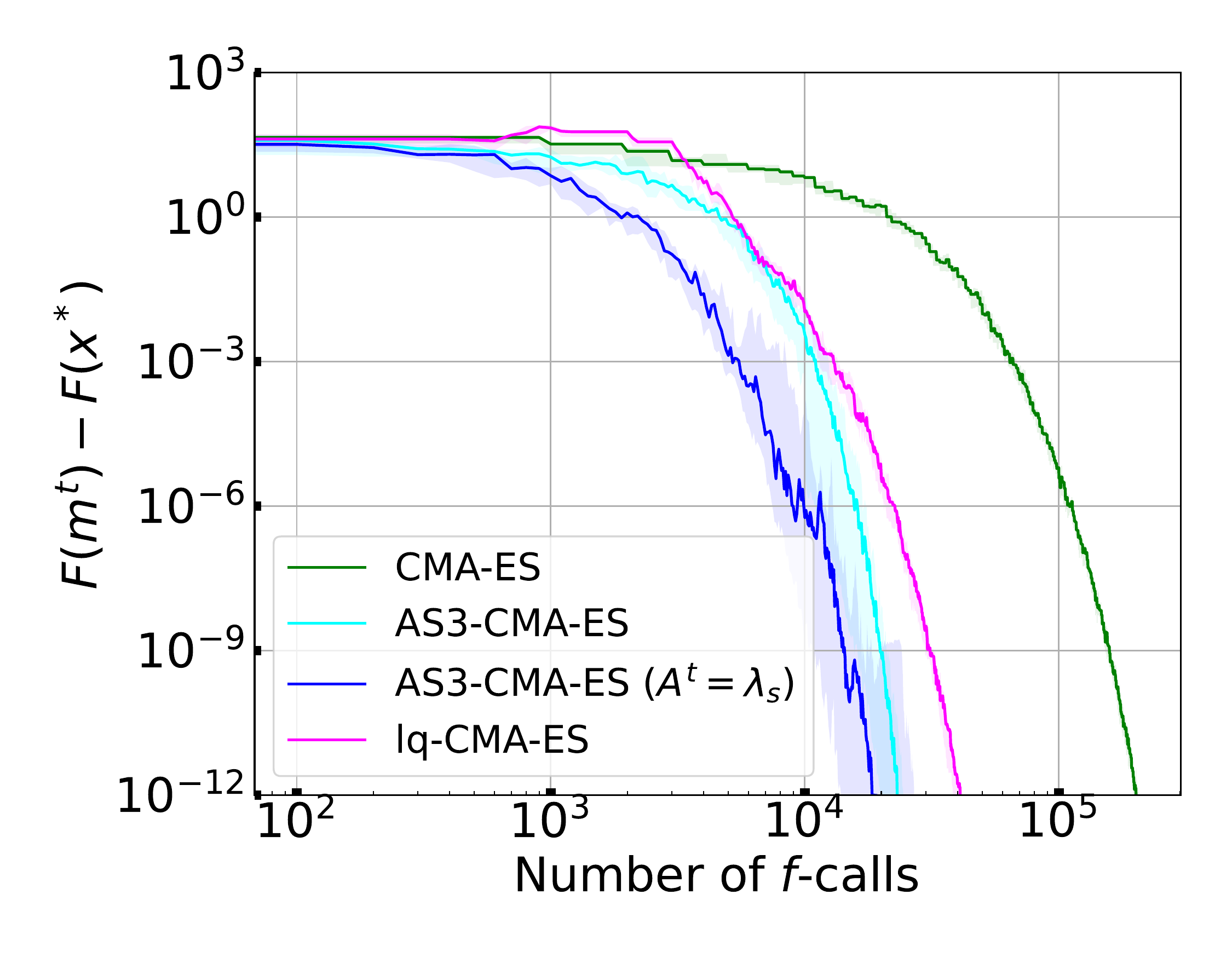}%
    \caption{$\abs{S_{\mathrm{support}}(x^*)}/m=0.05$ on P2}
\end{subfigure}
  \caption{
%   \textcolor{blue}{[R3C11] :
Gap $F(m^t)-F(x^*)$ of each approach. Solid line: median (50 percentile) over 20 runs. Shaded area: interquartile range (25–75 percentile) over 20 runs.
%   }
  }
  \label{fig:gap}

\end{figure}

\begin{figure}[t]
\centering
  \begin{subfigure}{\figsizefuncr\hsize}%
    \includegraphics[width=\hsize]{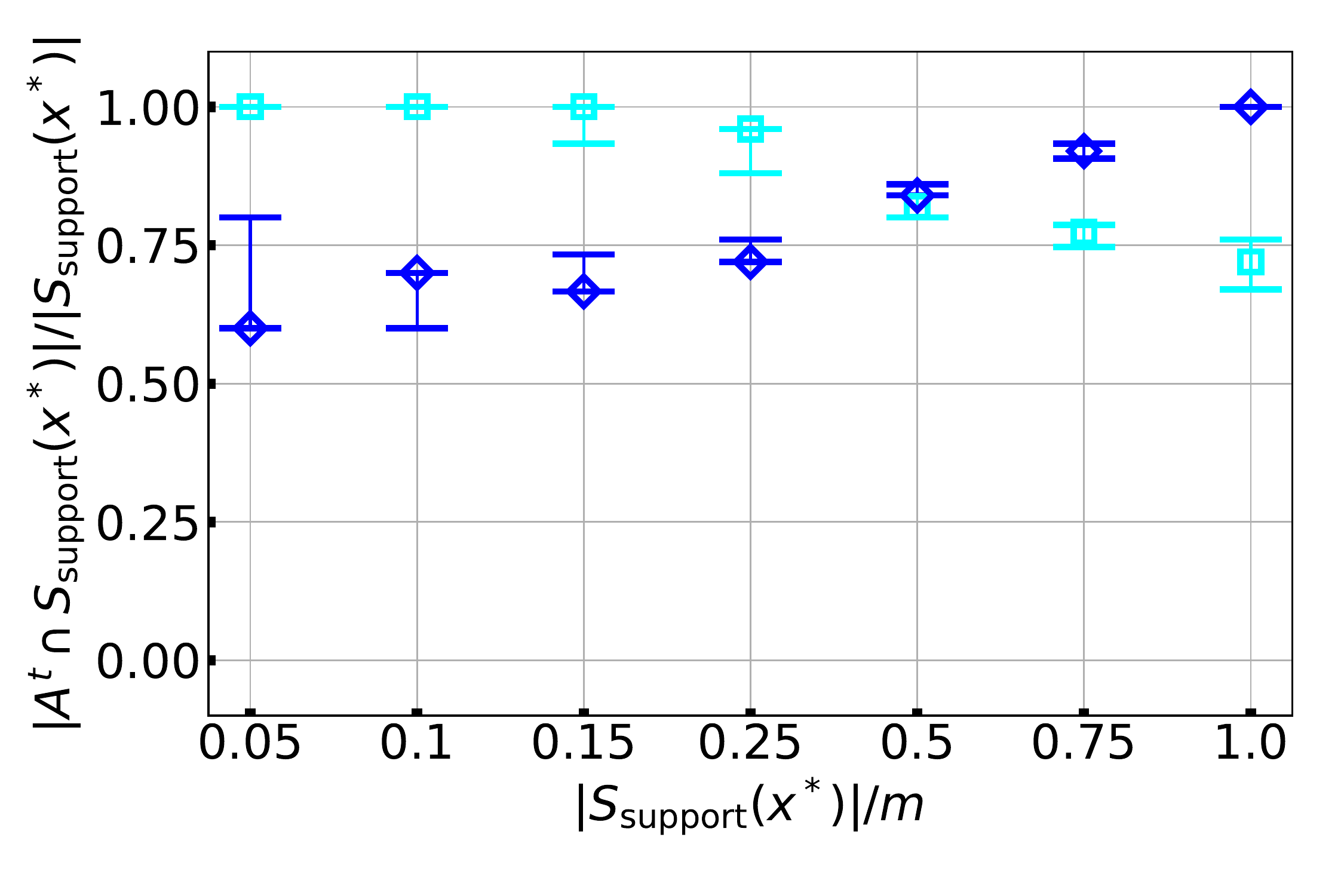}%
    \caption{P1}%    
  \end{subfigure}%
  \begin{subfigure}{\figsizefuncr\hsize}%
    \includegraphics[width=\hsize]{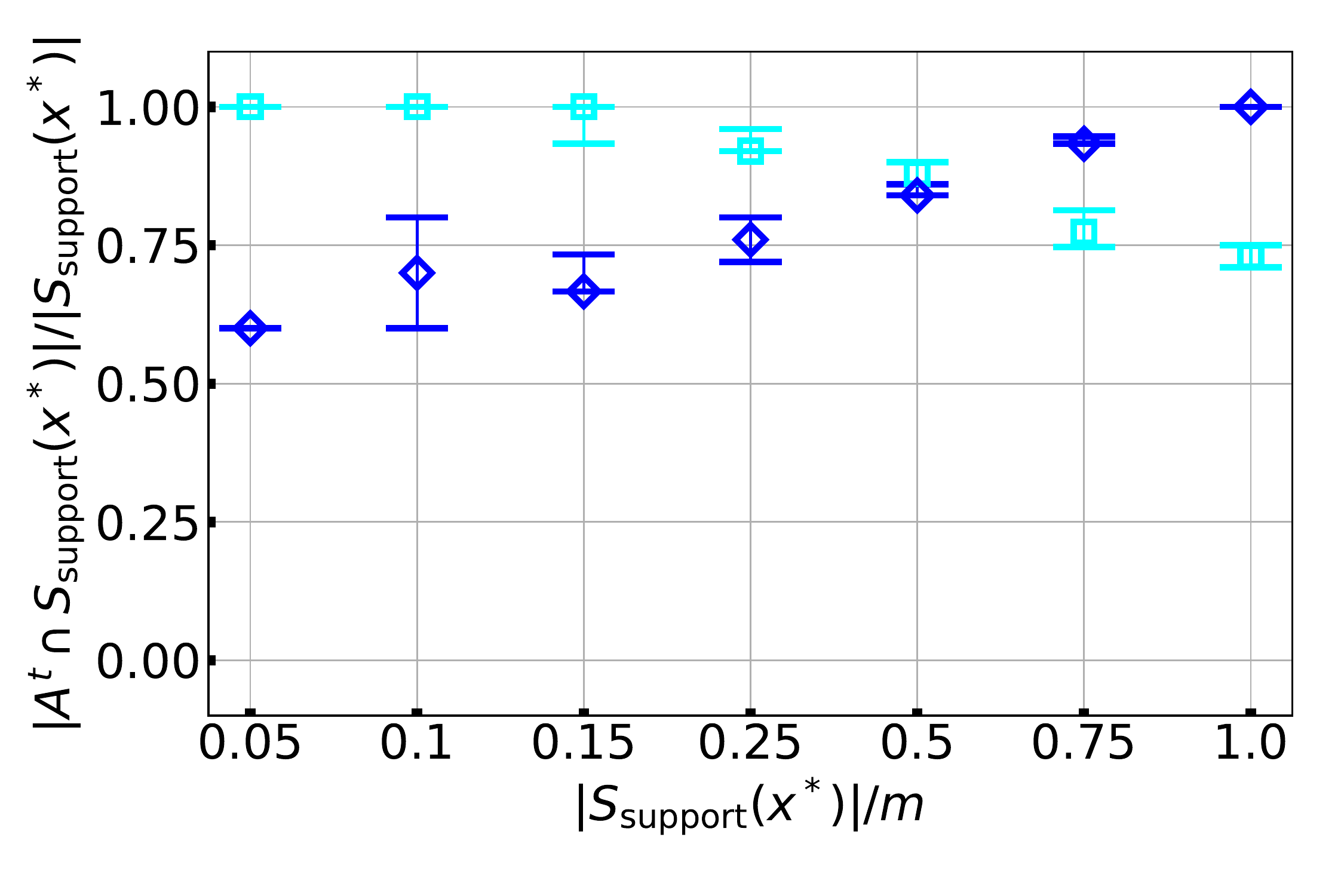}%
    \caption{P2}%    
  \end{subfigure}%
    \begin{subfigure}{\figsizefuncr\hsize}%
    \includegraphics[width=\hsize]{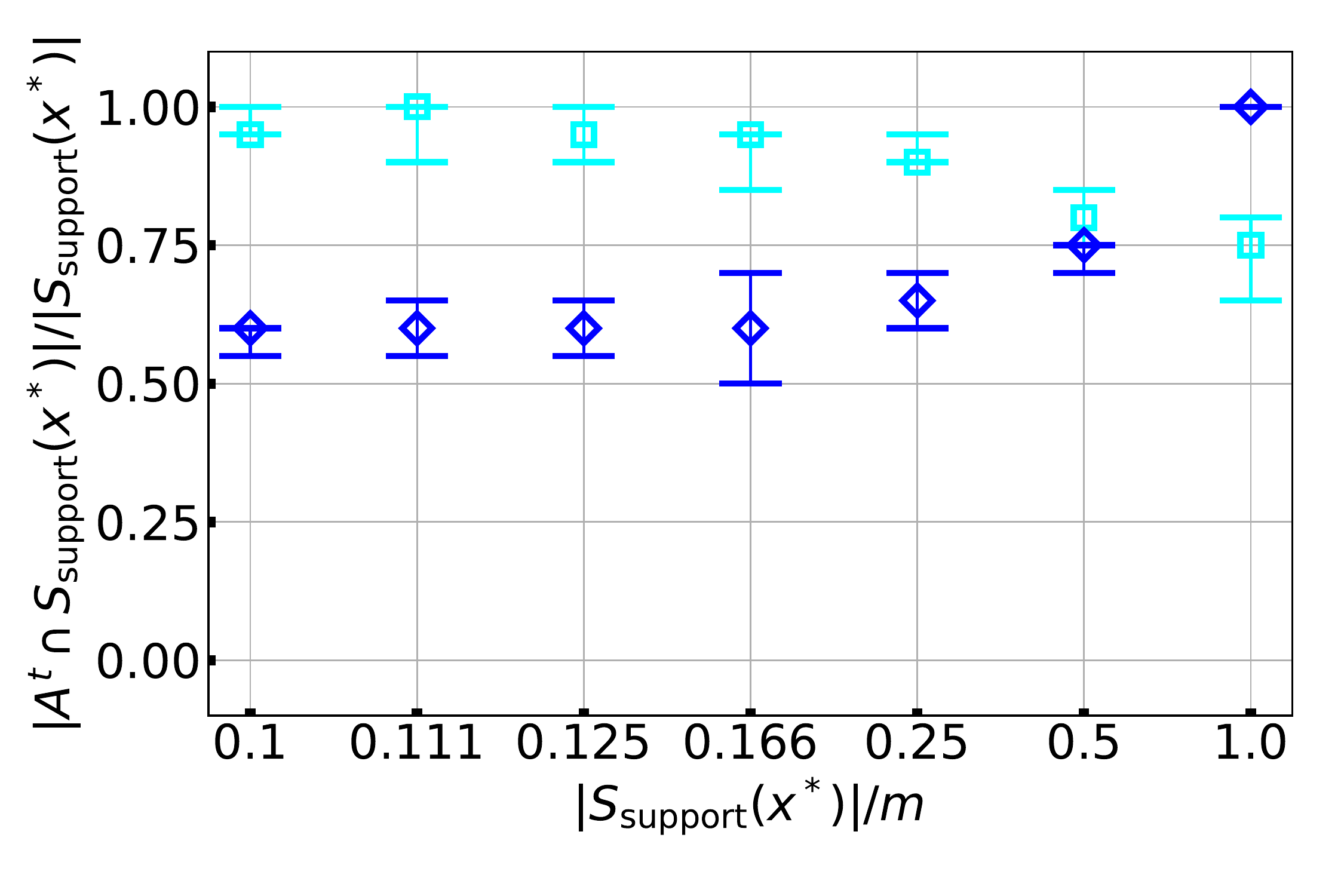}%
    \caption{P3}%    
  \end{subfigure}%
  \\%
  \begin{subfigure}{\figsizefuncr\hsize}%
    \includegraphics[width=\hsize]{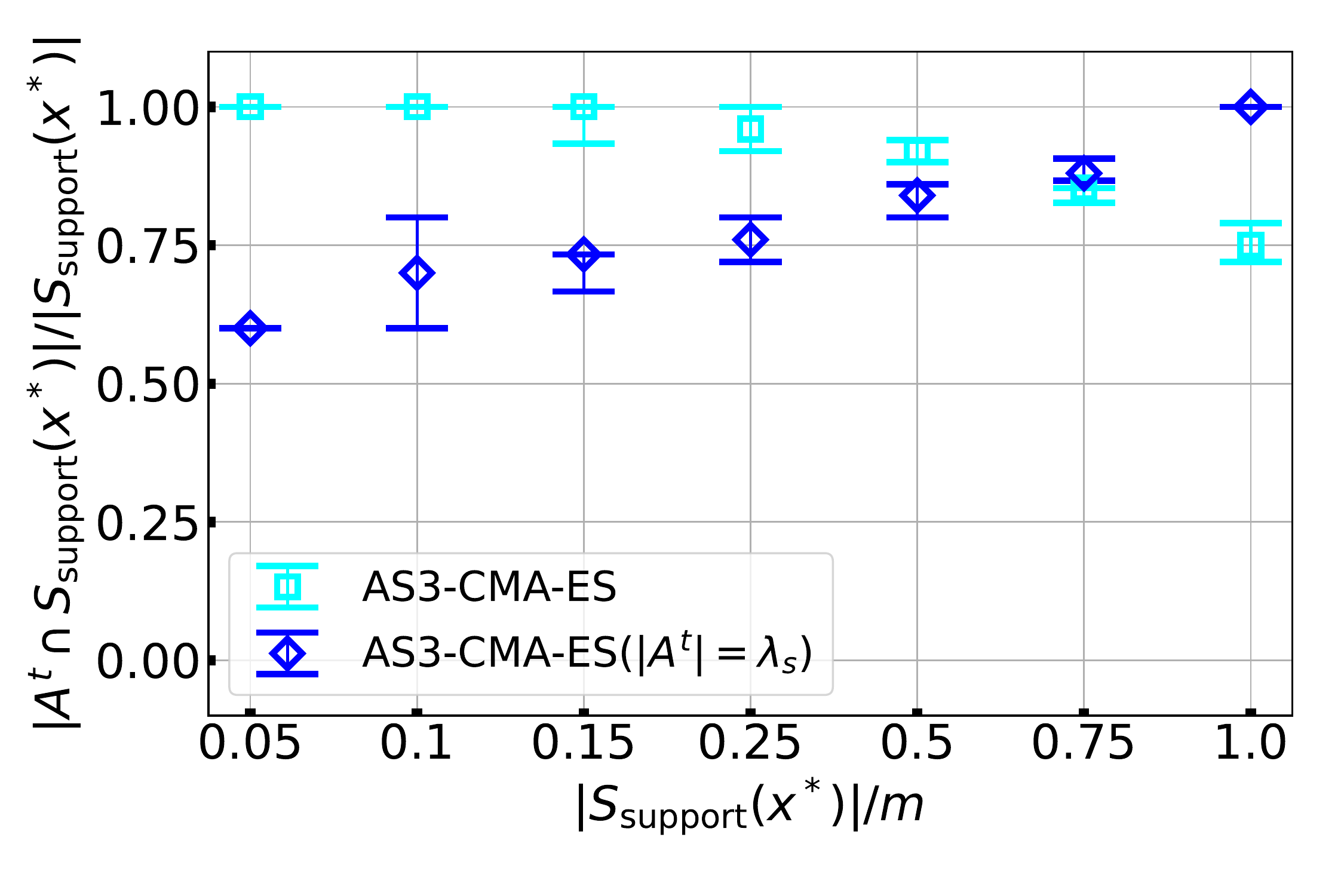}%
    \caption{P4}%    
  \end{subfigure}%  
  \begin{subfigure}{\figsizefuncr\hsize}%
    \includegraphics[width=\hsize]{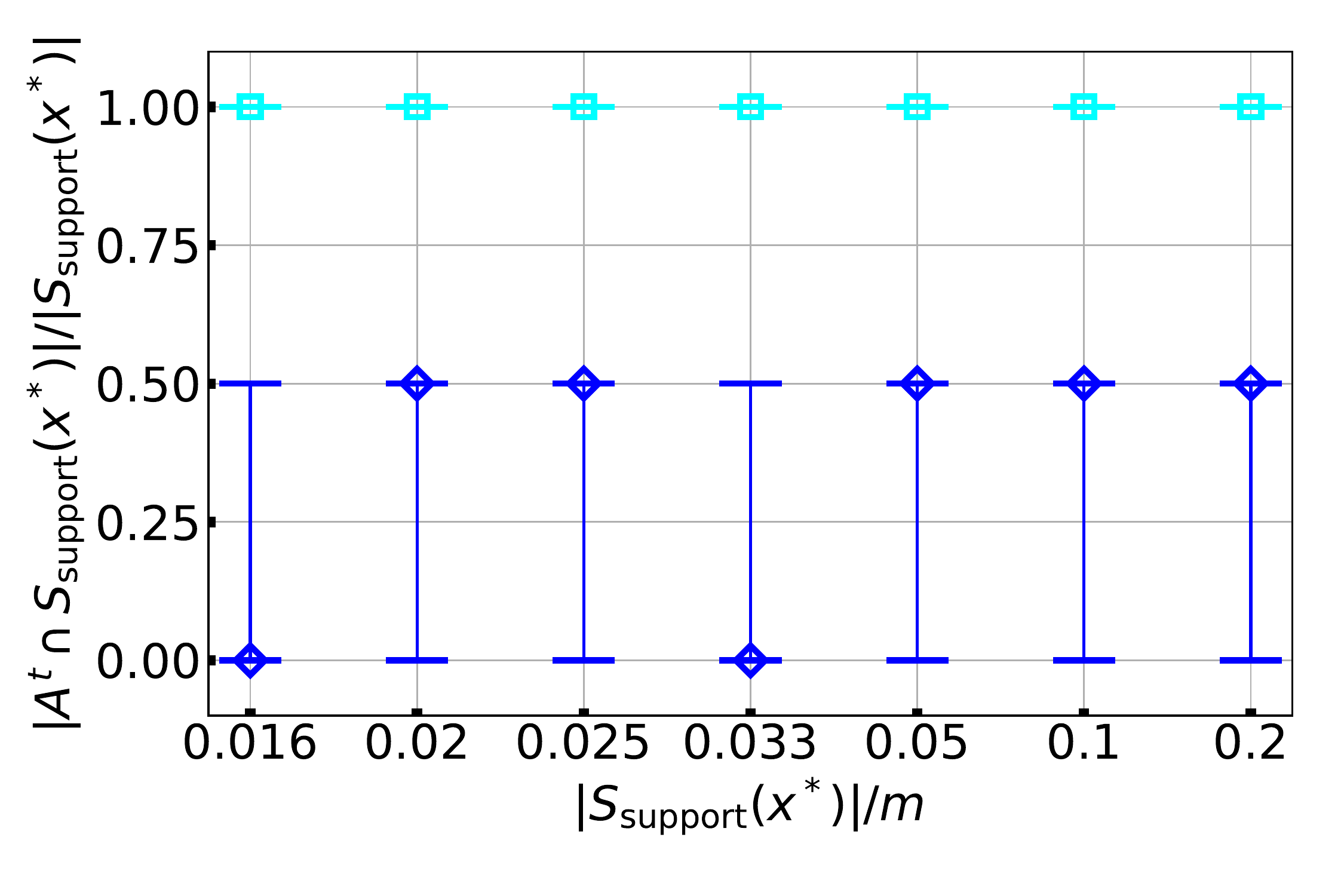}%
    \caption{P5}%    
  \end{subfigure}%    
  \caption{
%   \textcolor{blue}{[R3C10] : 
  Mean and standard deviation of the ratio $\abs{A^t \cap S_\mathrm{support}(x^*)} / \abs{S_\mathrm{support}(x^*)}$ at the end of optimization resulted from \pbilcma{} and \pbilcma{} with fixed $\lambda_s = \abs{S_\mathrm{support}(x^*)}$ (denoted as \pbilcma{} ($\abs{A^t}=\lambda_s$) at the legend) on problems P1--P5 over 20 trials.
%   }
  }
  \label{fig:lastscenarioratio}
%\end{minipage}
\end{figure}

\begin{figure}[t]
\centering
  \begin{subfigure}{\figsizefuncr\hsize}%
    \includegraphics[width=\hsize]{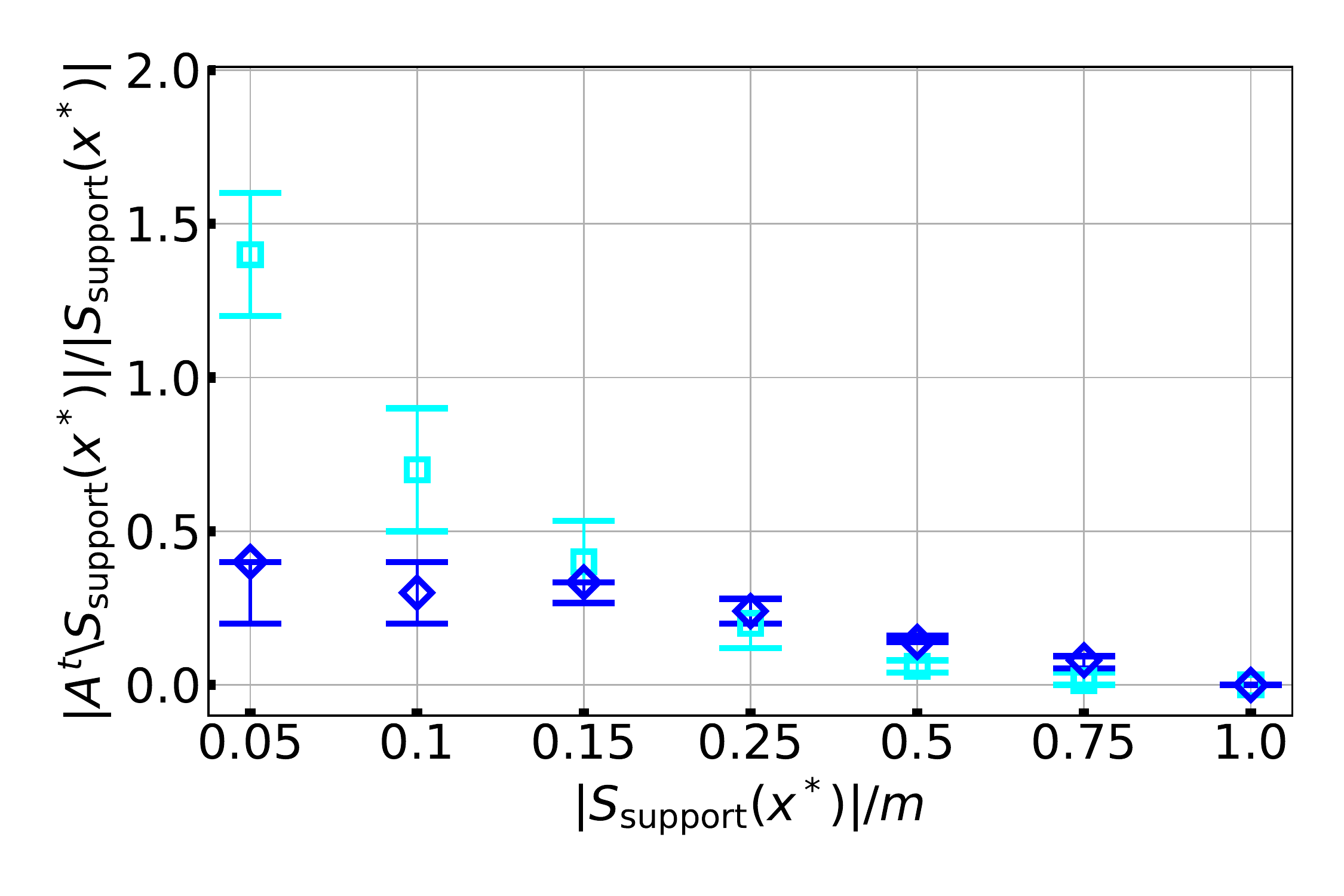}%
    \caption{P1}%    
  \end{subfigure}%
  \begin{subfigure}{\figsizefuncr\hsize}%
    \includegraphics[width=\hsize]{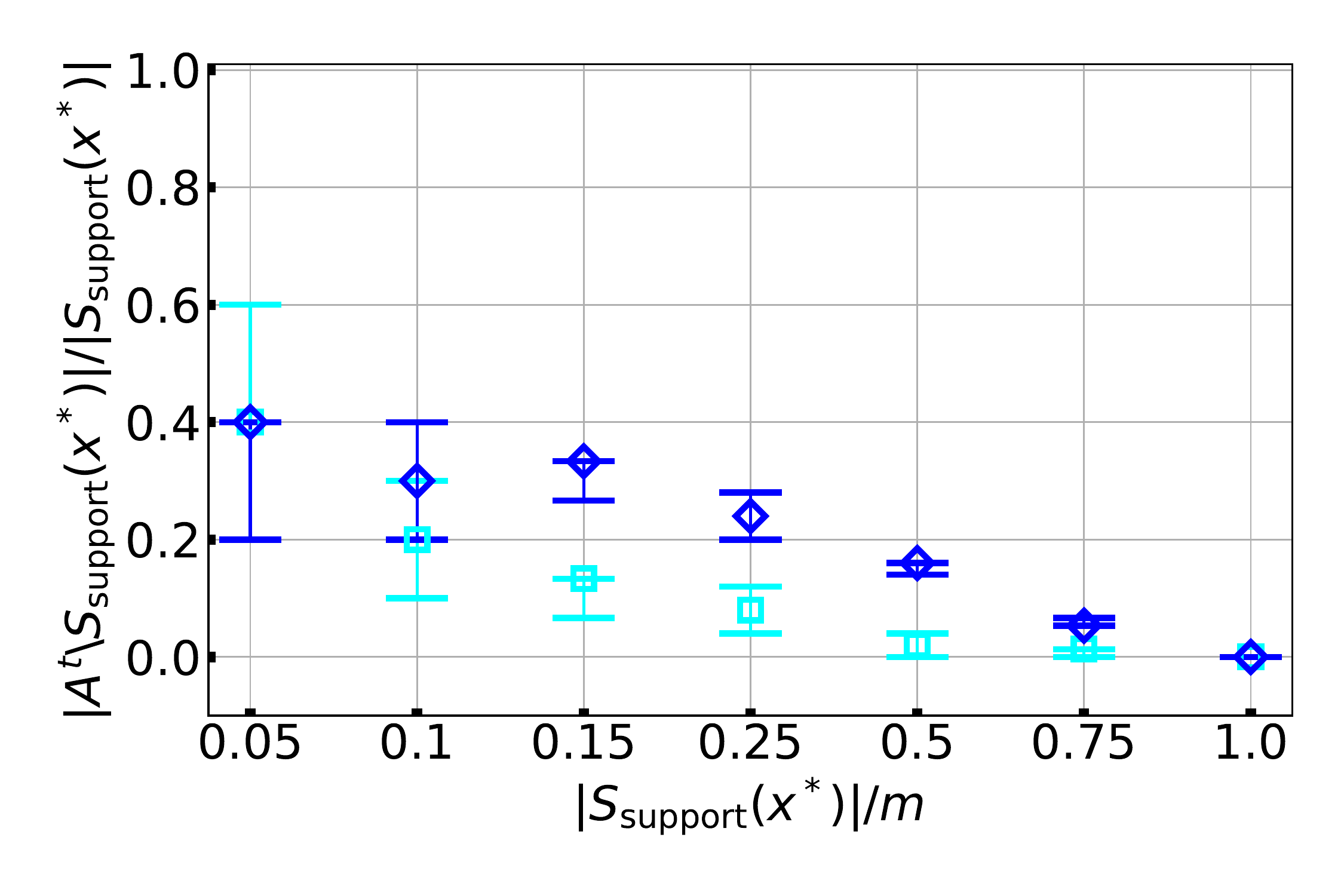}%
    \caption{P2}%    
  \end{subfigure}%
    \begin{subfigure}{\figsizefuncr\hsize}%
    \includegraphics[width=\hsize]{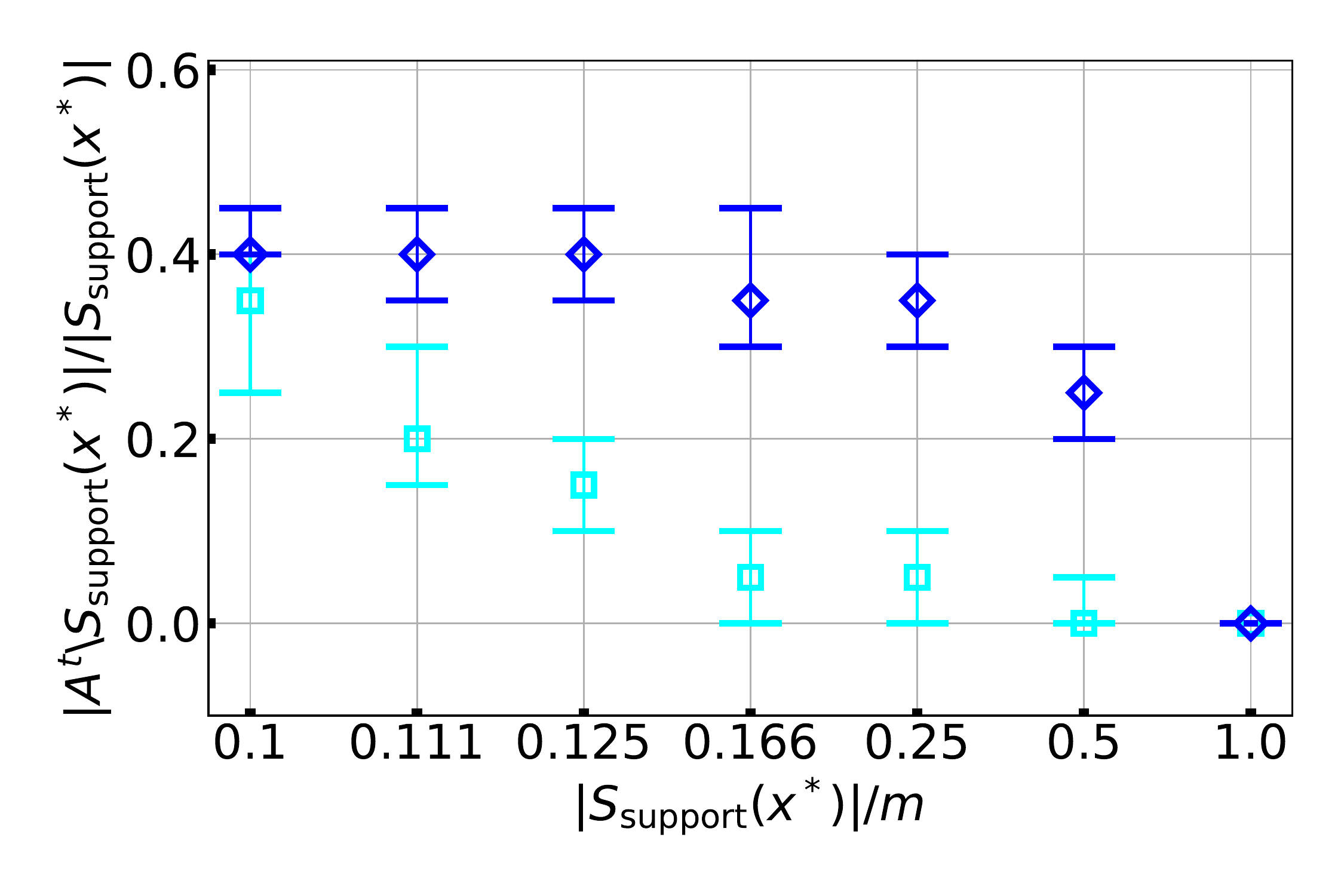}%
    \caption{P3}%    
  \end{subfigure}%
  \\%
  \begin{subfigure}{\figsizefuncr\hsize}%
    \includegraphics[width=\hsize]{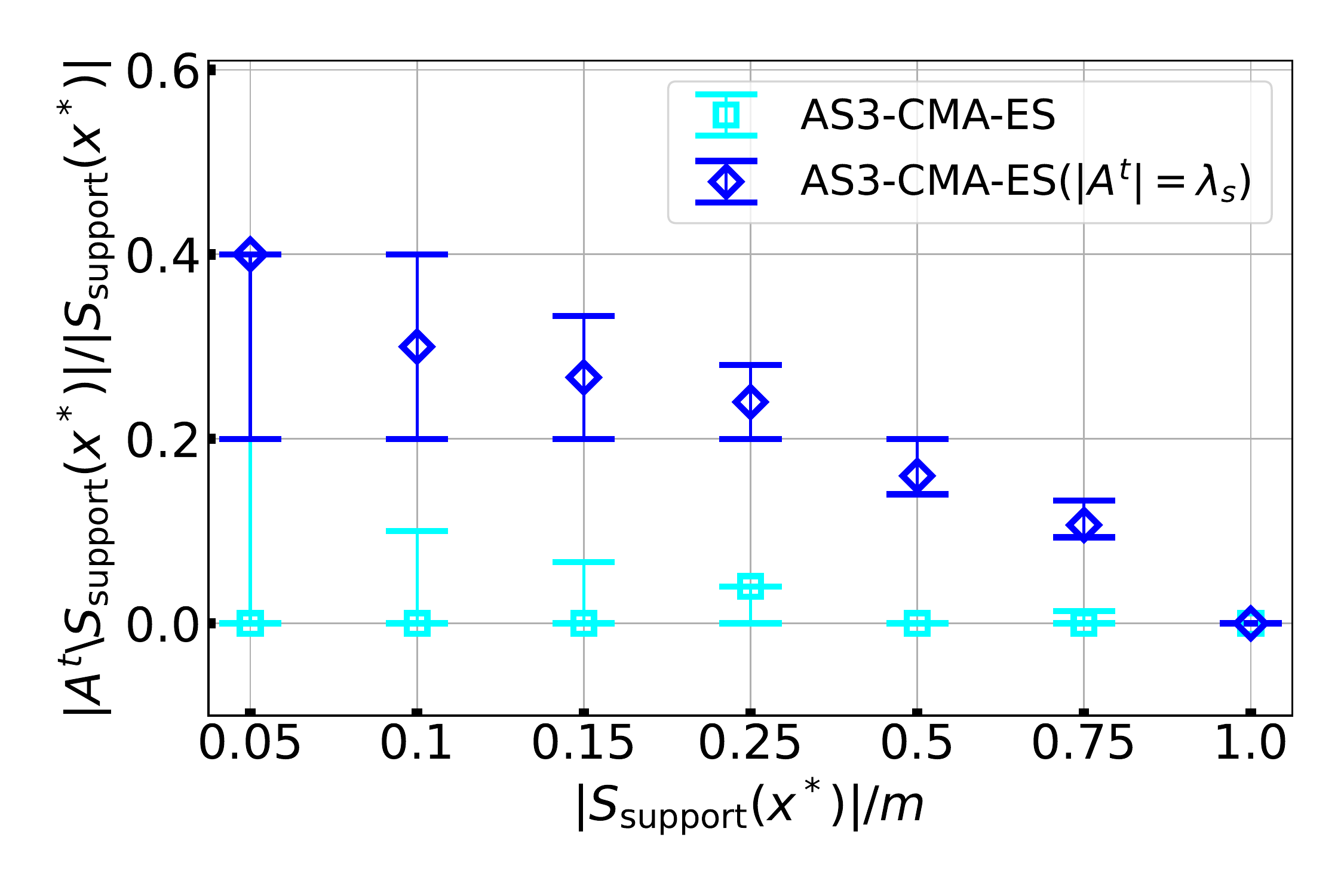}%
    \caption{P4}%    
  \end{subfigure}%  
  \begin{subfigure}{\figsizefuncr\hsize}%
    \includegraphics[width=\hsize]{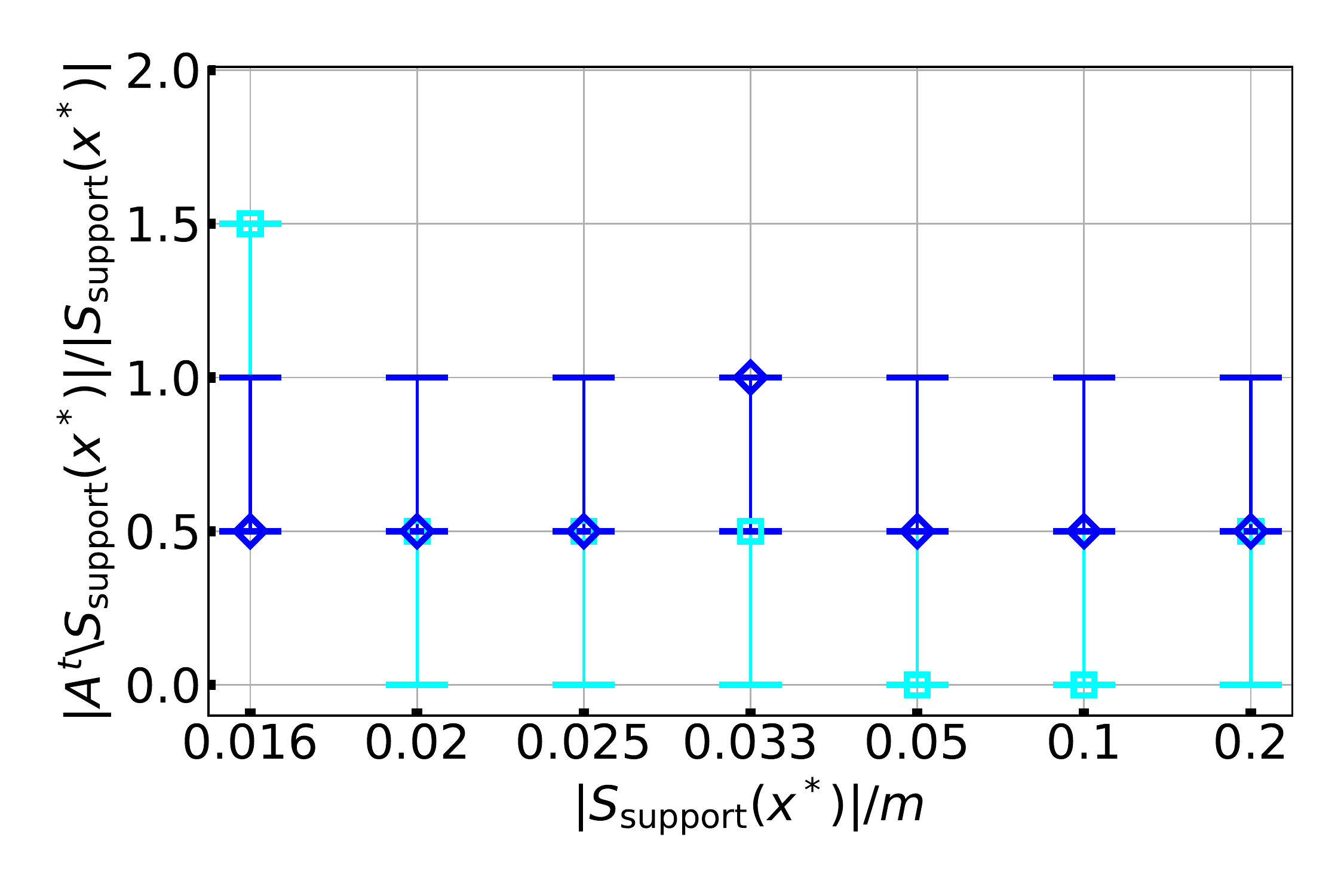}%
    \caption{P5}%    
  \end{subfigure}%    
  \caption{
%   \textcolor{blue}{[R3C10] : 
  Mean and standard deviation of the ratio $\abs{A^t \setminus S_\mathrm{support}(x^*)}/\abs{S_\mathrm{support}(x^*)}$ at the end of optimization resulted from \pbilcma{} and \pbilcma{} with fixed $\lambda_s = \abs{S_\mathrm{support}(x^*)}$ (denoted as \pbilcma{} ($\abs{A^t}=\lambda_s$) at the legend) on problems P1--P5 over 20 trials.
%   }
  }
  \label{fig:lastfaulsepositive}
%\end{minipage}
\end{figure}

\subsection{Comparison (Hypotheses (2)--(4))}

To compare the four approaches with different $m$ and different $\abs{S_\mathrm{support}(x^*)}$ (to test Hypotheses (2)--(4)), we set the problem control parameters as follows. 
We set $m = 100$ and $K =\{ 5, 15, 25, 50, 75, 100\}$ for P1 and P2, 
$m =\{ 20, 40, 80, 120, 160, 180, 200\}$ (hence $K =\{ 1, 2, 4, 6, 8, 9, 10\}$, respectively) for P3, 
$L =\{ 5, 10, 15, 25, 50, 75, 100\}$ and 
$m = 100$ for P4 and $m= \{10, 20, 40, 60, 80, 100, 120\}$ for P5. 
The problem dimension was $n = 10$. 
The scalability of \pbilcma{} against $n$ and $m$ was also tested; the results are presented in \ref{app:scal}.

The comparison results are shown in \Cref{fig:ssensitivity}. 
The average and standard deviation of the number of $f$-calls until the target objective value $\abs{F(m^t) - F(x^*)} < 10^{-12}$ was reached are displayed. 
If optimization failed, the run was treated as a case in which $10^6$ $f$-calls were exhausted.
As a reference, we also plotted the number of $f$-calls made by \texttt{CMA-ES} times the ratio $\abs{S_\mathrm{support}(x^*)} / m$. 
% \textcolor{blue}{[R3C11] :
Additionally, \Cref{fig:gap} shows the changes of gap $F(m^t)-F(x^*)$ obtained by four approaches on P1 with $\abs{S_{\mathrm{support}}(x^*)}/m=0.75$ and P2 with $\abs{S_{\mathrm{support}}(x^*)}/m=0.05$.
% }
% \textcolor{blue}{[R3C1] [R3C8] : 
To support the statistical significance of the differences in \Cref{fig:ssensitivity}, \Cref{table:exresult} shows the $p$-values from 
%\textcolor{blue}{[R3C1] :
Mann–Whitney's $U$-test 
%}
between the number of $f$-calls spent by \pbilcma{} and that of the other approaches on each problem.%
% }
% \textcolor{blue}{[R3C10] : 
To ascertain the number of support scenarios correctly selected in $A^t$ and the number of non-support scenarios wrongly selected in $A^t$, the ratios $\abs{A^t \cap S_\mathrm{support}(x^*)} / \abs{S_\mathrm{support}(x^*)}$ and $\abs{A^t \setminus S_\mathrm{support}(x^*)}/\abs{S_\mathrm{support}(x^*)}$ at the end of trials of \pbilcma{} and \pbilcma{} with fixed $\lambda_s = \abs{S_\mathrm{support}(x^*)}$ on each problem are visualized in \Cref{fig:lastscenarioratio} and \Cref{fig:lastfaulsepositive}. 
\texttt{CMA-ES} and \texttt{lq-CMA-ES} always use all scenarios and $\abs{A^t \cap S_\mathrm{support}(x^*)} / \abs{S_\mathrm{support}(x^*)} = 1$ and $\abs{A^t \setminus S_\mathrm{support}(x^*)}/\abs{S_\mathrm{support}(x^*)} = m/\abs{S_\mathrm{support}(x^*)} - 1$, therefore their results are omitted.
% }

\begin{table}[t]
    \centering
    \caption{
% \textcolor{blue}{[R3C1] [R3C8] :
    The $p$-values from
%\textcolor{blue}{[R3C1] :
    Mann–Whitney's $U$-test
%}
    between the number of $f$-calls spent by \pbilcma{} and that of the other approaches on each problem. The total number of tests is $105$; hence, the Bonferroni correction of the statistical significance $\alpha$ is $\alpha / 105 > 9.5 \times 10^{-5}$ for $\alpha = 10^{-2}$.
    The difference in \Cref{fig:ssensitivity} is regarded as statistically significant with $\alpha = 10^{-2}$ if the corresponding $p$-value is smaller than $9.5 \times 10^{-5}$.
    % }
    }
    \footnotesize
    \begin{tabular}{c|c|c|c|c|c|c|c}
    \toprule
    \multicolumn{8}{c}{P1} \\ \midrule
    \diagbox{Approach}{$\abs{S_\mathrm{support}(x^*)}/m$} & 0.05 & 0.1 & 0.15 & 0.25 & 0.5 & 0.75 & 1.0
    \\
    \hline 
    \texttt{CMA-ES} & $6.73 \times 10^{-8}$ & $6.71 \times 10^{-8}$ & $6.72 \times 10^{-8}$ & $6.75 \times 10^{-8}$ & $6.64 \times 10^{-8}$ &  $1.42 \times 10^{-7}$ &  $4.68 \times 10^{-2}$
    \\
    \pbilcma{} with fixed $\lambda_s$ & $6.79 \times 10^{-8}$ & $1.48 \times 10^{-3}$ & $4.6 \times 10^{-4}$ & $1.06 \times 10^{-7}$ & $1.48 \times 10^{-1}$ & $9.25 \times 10^{-1}$ & $4.38 \times 10^{-2}$ 
    \\
    \texttt{lq-CMA-ES} & $7.39 \times 10^{-6}$ & $6.74 \times 10^{-8}$ & $6.77 \times 10^{-8}$ & $6.79 \times 10^{-8}$ & $6.79 \times 10^{-8}$ & $6.77 \times 10^{-8}$ & $6.80 \times 10^{-8}$ 
    \\
    \bottomrule
    \toprule
    \multicolumn{8}{c}{P2} \\ \midrule
    \diagbox{Approach}{$\abs{S_\mathrm{support}(x^*)}/m$} & 0.05 & 0.1 & 0.15 & 0.25 & 0.5 & 0.75 & 1.0
    \\
    \hline
    \texttt{CMA-ES} & $6.77 \times 10^{-8}$ & $6.70 \times 10^{-8}$ & $6.75 \times 10^{-8}$ & $6.73 \times 10^{-8}$ & $6.77 \times 10^{-8}$ &  $2.74 \times 10^{-4}$ &  $3.36 \times 10^{-4}$
    \\
    \pbilcma{} with fixed $\lambda_s$ & $2.23 \times 10^{-2}$ & $9.89 \times 10^{-1}$ & $1.33 \times 10^{-2}$ & $4.38 \times 10^{-2}$ & $9.03 \times 10^{-3}$ & $2.61 \times 10^{-1}$ &  $4.15 \times 10^{-5}$
    \\
    \texttt{lq-CMA-ES} & $6.80 \times 10^{-8}$ & $2.06 \times 10^{-6}$ & $3.05 \times 10^{-4}$ & $6.71 \times 10^{-8}$ & $6.76 \times 10^{-8}$ &  $6.74 \times 10^{-8}$ &  $6.79 \times 10^{-8}$ 
    \\
    \bottomrule
    \toprule
    \multicolumn{8}{c}{P3} \\ \midrule
    \diagbox{Approach}{$\abs{S_\mathrm{support}(x^*)}/m$} & 0.1 & 0.111 & 0.125 & 0.166 & 0.25 & 0.5 & 1.0
    \\
    \hline
    \texttt{CMA-ES} & $6.77 \times 10^{-8}$ & $6.71 \times 10^{-8}$ & $6.73 \times 10^{-8}$ & $6.78 \times 10^{-8}$ & $1.20 \times 10^{-6}$ &  $1.38 \times 10^{-6}$ &  $6.77 \times 10^{-8}$ 
    \\
    \pbilcma{} with fixed $\lambda_s$ & $6.79 \times 10^{-8}$ & $6.79 \times 10^{-8}$ & $6.80 \times 10^{-8}$ & $7.90 \times 10^{-8}$ & $2.59 \times 10^{-5}$ &  $1.93 \times 10^{-2}$ &  $6.79 \times 10^{-8}$  
    \\
    \texttt{lq-CMA-ES} & $6.28 \times 10^{-8}$ & $6.77 \times 10^{-8}$ & $6.76 \times 10^{-8}$ & $6.79 \times 10^{-8}$ & $1.20 \times 10^{-6}$ &  $6.92 \times 10^{-7}$ &  $2.36 \times 10^{-6}$ 
    \\
    \bottomrule
    \toprule
    \multicolumn{8}{c}{P4} \\ \midrule
    \diagbox{Approach}{$\abs{S_\mathrm{support}(x^*)}/m$} & 0.05 & 0.1 & 0.15 & 0.25 & 0.5 & 0.75 & 1.0
    \\
    \hline 
    \texttt{CMA-ES} & $6.75 \times 10^{-8}$ & $6.78 \times 10^{-8}$ & $6.77 \times 10^{-8}$ & $6.79 \times 10^{-8}$ & $6.77 \times 10^{-8}$ &  $6.79 \times 10^{-8}$ &  $6.78 \times 10^{-8}$ 
    \\
    \pbilcma{} with fixed $\lambda_s$ & $1.20 \times 10^{-6}$ & $6.76 \times 10^{-8}$ & $6.80 \times 10^{-8}$ & $3.70 \times 10^{-5}$ & $6.74 \times 10^{-8}$ &  $6.75 \times 10^{-8}$ &  $6.77 \times 10^{-8}$ 
    \\
    \texttt{lq-CMA-ES} & $6.80 \times 10^{-8}$ & $6.80 \times 10^{-8}$ & $6.80 \times 10^{-8}$ & $6.80 \times 10^{-8}$ & $6.79 \times 10^{-8}$ &  $2.24 \times 10^{-4}$ &  $1.33 \times 10^{-1}$  
    \\
    \bottomrule
    \toprule
    \multicolumn{8}{c}{P5} \\ \midrule
    \diagbox{Approach}{$\abs{S_\mathrm{support}(x^*)}/m$} & 0.016 & 0.02 & 0.025 & 0.033 & 0.05 & 0.1 & 0.2
    \\
    \hline 
    \texttt{CMA-ES} & $6.74 \times 10^{-8}$ & $6.78 \times 10^{-8}$ & $6.76 \times 10^{-8}$ & $6.77 \times 10^{-8}$ & $6.76 \times 10^{-8}$ &  $6.76 \times 10^{-8}$ &  $6.73 \times 10^{-8}$
    \\
    \pbilcma{} with fixed $\lambda_s$ & $7.98 \times 10^{-9}$ & $8.01 \times 10^{-9}$ & $8.01 \times 10^{-9}$ & $7.99 \times 10^{-9}$ & $8.01 \times 10^{-9}$ & $7.99 \times 10^{-9}$ &  $8.01 \times 10^{-9}$ 
    \\
    \texttt{lq-CMA-ES} &$6.78 \times 10^{-8}$ & $6.80 \times 10^{-8}$ & $6.80 \times 10^{-8}$ & $6.79 \times 10^{-8}$ & $6.79 \times 10^{-8}$ &  $6.79 \times 10^{-8}$ &  $6.80 \times 10^{-8}$
    \\
    \bottomrule
    \end{tabular}
    \label{table:exresult}
\end{table}

\paragraph{\pbilcma{} vs \texttt{CMA-ES} (Hypothesis (2))}\label{sec:ex2}

\Cref{fig:ssensitivity} shows that \pbilcma{} spent fewer $f$-calls than \texttt{CMA-ES} in most cases. 
In particular, \pbilcma{} was more efficient when $\abs{S_\mathrm{support}(x^*)} / m$ was smaller.
% \textcolor{blue}{[R3C11] :
% As shown in \Cref{fig:gap}, at $\abs{S_{\mathrm{support}}(x^*)}/m=0.05$ on P2, the gradient of the gap $F(m^t)-F(x^*)$ obtained by \pbilcma{} was larger than that by \texttt{CMA-ES} around $10^4$ $f$-calls because $\abs{A^t}$ was less than $m$.
% }
% \textcolor{blue}{[R3C10]:
When $\abs{S_\mathrm{support}(x^*)} / m \leq 0.25$, as shown in \Cref{fig:lastscenarioratio} and \Cref{fig:lastfaulsepositive}, $\abs{A^t \cap S_\mathrm{support}(x^*)} / \abs{S_\mathrm{support}(x^*)}$ resulting from \pbilcma{} was close to $1$, and $\abs{A^t \setminus S_\mathrm{support}(x^*)}/\abs{S_\mathrm{support}(x^*)}$ was relatively small in comparison with the ratio $\abs{A^t \setminus S_\mathrm{support}(x^*)}/\abs{S_\mathrm{support}(x^*)}=m/\abs{S_\mathrm{support}(x^*)} - 1$ of \texttt{CMA-ES} (for example, in case of $m/\abs{S_\mathrm{support}(x^*)}=0.05$, $m/\abs{S_\mathrm{support}(x^*)} - 1$ is $19$). These results implicitly support that \pbilcma{} could approximate the worst-case objective function without sampling all scenarios at the convergence.
% }
By contrast, if $\abs{S_\mathrm{support}(x^*)} / m \approx 1$, \pbilcma{} is designed to obtain similar results to those of \texttt{CMA-ES}, and the advantage of \pbilcma{} over \texttt{CMA-ES} was reduced.

\paragraph{\pbilcma{} vs \pbilcma{} with fixed $\lambda_s = \abs{S_\mathrm{support}(x^*)}$(Hypothesis (3))} \label{sec:ex3}

On P1, P2, and P4, \pbilcma{} with $\lambda_s = \abs{S_\mathrm{support}(x^*)}$ achieved nearly ideal speed-up over \texttt{CMA-ES}, which is $\abs{S_\mathrm{support}(x^*)} / m$ times fewer $f$-calls (denoted as \texttt{Target}). 
% \textcolor{blue}{[R3C10]:
For the successful convergence on P1, P2, and P4, \Cref{fig:lastscenarioratio} shows that \pbilcma{} with $\lambda_s = \abs{S_\mathrm{support}(x^*)}$ needs more than half of the support scenarios at the convergence. 
% }
For a relatively small $\abs{S_\mathrm{support}(x^*)} / m$ situation, the speed-up factor is almost ideal for P2, but it was reduced on P1 and P4. This is because $\abs{S_\mathrm{support}(H_\gamma^t)} \leq \lambda_s$ is guaranteed at P2, whereas $\abs{S_\mathrm{support}(H_\gamma^t)}$ can be greater than $\lambda_s$, and the support scenarios can change over time on P1 and P4 until the search distribution is sufficiently concentrated around $x^*$. This, as well as the time required for the adaptation of $p_s^t$, may be the reason for non-ideal speed-up. 
On P3, this defect was observed even for a relatively high ratio $\abs{S_\mathrm{support}(x^*)} / m$. On P5, $\lambda_s = \abs{S_\mathrm{support}(x^*)}= 2$ was too small to approximate the worst-case objective; hence, the optimization failed.

We observe that the efficiency of \pbilcma{} is competitive or slightly worse than that of \pbilcma{} with fixed $\lambda_s = \abs{S_\mathrm{support}(x^*)}$ on P1, P2, and P4. 
This is a promising result, as \pbilcma{} with fixed $\lambda_s = \abs{S_\mathrm{support}(x^*)}$ exhibited nearly ideal performance, and \pbilcma{} achieved competitive performance without knowing $\abs{S_\mathrm{support}(x^*)}$. 
On P3 and P5, \pbilcma{} exhibited even better performance than \pbilcma{} with fixed $\lambda_s = \abs{S_\mathrm{support}(x^*)}$, where the number $\abs{S_\mathrm{support}(H_\gamma^t)}$ of the support scenarios in the search area $H_\gamma^t$ may be significantly greater than $\abs{S_\mathrm{support}(x^*)}$ and changes over time. The adaptive behavior of the number $\abs{A^t}$ of subsampled scenarios is helpful for such situations.

\paragraph{\pbilcma{} vs \texttt{lq-CMA-ES}(Hypothesis (4))}\label{sec:ex4}

\texttt{lq-CMA-ES} achieved a speed-up of factor $2$ to $3$ over \texttt{CMA-ES}, independently on the ratio $\abs{S_\mathrm{support}(x^*)} / m$. 
Therefore, 
if $\abs{S_\mathrm{support}(x^*)} / m$ is relatively small, \pbilcma{} is a better choice than \texttt{lq-CMA-ES}, whereas
if $\abs{S_\mathrm{support}(x^*)} / m$ is relatively high, (in particular when $\abs{S_\mathrm{support}(x^*)} / m = 1$), \texttt{lq-CMA-ES} is a better choice than \pbilcma. 
% \textcolor{blue}{[R3C11] :
This tendency is also observed in \Cref{fig:gap}.
% As shown in \Cref{fig:gap}, at $\abs{S_{\mathrm{support}}(x^*)}/m=0.75$ on P1, the gap $F(m^t)-F(x^*)$ obtained from \texttt{lq-CMA-ES} was decreased faster than that obtained from \pbilcma{}.
% }
% }
However, there are also cases such as on P4 and P5 where \pbilcma{} is significantly better than \texttt{lq-CMA-ES} even when the ratio $\abs{S_\mathrm{support}(x^*)} / m$ is close to $1$.
By contrast, \pbilcma{} is more advantageous as the ratio $\abs{S_\mathrm{support}(x^*)} / m$ is lower.

\section{Application to Well Placement Optimization}\label{sec:rwa}

We demonstrate the usefulness of \pbilcma{} in comparison with \texttt{CMA-ES} and \texttt{lq-CMA-ES} for well placement optimization for carbon dioxide capture and storage (CCS) with multiple geological models. 

\subsection{Problem Description}

The problem is to optimize the placement of three injection wells to maximize the injectable $\mathrm{CO_2}$ volume.
The objective function is the total injection volume of $\mathrm{CO_2}$ from the three wells. The design variable $x = (x_1, x_2, x_3)$ is the 2D coordinate, where $x_i = (w_1^i, w_2^i)$ is the coordinate of the placement of the $i$th well.
The problem dimension is $n = 6$.

We consider obtaining a robust well placement for the uncertainty of the geological property distribution (e.g., porosity or permeability distribution). 
The uncertainty of the geological property distribution is worth considering because the performance of injection wells greatly depends on the geological properties of their location. 
In this experiment, we created $m = 50$ models with different geological property distributions to represent geological uncertainty. 
The total injectable $\mathrm{CO_2}$ volume through three wells varies for different models.
Here, each model is considered for each scenario, and the objective function value for the $s$th model is $f(x, s)$. 

The optimization problem is formulated as the following max--min problem     
\begin{equation}
  \argmax_{x \in \mathbb{X}} F(x), \quad \text{where} \quad F(x) = \min_{s=1,\cdots,m} f(x, s) \enspace.\label{eq:maxmin} 
\end{equation}
Here, $F(x)$ is the minimum injectable $\mathrm{CO_2}$ volume among the $m$ models. 

We formulated the objective function as follows. 
Let $f_{i,j,s}$ be the injectable $\mathrm{CO_2}$ volume in model $s$ at a single injection well located at $(i, j)$ for $i = 1,\dots, 50$ and $j = 1, \dots, 50$.
We computed $f_{i,j,s}$ by performing numerical simulations, which amounted to 125,000 simulations in total. 
Let $f_b(\check{x}, s)$ be the approximated injectable $\mathrm{CO_2}$ volume in model $s$ at a single injection well located at $\check{x} \in [1, 50]^2$. 
This value is computed by the bilinear interpolation of $f_{i,j,s}$. 
The objective function is defined as follows.
\begin{equation} 
f(x, s) = \sum_{i=1}^{3} f_b(x_{i:C},s) \prod_{j=1}^{i - 1} (1 - \exp(- d(x_{i:3}, x_{j:3}))) \enspace, \label{eq:IWOobj}
\end{equation} 
where $x_{i:3}$ is the well coordinate with the $i$th greatest injectable $\mathrm{CO_2}$ volume $f_b(x_{i:3}, s)$, i.e.,  $f_b(x_{1:3}, s) \geq f_b(x_{2:3}, s) \geq f_b(x_{3:3}, s)$, and $d(\check{y}, \check{z})$ represents the distance between well $\check{y}$ and well $\check{z}$.
The rationale behind \eqref{eq:IWOobj} is as follows.
First, the total injectable $\mathrm{CO_2}$ volume $f(x, s)$ must be no less than the injectable $\mathrm{CO_2}$ volume $f_b(x_i, s)$ from each injection well. 
Second, $f(x,s)$ must be less than the sum of the injectable $\mathrm{CO_2}$ volume from each well when there is only one well because of the pressure interference between the wells. 
That is, we require $\max_{i=1,2,3}f_s(x_i, s) < f(x,s) < \sum_{i=1}^3 f_s (x_i, s)$. Our objective function \eqref{eq:IWOobj} reflects this requirement, where the pressure interference is represented by $\prod_{j=1}^{i - 1} (1 - \exp(- d(x_{i:3}, x_{j:3})))$.\footnote{
Our motivation is to compare the worst-case performance of \pbilcma{}, \texttt{CMA-ES}, and \texttt{lq-CMA-ES}. For this purpose, it would be preferable to run several trials for each approach and to compute the worst-case performance at each iteration of each trial.
However, because $\mathrm{CO_2}$ flow simulation requires high computational resources, such as supercomputers \cite{yamaoto2012,miyagi@ghgt}, running multiple trials for each approach was impossible. 
By contrast, \Cref{eq:IWOobj} is computationally inexpensive to evaluate and we expect that it reflects the characteristics of the reality. 
} 

\begin{figure}[t]
  \centering
  \begin{minipage}{0.55\textwidth}
  \begin{subfigure}{0.5\hsize}%
    \includegraphics[trim=20 20 20 20,clip, width=\hsize]{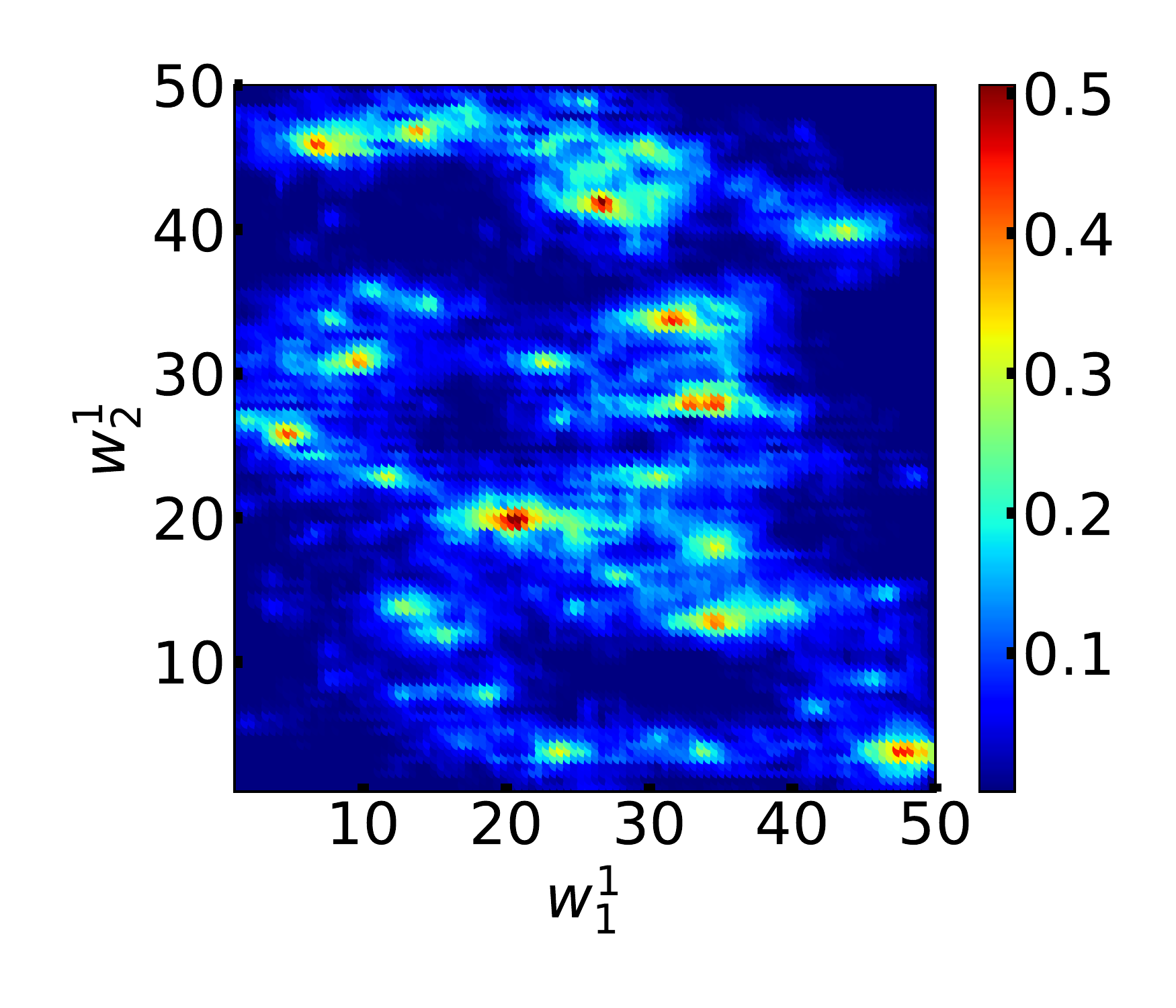}%
    \caption{$\max_{s \in S}f_b(x_1, s)$}%    
  \end{subfigure}%
  \begin{subfigure}{0.5\hsize}%
    \includegraphics[trim=20 20 20 20,clip, width=\hsize]{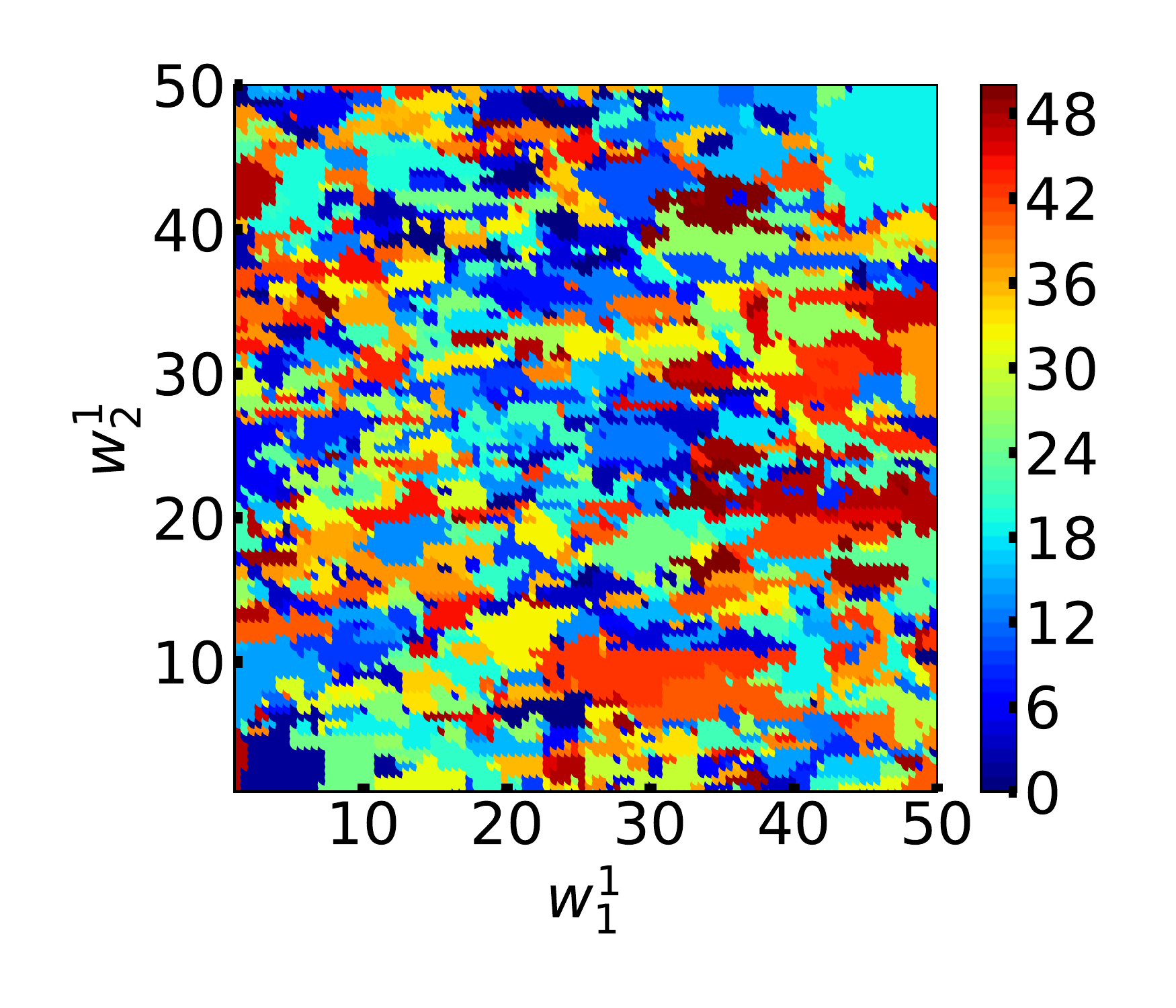}%
    \caption{$\argmax_{s \in S} f_b(x_1, s)$}%    
  \end{subfigure}%
  \caption{Visualization of the optimization problem.}%
  \label{fig:2Dlandscape}%
\end{minipage}%
\begin{minipage}{0.45\textwidth}%
  \centering%
    \includegraphics[trim=20 30 20 30,clip,width=\hsize]{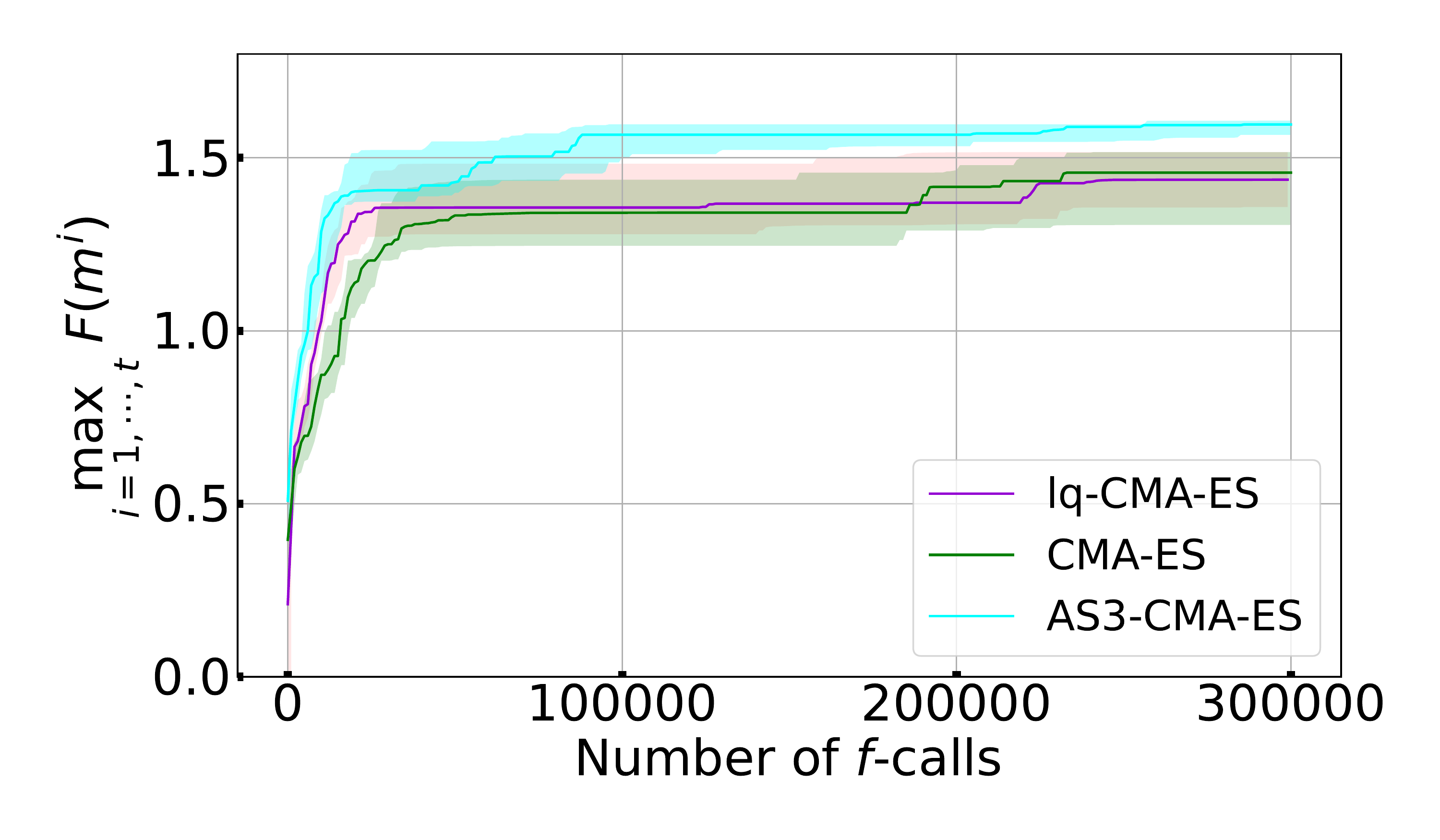}%  
    \caption{Best-so-far worst-case objective value $\max_{i=1,\dots,t}F(m^i)$. Solid line: median (50 percentile) over 20 runs; Shaded area: interquartile range (25--75 percentile range) over 20 runs. 
    }%
  \label{fig:IWPO-20runs}%
\end{minipage}%
\end{figure}

\Cref{fig:2Dlandscape} visualizes the injectable $\mathrm{CO_2}$ volume at the worst case, $\max_{s=1, \dots, m} f_b(\check{x}, s)$, when there is only one injection well and the support scenario at each location, $\check{x}$. 
All the scenarios were included in the support scenarios $S_\mathrm{support}([1, 50]^2)$ over the search space. 
However, if we focus on the neighborhood of a local maximum, e.g., $\check{x} \approx (20, 20)$, the number of support scenarios is no greater than 10. 
We expect that this characteristic is inherited by $f$. Based on the results observed in \Cref{sec:exp}, we expect that \pbilcma{} is more efficient than \texttt{lq-CMA-ES} because the ratio $S_\mathrm{support}(x^*) / m$ is relatively small.

\subsection{Experimental Settings}

We applied \pbilcma, \texttt{CMA-ES}, and \texttt{lq-CMA-ES} to this problem. 
The search domain was $\X = [1,50]^6$, with $S = \llbracket 1, 50\rrbracket$. 
We used \texttt{pycma} for the implementation of \texttt{lq-CMA-ES}. We implemented other approaches using the version of the CMA-ES proposed in \cite{akimoto2019} as the baseline. For a fair comparison between \texttt{lq-CMA-ES} and the other approaches, we turned off the diagonal acceleration mechanism of \cite{akimoto2019}. All hyperparameters were set to their default values.
The initial mean vector and covariance matrix of the CMA-ES were set as $m^0 \sim \mathcal{U}(\X)$ and $\Sigma^0 = (12.5)^2\cdot I_n$. 
The parameters for \pbilcma{} were set as follows: $c_p = 0.3$, $\eta = 0.3$, $\epsilon = 1/m$, $\gamma = 0.99$, and $p_s^0 = 0.1$ for all $s \in S$.
% \textcolor{blue}{[R3C7]: 
We used the same initial mean vector and covariance matrix as initial settings for \texttt{lq-CMA-ES}. 
The other hyperparameters for \texttt{lq-CMA-ES} were set to their default values implemented in \texttt{pycma}.
% }

In this experiment, we employed a simple restart strategy with default $\lambda_x$ to deal with multimodality.\footnote{
The typical approach to dealing with multimodality is to increase $\lambda_x$. However, for problems without a global structure, a greater $\lambda_x$ is not helpful in converging to a better local optimal solution. Conversely, the CMA-ES tends to converge to the same local optimal solutions as $\lambda_x$ increases. Moreover, the number of possible restarts decreases if we increase $\lambda_x$. Because \Cref{fig:2Dlandscape} does not exhibit a global structure, we employed a simple restart strategy.
We ran the same experiments with the IPOP restart strategy \cite{Auger2018cec}, where $\lambda_x$ is doubled at each restart. We observed similar differences between the compared approaches as in \Cref{fig:IWPO-20runs}, but the number of restarts performed by each approach was smaller and the performance of each approach was lower.} 
The termination condition for restart was an excessively small coordinate-wise standard deviation $\max_{i =1,\dots,n} \Sigma^t_{i,i} < 10^{-8}$, where $\Sigma^t_{i,i}$ is the $i$th diagonal element of the covariance matrix $\Sigma^t$.
If this condition was satisfied, the mean vector and covariance matrix were initialized as $m^0 \sim \mathcal{U}(\X)$ and $\Sigma^0 = (12.5)^2\cdot I_n$, and $p_s^0 = 0.1$ for all $s \in S$ for \pbilcma.

We evaluated the performance of each algorithm by computing the worst-case performance $F(m^t)$ evaluated at the mean vector $m^t$ at each iteration. We performed 20 independent trials for each algorithm with a maximum of $f$-calls of $3 \times 10^5$.

\subsection{Results and Analysis}

\begin{figure}[t]
  \centering
  \begin{minipage}{0.35\textwidth}
  \centering
    \includegraphics[width=\hsize]{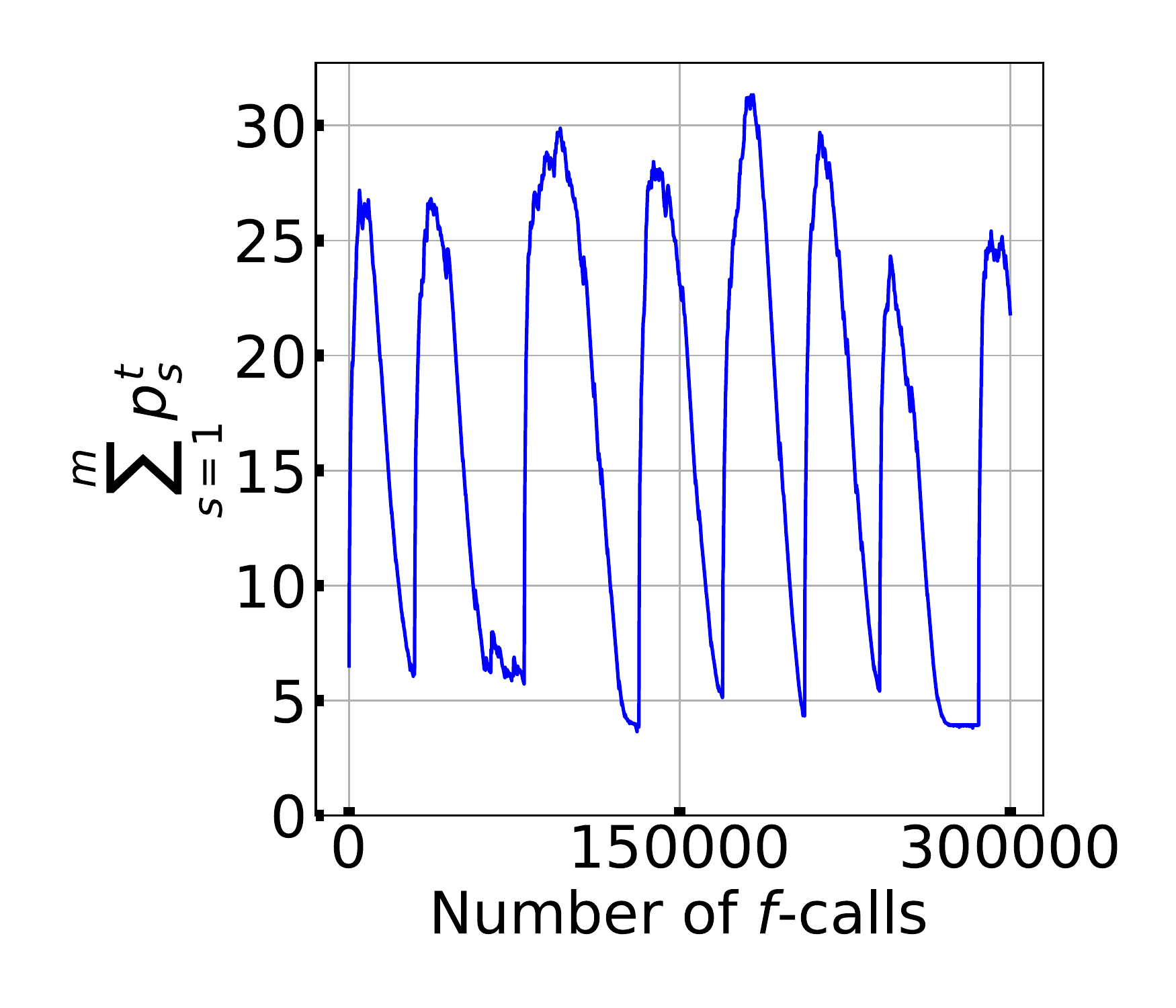}%  
  \caption{History of $\E[\abs{A^t}] = \sum_{s=1}^m p^t_s $ on a typical run of \pbilcma{}.}%
  \label{fig:IWPO-pts}
  \end{minipage}
\quad
  \begin{minipage}{0.6\textwidth}
  \centering
  \begin{subfigure}{0.5\hsize}%
    \includegraphics[width=\hsize]{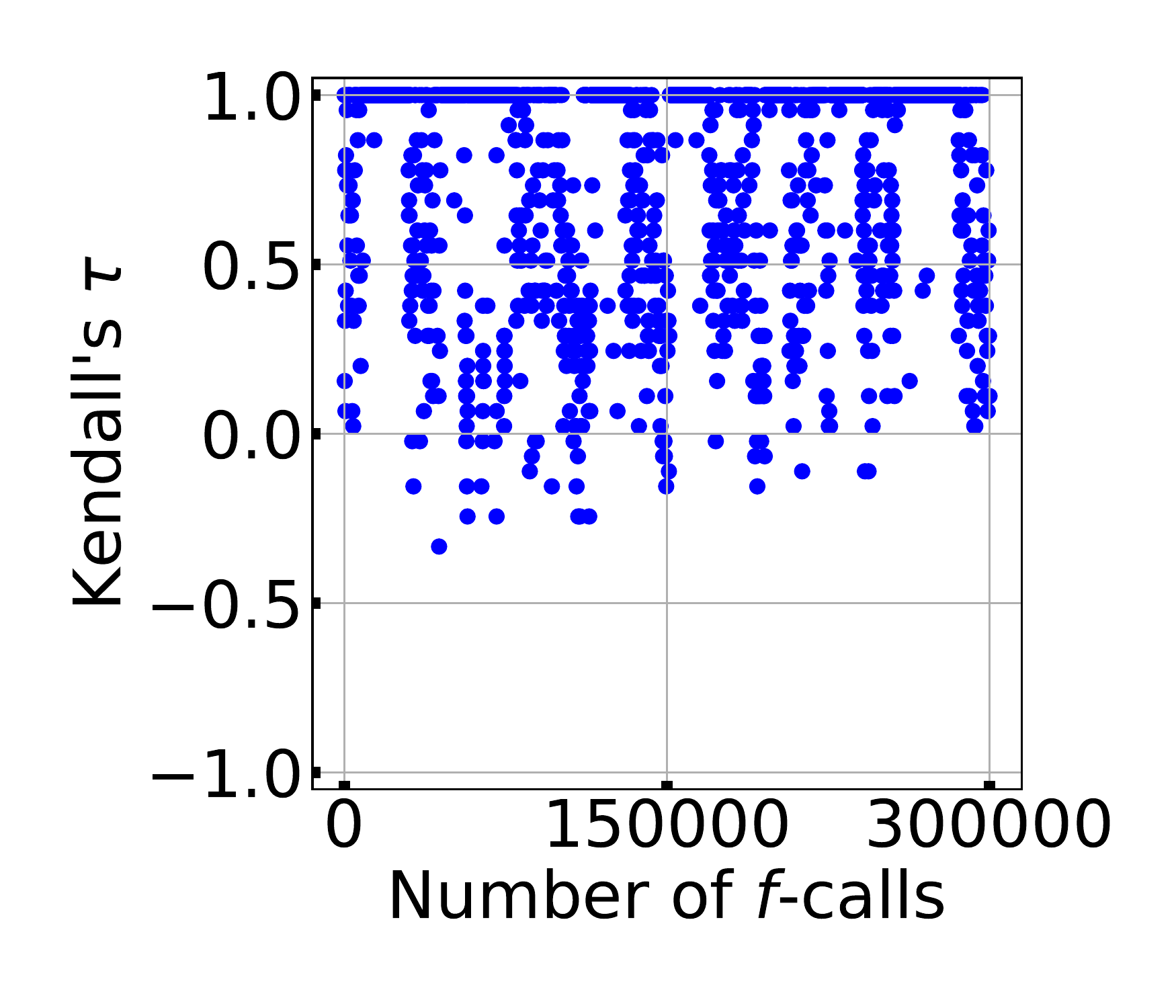}%
    \caption{\pbilcma{}}%    
  \end{subfigure}%
  \begin{subfigure}{0.5\hsize}%
    \includegraphics[width=\hsize]{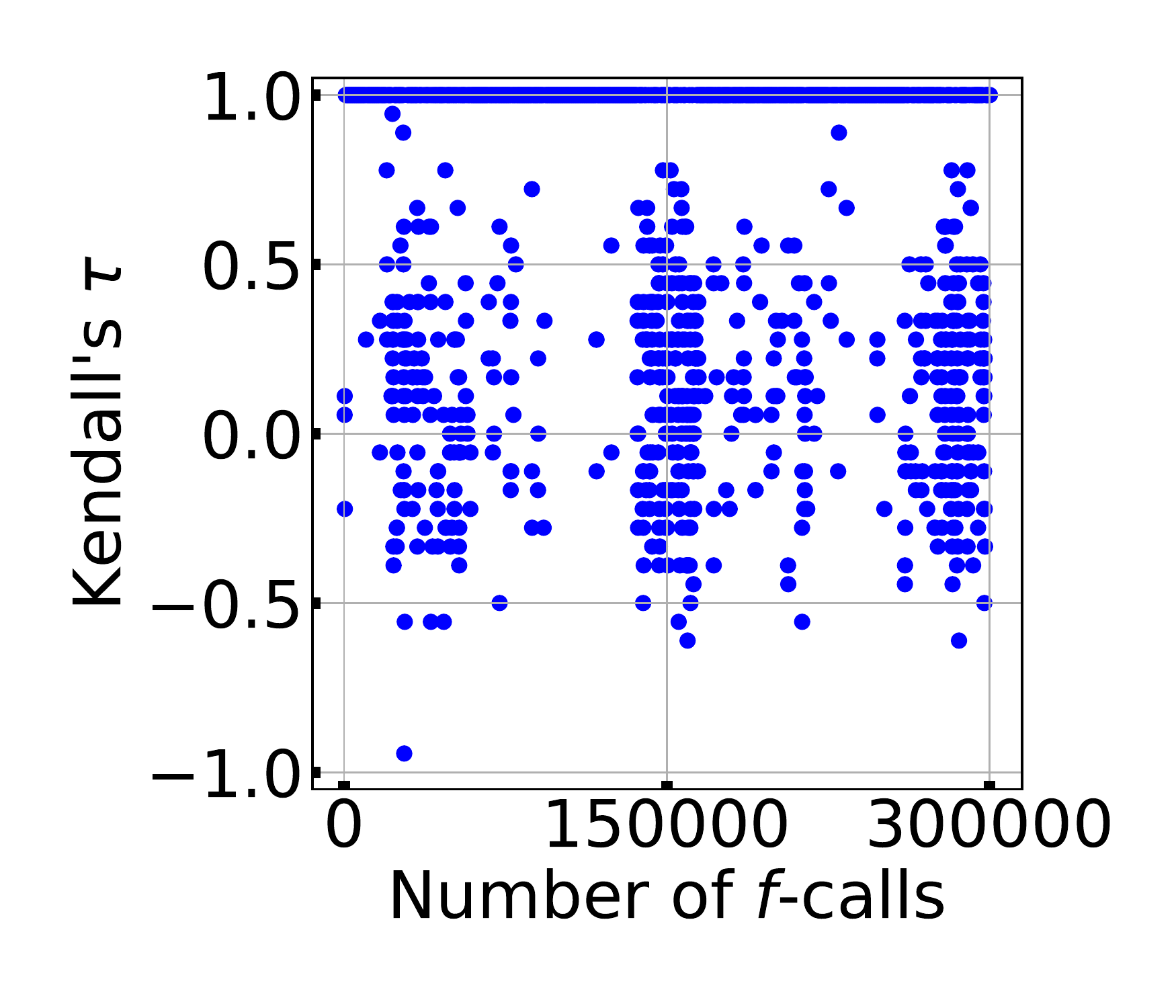}%
    \caption{\texttt{lq-CMA-ES}}%    
  \end{subfigure}%
  \caption{History of Kendall's $\tau$ between the rankings of the solution candidates based on the worst-case objective function values and the ranking computed inside each algorithm in a typical run.}
  \label{fig:IWPO-kendall}
  \end{minipage}
\end{figure}

\Cref{fig:IWPO-20runs} shows the median (50 percentile) and interquartile range (25--75 percentile range) of the best-so-far worst-case performance values $\max_{i = 1,\dots,t} F(m^i)$.
% \textcolor{blue}{[R1C3]:
As a summary of the computational results, the median and interquartile range of the best-so-far worst-case performance values at the end of the optimization and $p$-values against $\pbilcma{}$ resulting from 
%\textcolor{blue}{[R3C1] :
Mann–Whitney's $U$-test
%}
are shown in \Cref{table:wellplacementsummary}. 
\Cref{fig:IWPO-Fmt} shows the history of the worst-case performance at each iteration of each algorithm on a typical trial.
The sharp drops of $F(m^t)$ indicate restarts of the algorithms.
\Cref{fig:IWPO-pts} shows the history of the sum $\sum_{s = 1}^{m} p_s^t$ of the sampling probability of scenarios in \pbilcma{} on a typical trial. 
Note that this is the expected number of sampled scenarios at each iteration, i.e., $\E[\abs{A^t}]$. 
\Cref{fig:IWPO-kendall} shows Kendall's $\tau$ between the ranking of the worst case objective function values $\{F(x_i^t)\}_{i=1}^{\lambda_x}$ and the ranking of the solution candidates computed inside the algorithms, which are the ranking of $\{F(x_i^t; A^t)\}$ in \pbilcma{}. Higher values indicate higher correlations between the true and estimated values. 

\paragraph{\pbilcma{} vs \texttt{CMA-ES}}
\Cref{fig:IWPO-20runs} and \Cref{table:wellplacementsummary} show that \pbilcma{} outperformed \texttt{CMA-ES}. \pbilcma{} obtained higher $50$ percentile values of $\max_{t} F(m^t)$ from the beginning of the optimization, and smaller interquartile ranges at $300,000$ $f$-calls. 
We consider this to be because \pbilcma{} performed more restarts within the fixed budget of $f$-calls, as may be noted from \Cref{fig:IWPO-Fmt}.
Because the worst-case objective has multiple local optima, restarts are essential to obtain a better local optimum. 
\pbilcma{} performed more restarts because it saves $f$-calls for each restart by subsampling a scenario subset, whose cardinality decreased to around $10$ at the end of each restart (see \Cref{fig:IWPO-pts}). 
The smaller interquartile range can also be attributed to the greater number of restarts because the best among more local maxima have less variation than the best among fewer local maxima.

\begin{table*}[]
    \centering\small
    \caption{
% \textcolor{blue}{[R1C3] :
    Median and interquartile range of best-so-far worst-case objective value over 20 runs at the end of the optimization, and $p$-values from 
%\textcolor{blue}{[R3C1] :
    Mann–Whitney's $U$-test 
%}
    on the best-so-far worst-case values obtained by \pbilcma{} and \texttt{CMA-ES} at $100,000$, $200,000$, and $300,000$ $f$-calls.
    The number of tests is $6$; hence, the Bonferroni correction of the statistical significance $\alpha$ is $\alpha / 6 > 1.66 \times 10^{-3}$ for $\alpha = 10^{-2}$, indicating that all the results were statistically significant with $\alpha = 10^{-2}$. }
    % }
    \label{tab:result}
    \begin{tabular}{c|c|c|c|c|c}
    \toprule
    Approach & Median & Interquartile range & \multicolumn{3}{c}{$p$-values against \pbilcma{}}  
    \\
    \cline{4-6}
    & 300,000 $f$-calls & 300,000 $f$-calls & 100,000  $f$-calls & 200,000 $f$-calls & 300,000 $f$-calls \\
    \hline 
    \pbilcma{} &  $1.60$ &  $0.04$ & -- & -- & --
    \\
    % \hline
    \texttt{CMA-ES} & $1.46$ & $0.21$ &  $1.06 \times 10^{-2}$ & $1.79 \times 10^{-4}$ & $1.25 \times 10^{-5}$
    \\
    % \hline
    \texttt{lq-CMA-ES} & $1.44$ & $0.16$ & $1.12 \times 10^{-3}$ & $1.38 \times 10^{-6}$ & $1.20 \times 10^{-6}$
    \\
   \bottomrule
    \end{tabular}
    \label{table:wellplacementsummary}
\end{table*}

\providecommand{\figsizeiwpo}{0.33}
\begin{figure}[t]
  \centering
  \begin{subfigure}{\figsizeiwpo\hsize}%
    \includegraphics[width=\hsize]{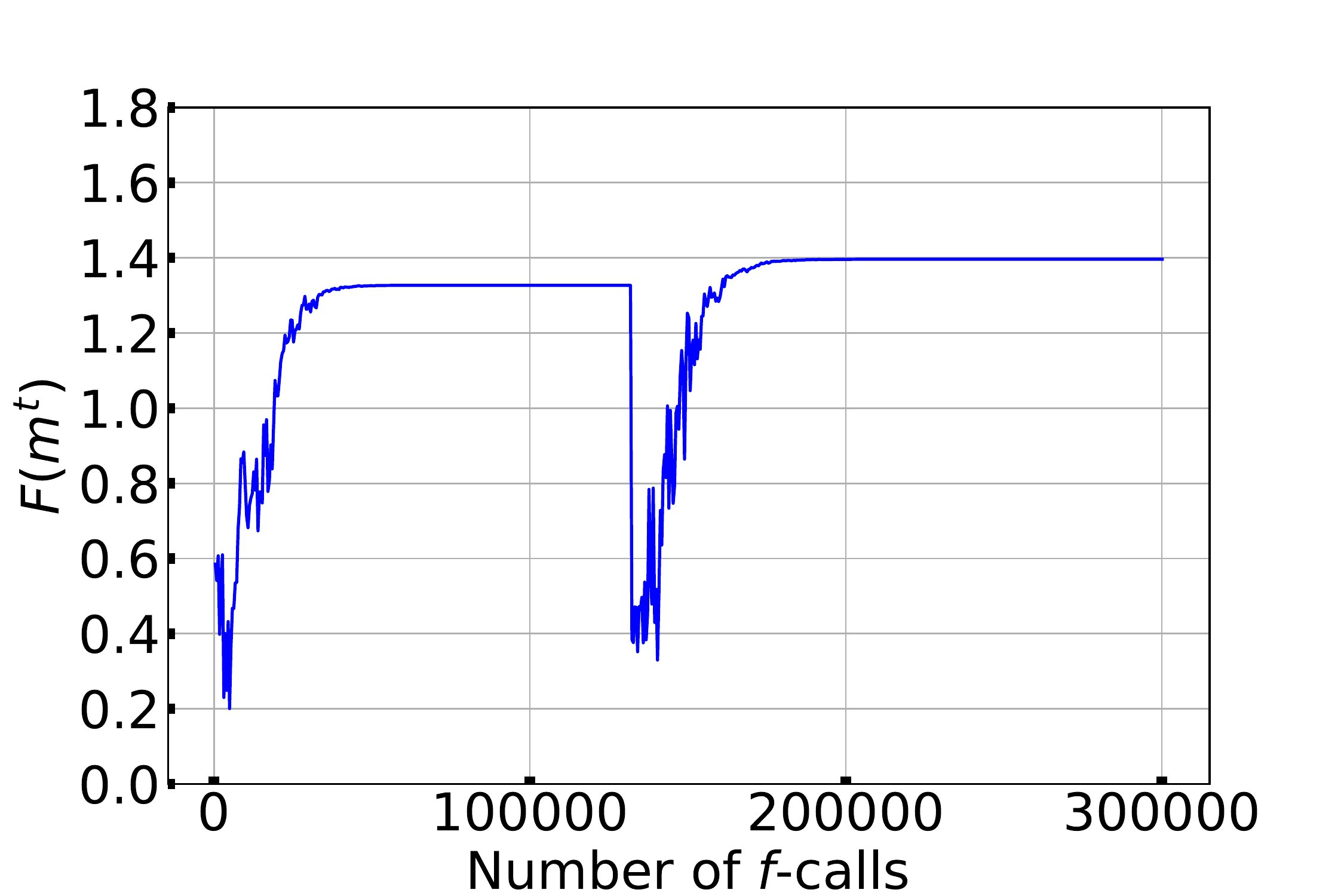}%
    \caption{\texttt{CMA-ES}}%    
  \end{subfigure}%
  \begin{subfigure}{\figsizeiwpo\hsize}%
    \includegraphics[width=\hsize]{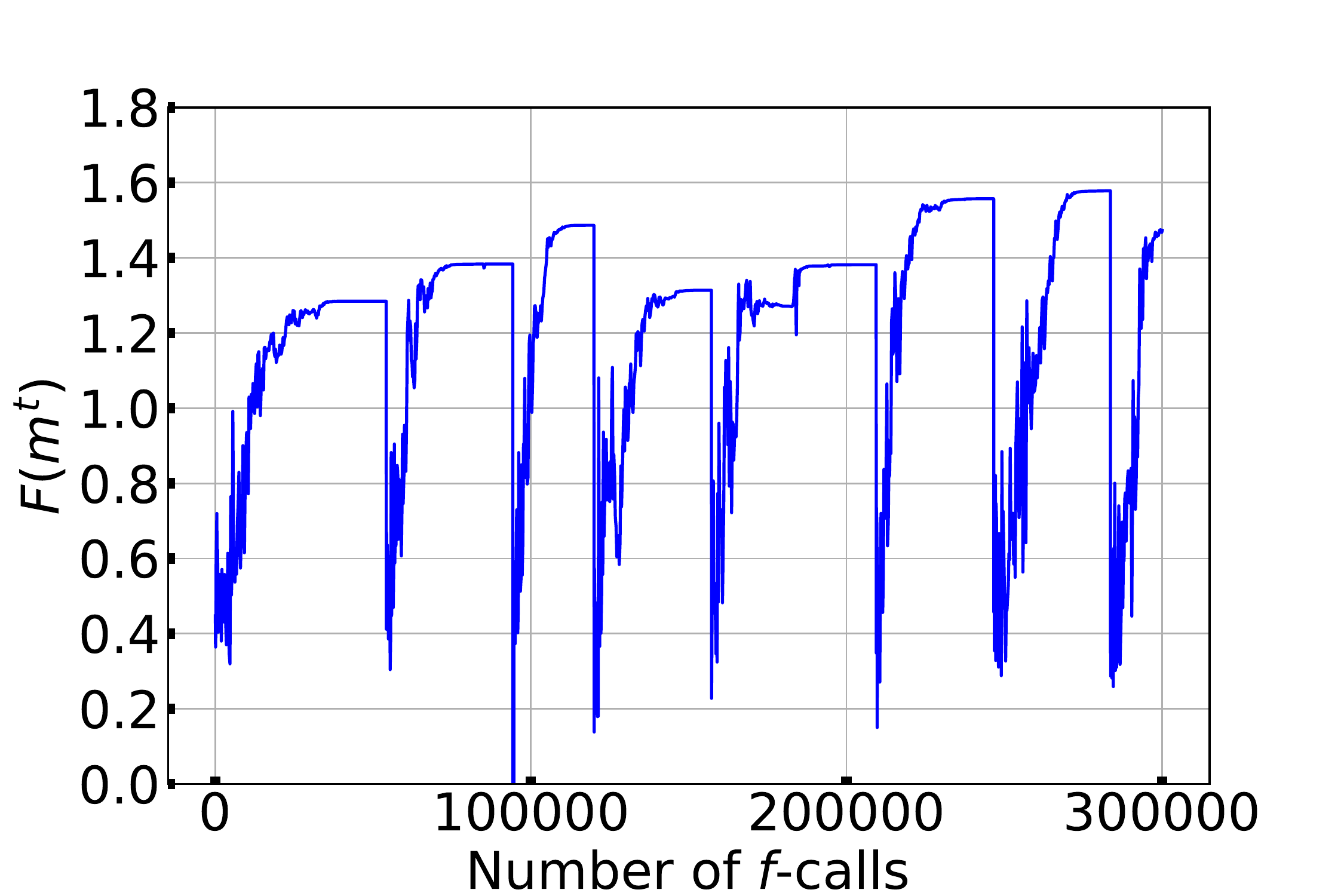}%
    \caption{\pbilcma{}}%    
  \end{subfigure}%
  \begin{subfigure}{\figsizeiwpo\hsize}%
    \includegraphics[width=\hsize]{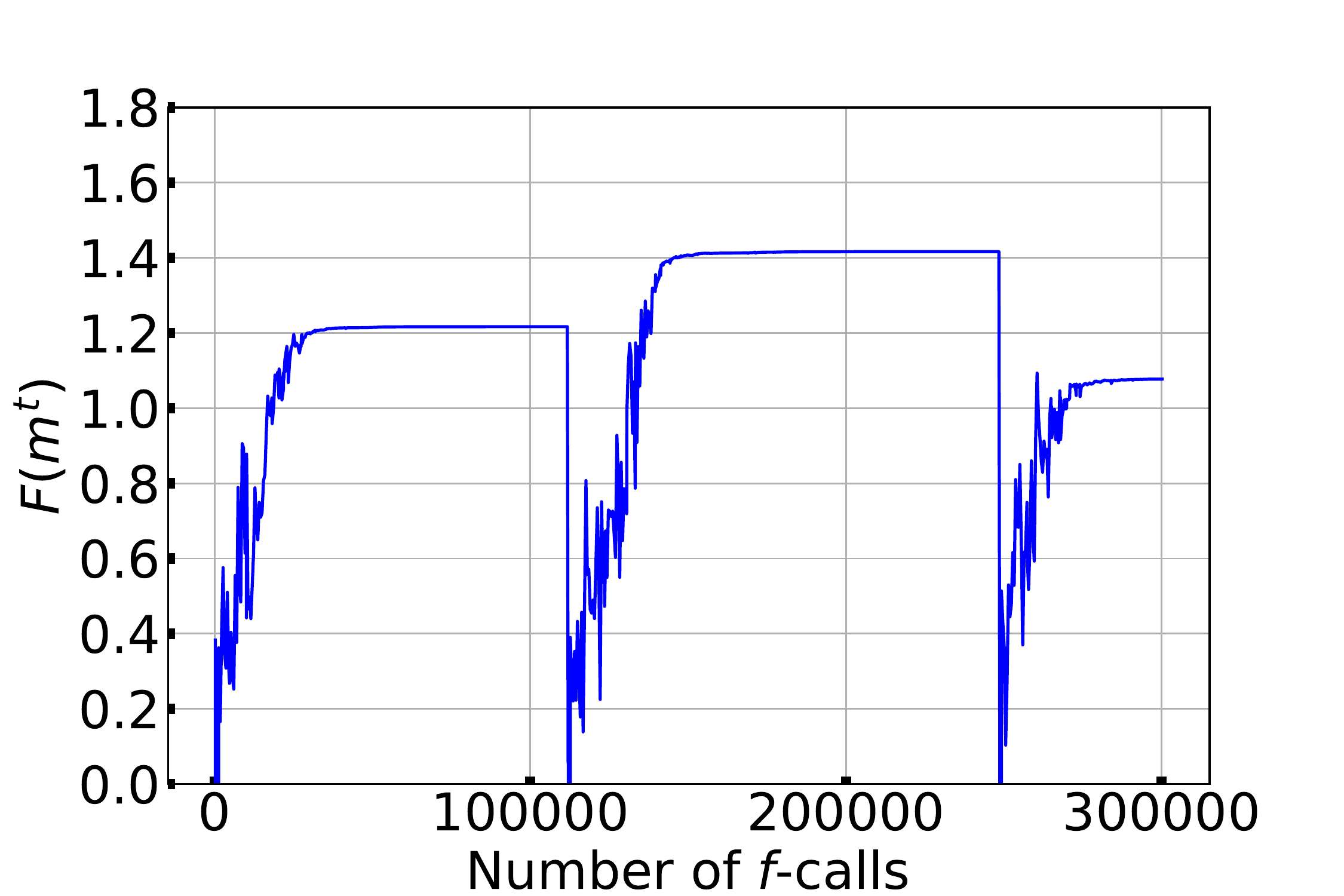}%
    \caption{\texttt{lq-CMA-ES}}%    
  \end{subfigure}%
  \caption{History of $F(m^t)$ in a typical run for each algorithm.}
  \label{fig:IWPO-Fmt}
\end{figure}

\paragraph{\pbilcma{} vs \texttt{lq-CMA-ES}}
\Cref{fig:IWPO-20runs} and \Cref{table:wellplacementsummary} show that \pbilcma{} outperformed \texttt{lq-CMA-ES}.
As shown in \Cref{fig:IWPO-20runs}, \pbilcma{} obtained higher $50$ percentile values than \texttt{lq-CMA-ES} from the beginning of the optimization, and smaller interquartile ranges at $300,000$ $f$-calls. 
Similarly to the advantage of \pbilcma{} over \texttt{CMA-ES}, \pbilcma{} was able to perform more restarts than \texttt{lq-CMA-ES}, whereas \texttt{lq-CMA-ES} performed more restarts than \texttt{CMA-ES} on average. 
That is, \pbilcma{} converged to a local optimum faster than \texttt{lq-CMA-ES} for each restart.
The reason may be twofold. 
First, the ratios $\abs{S_\mathrm{support}(\hat{x})}/m$ around the local optima $\hat{x}$ are sufficiently small for \pbilcma{} to be more efficient than \texttt{lq-CMA-ES}. 
As confirmed in the previous sections, the efficiency of \pbilcma{} over \texttt{CMA-ES} was greater when this ratio was smaller, whereas the efficiency of \texttt{lq-CMA-ES} over \texttt{CMA-EX} was virtually constant.
In this problem, the ratio $\abs{A^t} / m$ decreased by approximately $1/5$ in \pbilcma{}, as shown in \Cref{fig:IWPO-pts}.
Second, the surrogate model inside \texttt{lq-CMA-ES}, which is a linear-quadratic model, may not be suitable for this problem, possibly because of multimodality and non-smoothness. 
As may be noted from \Cref{fig:IWPO-kendall}, the rank correlation between the true worst-case values and the output of the surrogate model tends to be frequently lower than $0$. 
If it is smaller than the predefined threshold, \texttt{lq-CMA-ES} spends $f$-calls to train the surrogate model. 
Therefore, \Cref{fig:IWPO-kendall} indicates that \texttt{lq-CMA-ES} frequently updates the surrogate model by spending $f$-calls, resulting in a slower convergence than \pbilcma{}.

\section{Conclusions}\label{sec:conc}

We targeted the worst-case optimization with a finite scenario set $S=\{1,\cdots, m\}$, and the objective function values were evaluated using computationally expensive numerical simulations. In this study, we focused on reducing the number of simulation executions (referred to as $f$-calls for simplicity) for the objective function value of the problem. The conclusions of this study are summarized as follows.

\begin{enumerate}
 \item The definition of support scenarios $S_\mathrm{support}(H)$ at a neighborhood $H$ was introduced to elucidate the idea of approximating the worst-case objective function without sampling every possible scenario. We designed five test problems in which we could control the number of support scenarios around the optimal solution.
 
 \item We proposed a new optimization algorithm, denoted adaptive scenario subset selection CMA-ES (\pbilcma), which optimizes continuous variables vector $x$ by the CMA-ES while approximating the worst-case objective function $F$ by adaptively subsampling a set of support scenarios in the current search area.

 \item Numerical experiments were conducted on test problems to compare \pbilcma{} with a brute-force approach (\Cref{alg:baseline}) and a surrogate-assisted approach \texttt{lq-CMA-ES}. We confirmed that \pbilcma{} generally outperformed the brute-force approach. Moreover, \pbilcma{} outperformed \texttt{lq-CMA-ES} when the ratio of the number of support scenarios to the total number of scenarios was relatively small (e.g., $< 1/3$). 

 \item The effectiveness of \pbilcma{} was demonstrated on well placement optimization problems. \pbilcma{} was able to obtain a better well placement than \texttt{lq-CMA-ES} and the brute-force approach because of more frequent restarts due to the greater efficiency of \pbilcma{} compared to the other approaches.
 
\end{enumerate}

% \textcolor{blue}{[R1C5] :
In this study, numerical experiments on benchmark problems with various characteristics were not conducted. Five benchmark problems were considered, and the worst-case objective function in all of them was a single-peak function. Therefore, in future work, the performance of the proposed approach should be investigated on benchmark problems whose worst-case objective function is ill-conditioned, multimodal, or has variable dependencies. 
% }

% \textcolor{blue}{[R1C5] :
Another direction of future work is to combine \pbilcma{} and the approaches smoothing the worst-case objective function. As introduced in \Cref{sec:relatedworks}, the worst-case objective function becomes naturally non-smooth owing to its construction. 
We expect that the CMA-ES for minimization can become more efficient by smoothing the worst-case objective function obtained by \emph{AS3}.  
% }

\section*{Acknowledgements}
This work is partially supported by JSPS KAKENHI Grant Number 19H04179.

\appendix

\section{Sensitivity Analysis}\label{app:sensitivity}

The sensitivities of \pbilcma{} on $\eta$, $c_p$, $p^0$, and $\epsilon$ were investigated to demonstrate the effect of these hyperparameters and the robustness of the proposed approach. Unless otherwise specified, we followed the experimental setting described in \Cref{sec:common}, except for the maximum number of $f$-calls, which was set to $2 \times 10^6$ in this analysis.

The test problems were set as follows. 
$n = 10$ for all problems, 
$K = \{5, 10, 15, 25, 50, 75, 100\}$ and $m = 100$ for P1 and P2,
$m = \{20, 40, 80, 120, 160, 180, 200\}$ (hence, $K = \{1, 2, 4,6,8,9, 10 \}$) for P3, 
$L =\{ 5, 10, 15, 25, 50, 75, 100 \}$ and $m = 100$ for P4. 
They were set to test the performance of \pbilcma{} on different ratios $S_\mathrm{support}(x^*) / m$. 

\subsection{Sensitivity to $\eta$}\label{sec:eta}

A higher $\eta$, and hence a higher $c_n$, is expected to require fewer $f$-calls to decrease $p_s^t$ for $s \notin S_\mathrm{support}(H^t_{\gamma})$. However, a higher $\eta$ has a risk of not increasing $p_s^t$ for $ s \in S_\mathrm{support}(H_{\gamma}^t)$, which does not satisfy condition \eqref{eq:p_condition}. 

The results on P3 and P4 are shown in \Cref{fig:dis1-eta}. 
We observed a tendency on P1 and P2 similar to that observed on P3, and hence they are omitted.

The results for P3 show that a higher $\eta$ converges faster when $\abs{S_\mathrm{support}(x^*)}/m$ is relatively small, such as $0.1$. A higher $\eta$ contributes to a faster decrease in $p^t_s$ for each $s \notin S_\mathrm{support}(H^t_{\gamma})$ to avoid sampling unnecessary scenarios. 
However, when $\abs{S_\mathrm{support}(x^*)}/m$ on P3 increases, a higher $\eta$ requires more $f$-calls. Additionally, setting $\eta \geq 0.7$ led to the optimization failure when $\abs{S_\mathrm{support}(x^*)}/m = 1.0 $. This occurred because a high $\eta$ decreased the number of support scenarios whose $p^t_s$ was kept at a relatively high value. As a result, \pbilcma{} failed to sample a sufficient  number of support scenarios to approximate $F$. 
For example, on P3 with $\abs{S_\mathrm{support}(x^*)} = m = 20$, the expected number of sampled scenarios, i.e., $\E[\abs{A^t}] = \sum^m_{s=1}p^t_s$, was maintained at approximately $16$ with $\eta = 0.1$, whereas it was maintained at approximately $8$ with $\eta = 0.9$.

\renewcommand{\figsizefuncr}{0.5}
\begin{figure}[t]
\centering
  \begin{subfigure}{0.485\hsize}%
    \includegraphics[width=\hsize]{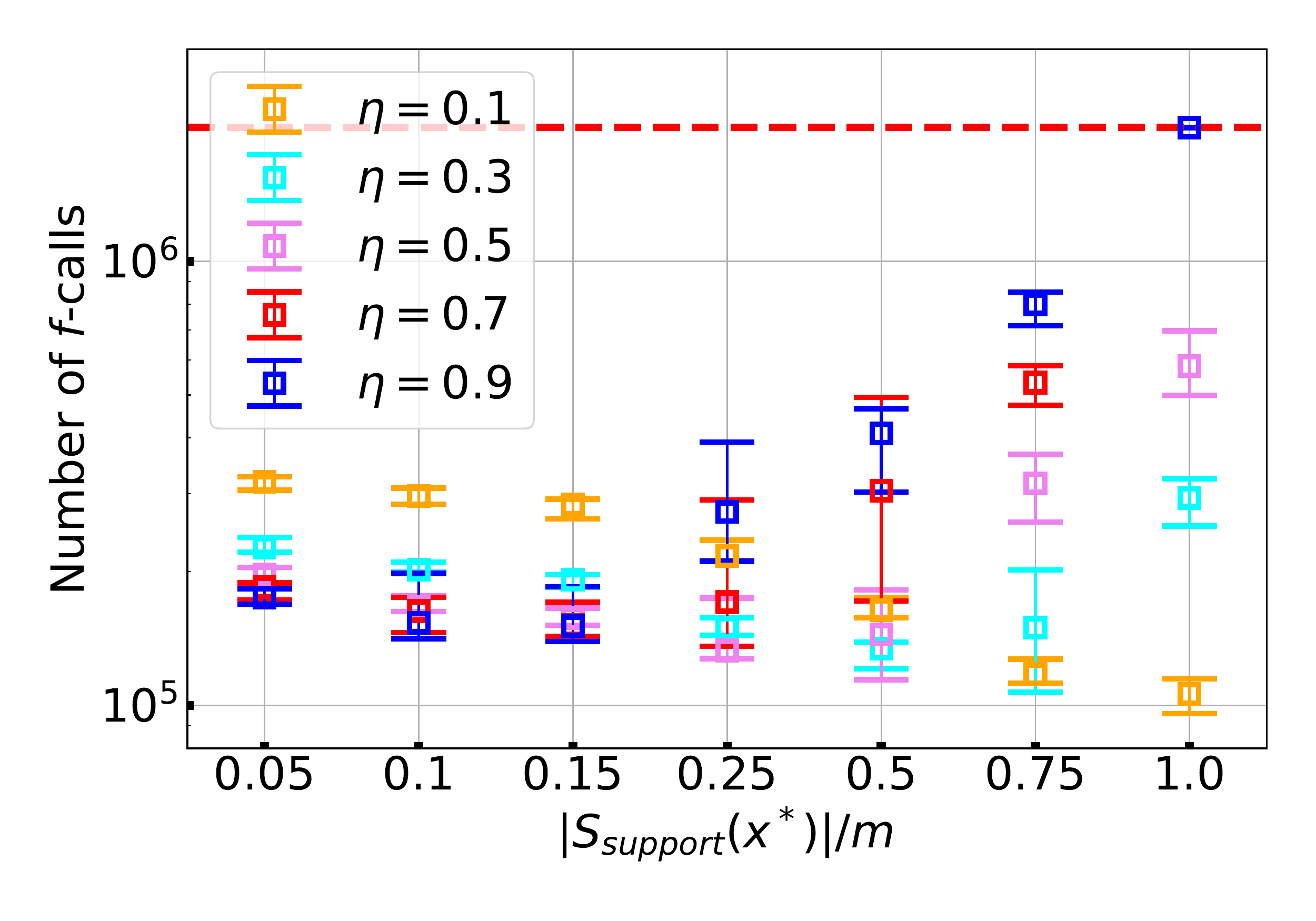}%
    \caption{P3}%    
  \end{subfigure}%
  \begin{subfigure}{\figsizefuncr\hsize}%
    \includegraphics[width=\hsize]{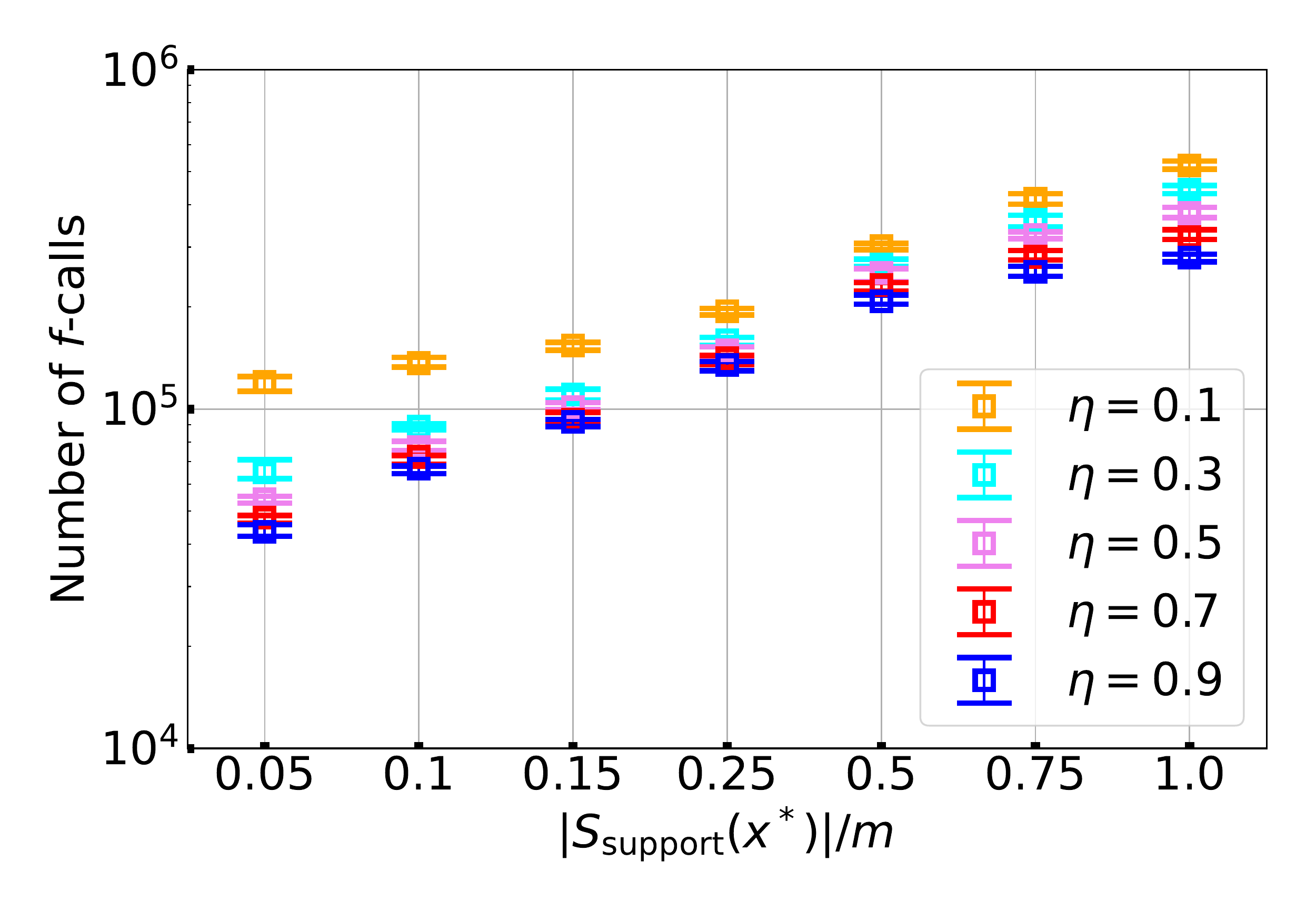}%
    \caption{P4}%    
  \end{subfigure}%   
  \caption{Mean and standard deviation of the number of $f$-calls over 20 trials obtained from \pbilcma{} with various $\eta$ on P3 and P4. The number of $f$-calls in all cases with $\eta= 0.7, 0.9$ on $P3$ reached $2\times10^6$, and they were evaluated as optimization failures in this experiment.}
  \label{fig:dis1-eta}
\end{figure}

By contrast, on P4, \pbilcma{} with a higher $\eta$ could successfully determine the optimal solution for all problem instances, and the number of $f$-calls was smaller. 
This is attributed to the characteristics of P4 and the comparison-based nature of CMA-ES.
In contrast to P1--P3 and P5, we can select a subset $A \subset S_\mathrm{support}(x^*)$ such that $\min_{x \in \X} F(x; A) = \min_{x \in \X} F(x)$ with $\abs{A} = 2$ or $3$. If $L$ is even or odd, $\abs{A} = 2$ or $3$, respectively.  
That is, even if $\abs{A^t} \ll \abs{S_\mathrm{support}(x^*)}$, it was possible to locate the optimum of $F$ by solving $\min_{x \in \X} F(x; A)$, whereas $F(x)$ was not necessarily approximated well by $F(x; A^t)$. 
For example, on P4 with $m = L = \abs{S_\mathrm{support}(x^*)} = 100$, we observed that $\E[\abs{A^t}] = \sum^m_{s=1}p^t_s$ was maintained at approximately 40--50, which is smaller than half of $\abs{S_\mathrm{support}(x^*)}$, while the Kendall's $\tau$ between $F(x)$ and $F(x; A^t)$ computed for solution candidates generated at each iteration was maintained at nearly one during the optimization. 
We consider that the characteristics of P4 were unusual, and note that setting $\eta$ to a small value is advisable in general. 

\subsection{Sensitivity to $c_p$}

A higher $c_p$ is expected to result in a faster adaptation of $p_s^t$, leading to faster convergence. However, $p^t_s$ for a scenario $s \notin S_\mathrm{support}(x^*)$ also increases, resulting in spending more $f$-calls.  
Another impact of a higher $c_p$ is a higher $c_n$ because of \eqref{eq:cn}, and we have already discussed the sensitivity of $c_n$ in \ref{sec:eta}. 

To distinguish the effect of $c_p$ from the effect of $c_n$, we set $c_n =0.003$, which is the value when we set $\eta = 0.1$ and $c_p = 0.3$ for P1, P2, and P4 in this analysis.  

The results are presented in \Cref{fig:dis1-cpcn}. The results for P2 and P3 showed a tendency similar to that observed for P1; hence, they are omitted.
We observed that a smaller $c_p$ converges faster if $\abs{S_\mathrm{support}(x^*)} / m \lessapprox 1/2$. 
In the case of $\abs{S_\mathrm{support}(x^*)} / m \gtrapprox 1/2$, a greater $c_p$ tended to converge faster on P1--P3 and P5, whereas a smaller $c_p$ was better on P4. 
However, the differences in the number of $f$-calls made for $c_p = 0.01$, $0.1$, $0.3$, and $1.0$ were at most a factor of $2$ for all cases in the experiments. Therefore, we conclude that the performance of \pbilcma{} is not sensitive to $c_p$ value.

\providecommand{\figsizefuncr}{0.5}
\begin{figure}[t]
  \centering
  \begin{subfigure}{\figsizefuncr\hsize}%
    \includegraphics[width=\hsize]{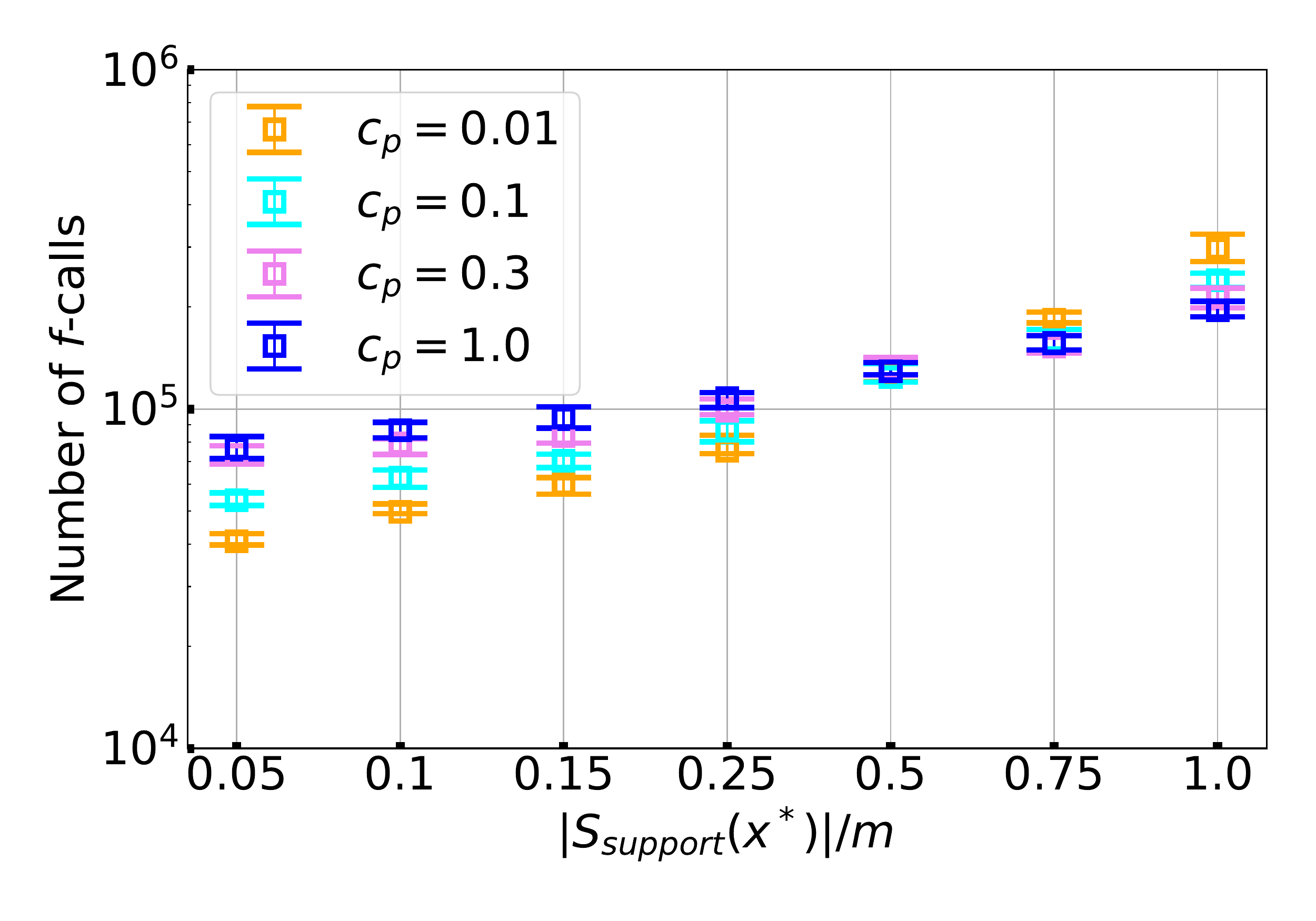}%
    \caption{$P1$}%    
  \end{subfigure}%
  \begin{subfigure}{\figsizefuncr\hsize}%
    \includegraphics[width=\hsize]{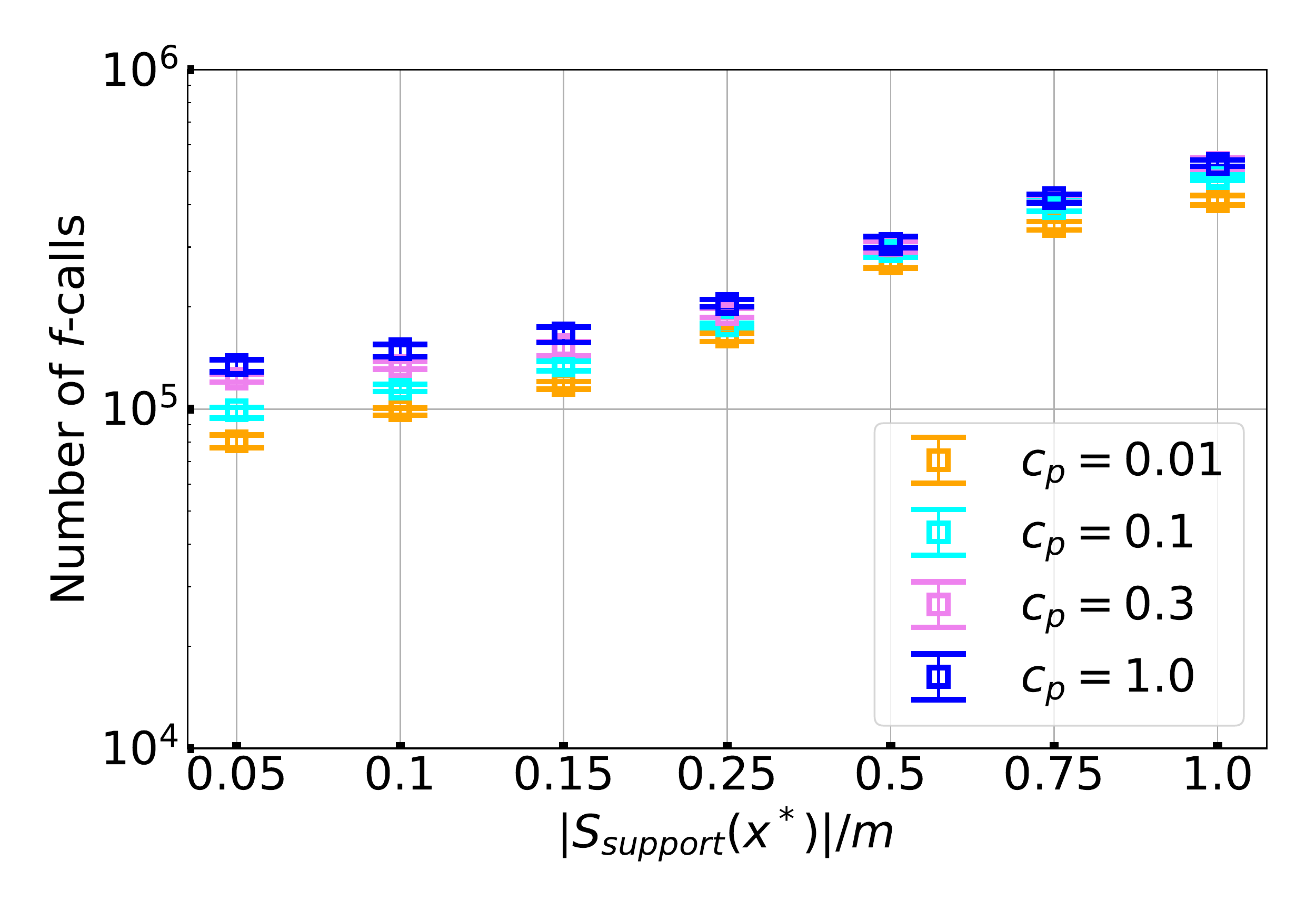}%
    \caption{P4}%    
  \end{subfigure}%   
  \caption{Mean and standard deviation of the number of $f$-calls over 20 trials obtained from \pbilcma{} with various $c_p$.
}
  \label{fig:dis1-cpcn}
\end{figure}

\subsection{Sensitivity to $p^0$}

\begin{figure}[t]
  \centering
  \begin{subfigure}{0.48\hsize}%
    \includegraphics[width=\hsize]{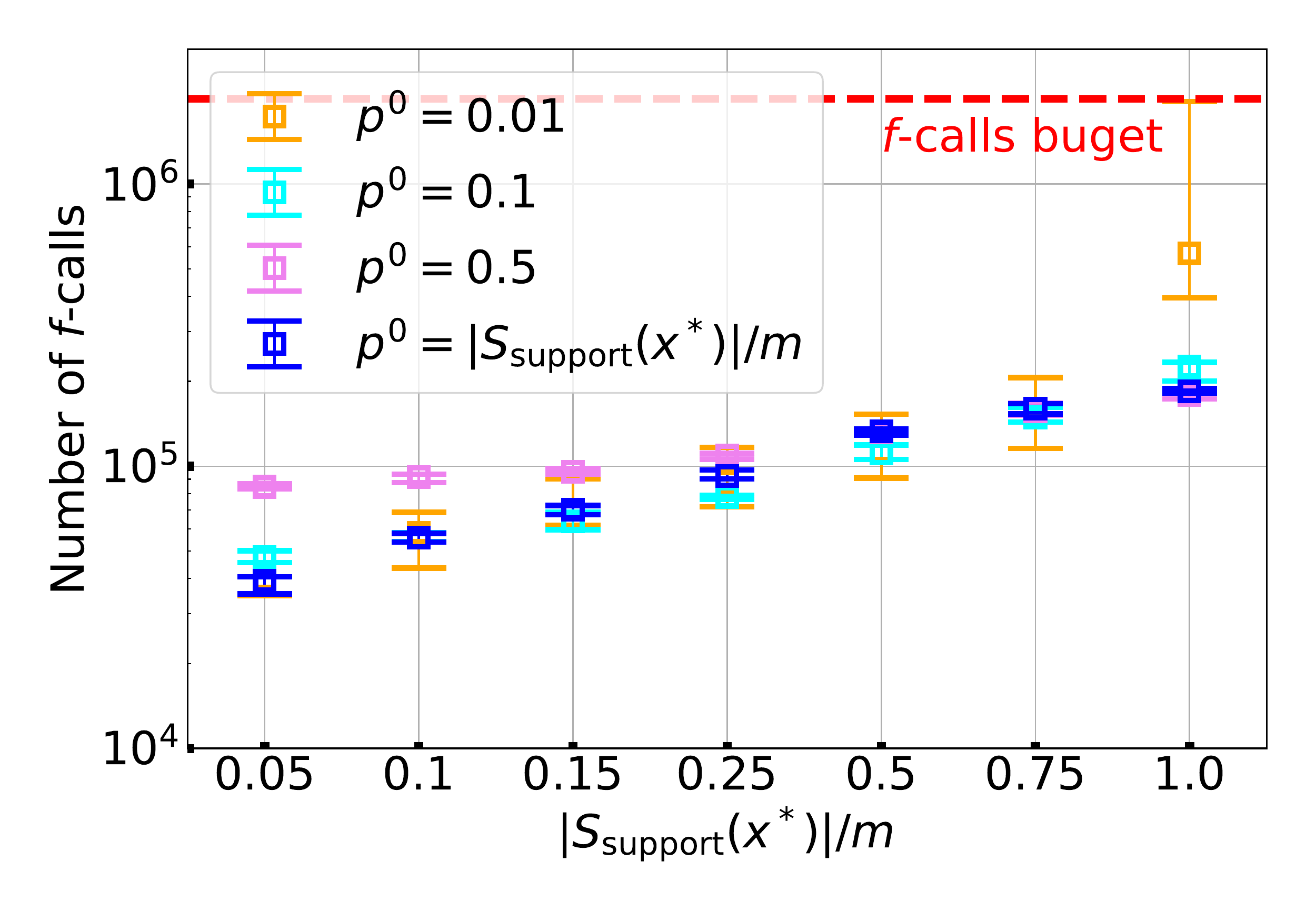}%
    \caption{P1}%    
  \end{subfigure}%
  \begin{subfigure}{0.5\hsize}%
    \includegraphics[width=\hsize]{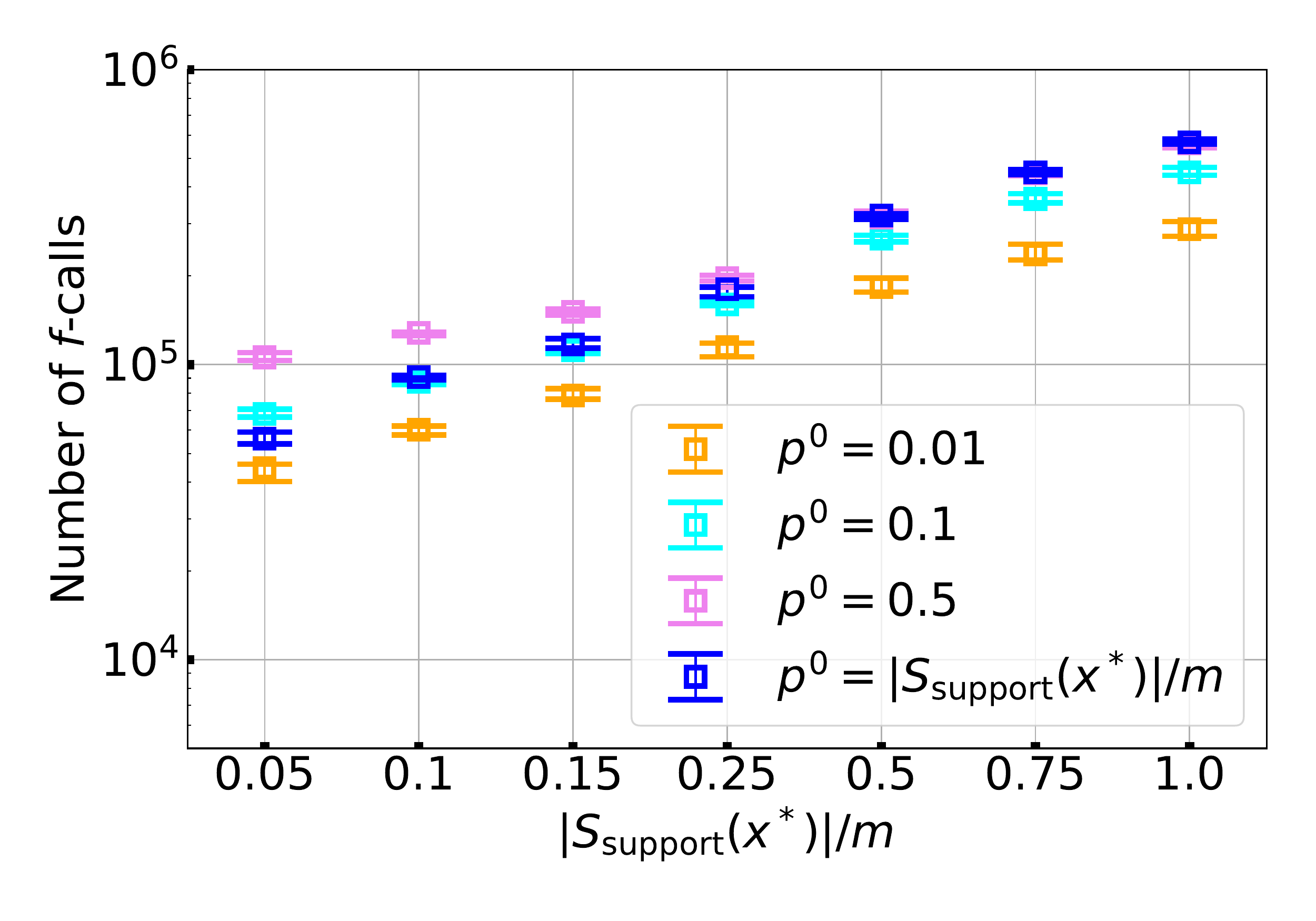}%
    \caption{P4}%    
  \end{subfigure}%   
  \caption{Mean and standard deviation of the number of $f$-calls over 20 trials obtained from \pbilcma{} with various $p^0$ on P1. Cases with $f$-calls reaching $2\times10^6$ were evaluated as optimization failures in this experiment.}
  \label{fig:dis4-1}
\end{figure}

The impact of the initial $p_s^{0}$ was investigated. 
The results are presented in \Cref{fig:dis4-1}. The results for P2 and P3 showed a tendency similar to that observed for P1; hence, they are omitted.

We observed the tendency that a smaller $p_s^0$ converges faster if $\abs{S_\mathrm{support}(x^*)} / m$ is small, whereas a larger $p_s^0$ converges faster if $\abs{S_\mathrm{support}(x^*)} / m$ is large for P1--P3 and P5. This is advantageous if $\ind{s \in S_\mathrm{support}(H_\gamma^0)}$ is approximated by the initial $p_s^0$, as it does not need to adapt $p_s^t$ at the beginning, thereby minimizing $f$-calls. If $p_s^0$ is set to a greater value, it requires more $f$-calls to decrease $p_s^t$ for non-support scenarios. If $p_s^0$ is set to a smaller value, more $f$-calls are required to increase $p_s^t$ for support scenarios. The experimental results reflect these expectations. On P4, we observed that a smaller $p_s^0$ resulted in a faster convergence. This is due to the characteristics of P4, as discussed above. 

We note that an excessively small $p_s^0$ value sometimes leads to optimization failure. 
On P1 with $m = \abs{S_\mathrm{support}(x^*)}$, \pbilcma{} with $p_s^0 = 0.01$ failed to converge. We observed divergent behavior in the search distribution in this situation, in which $\norm{m^t - x^*}$ increased to $10^{11}$ at the beginning of the search. This is possibly because the landscape of $F(x; A^t)$ changed drastically at each iteration. Therefore, it is safer to set $p_s^0$ to a relatively high value, although doing so may reduce efficiency.

\providecommand{\figsizefuncr}{0.5}
\begin{figure}[t]
  \centering
  \begin{subfigure}{\figsizefuncr\hsize}%
    \includegraphics[width=\hsize]{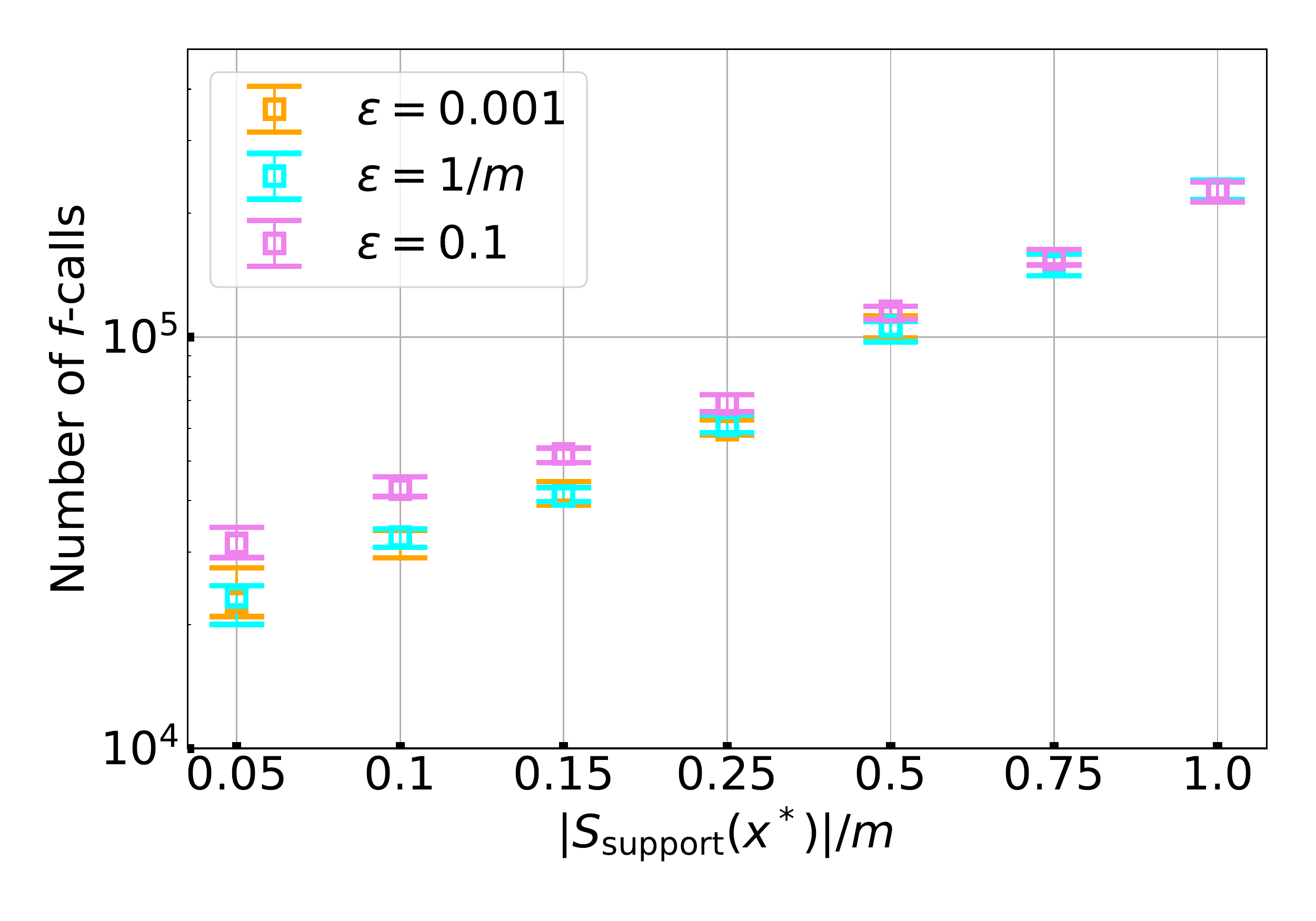}%
    \caption{P2}%    
  \end{subfigure}%
%  \\
  \begin{subfigure}{\figsizefuncr\hsize}%
    \includegraphics[width=\hsize]{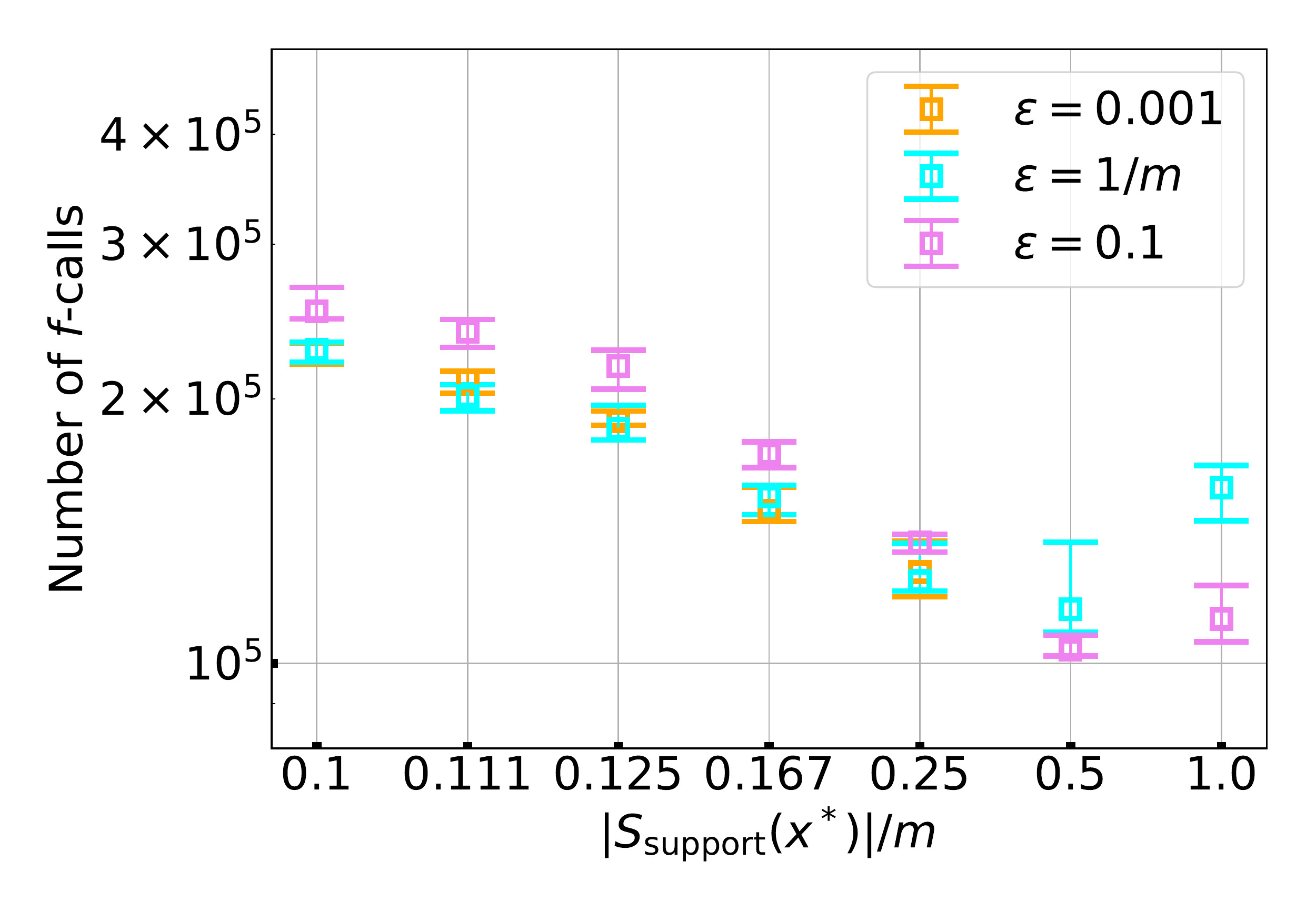}%
    \caption{P3}%    
  \end{subfigure}%
  \caption{Mean and standard deviation of the number of $f$-calls over 20 trials obtained from \pbilcma{} with various $\epsilon$ on P2 and P3. No trials with $\epsilon = 0.001$ were successful when $\abs{S_\mathrm{support}(x^*)}/m \geq 0.5$ on P3.
  }
  \label{fig:dis1-epsilon}
\end{figure}

\subsection{Sensitivity to $\epsilon$}\label{sec:epsilon}

The minimal probability $\epsilon$ is introduced to prevent $p_s^t$ from converging to $0$, resulting in the algorithm not sampling $s$ forever. With $\epsilon$, all scenarios are guaranteed to be sampled every $1/\epsilon$ iterations in expectation. However, because the expected number of sampled scenarios is $\sum_{s=1}^{m} p_s^t \geq \epsilon \cdot m$, this limits the upper bound of the speed-up factor over the brute-force approach. In this study, we investigated the impact of $\epsilon$.

The results are shown in \Cref{fig:dis1-epsilon}. The results for P1 and P4 showed a tendency similar to that observed on P2; hence, they are omitted. 

We observed that a small $\epsilon$ required fewer $f$-calls when $\abs{S_\mathrm{support}(x^*)}/m$ was relatively small at P1--P4, as a small $\epsilon$ allows $\sum_s^m p^t_s$ to be as small as $\abs{S_\mathrm{support}(x^*)}$ at the convergence, resulting in fewer $f$-calls.
We note that an excessively small $\epsilon$ led to failure in some problem instances. On P3, \pbilcma{} with $\epsilon = 0.001$ failed to converge at $x^*$ when $\abs{S_\mathrm{support}(x^*)}/m \geq 0.5$.
Therefore, we advise setting $\epsilon$ to a relatively high value, while the efficiency of \pbilcma{} may be lost.

\section{Scalability Analysis}\label{app:scal}

The efficiency of \pbilcma{} for problems with higher $n$ and $m$ than those in \Cref{sec:exp} was investigated.
Unless otherwise specified, we followed the experimental setting described in \Cref{sec:common}, except for the maximum number of $f$-calls, which was  set to $2 \times 10^7$ in this analysis.

\begin{figure}[t]
\begin{minipage}{0.45\textwidth}
  \centering
    \includegraphics[width=\hsize]{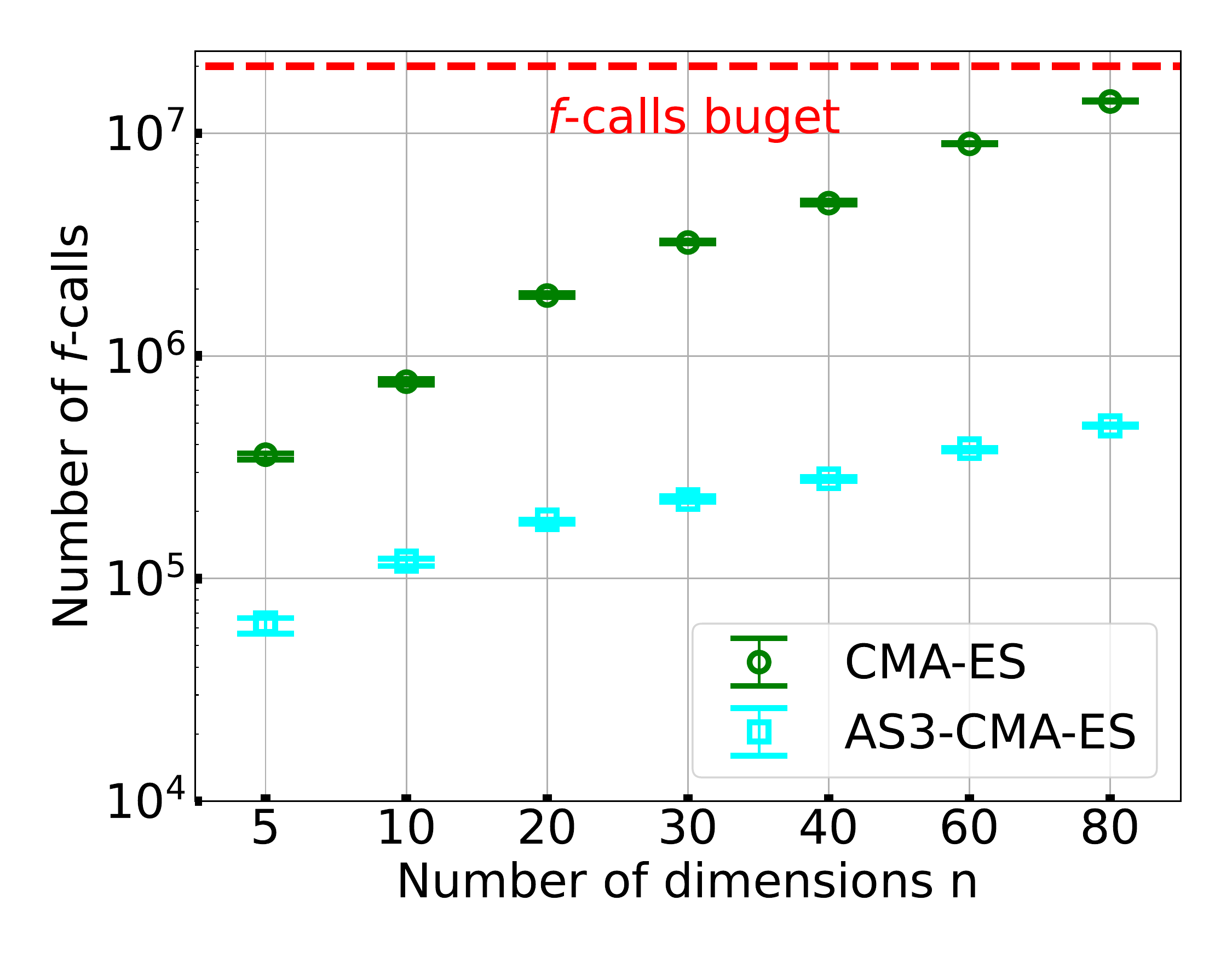}%
  \caption{Mean and standard deviation of the number of $f$-calls  over 20 trials with various $n$ on P4.}%
  \label{fig:Nsensitivity}
\end{minipage}
\begin{minipage}{0.55\textwidth}
  \centering
  \begin{subfigure}{0.5\hsize}%
    \centering%
    \includegraphics[width=\hsize]{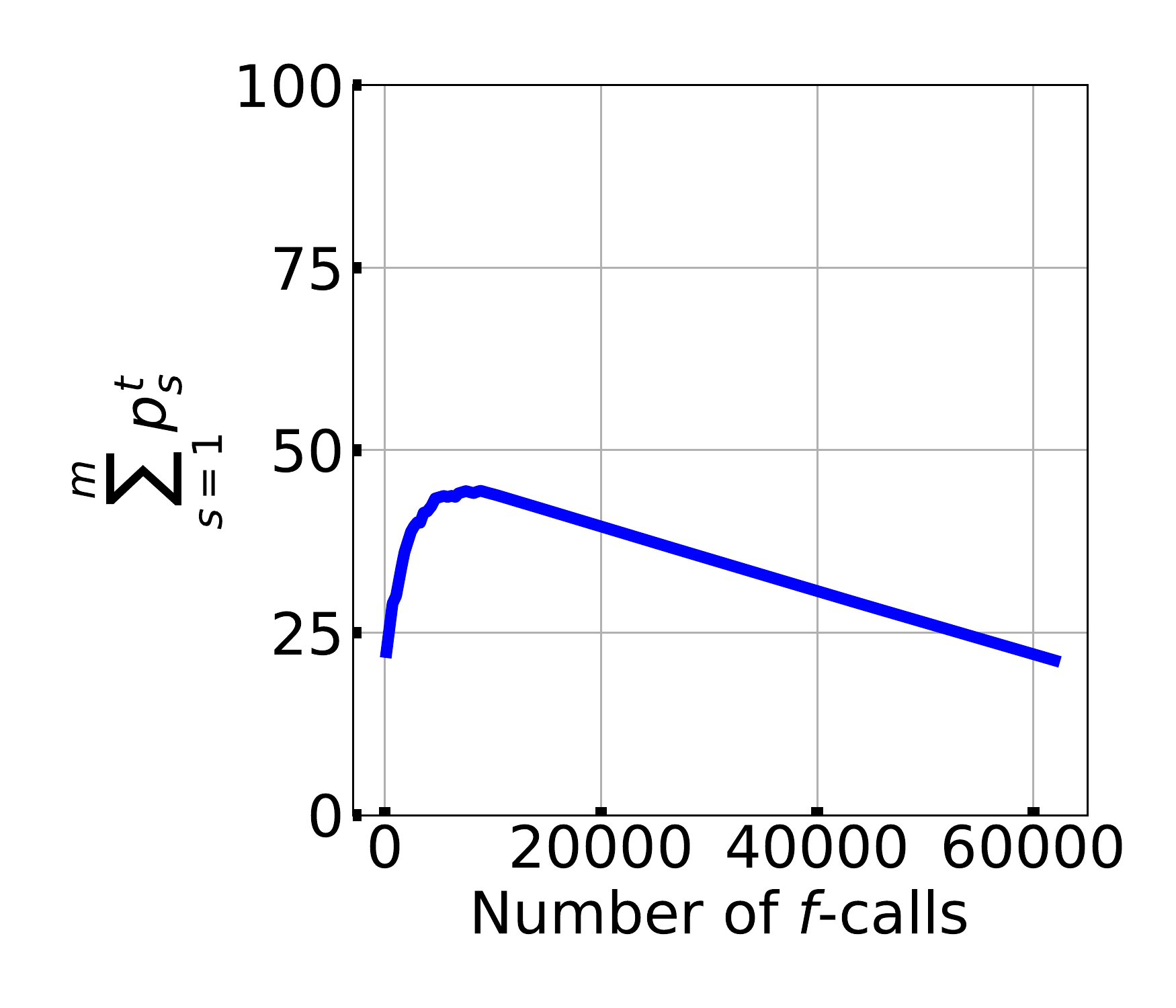}%
    \caption{$n=5$}%
  \end{subfigure}%
  \begin{subfigure}{0.5\hsize}%
    \centering%    
    \includegraphics[width=\hsize]{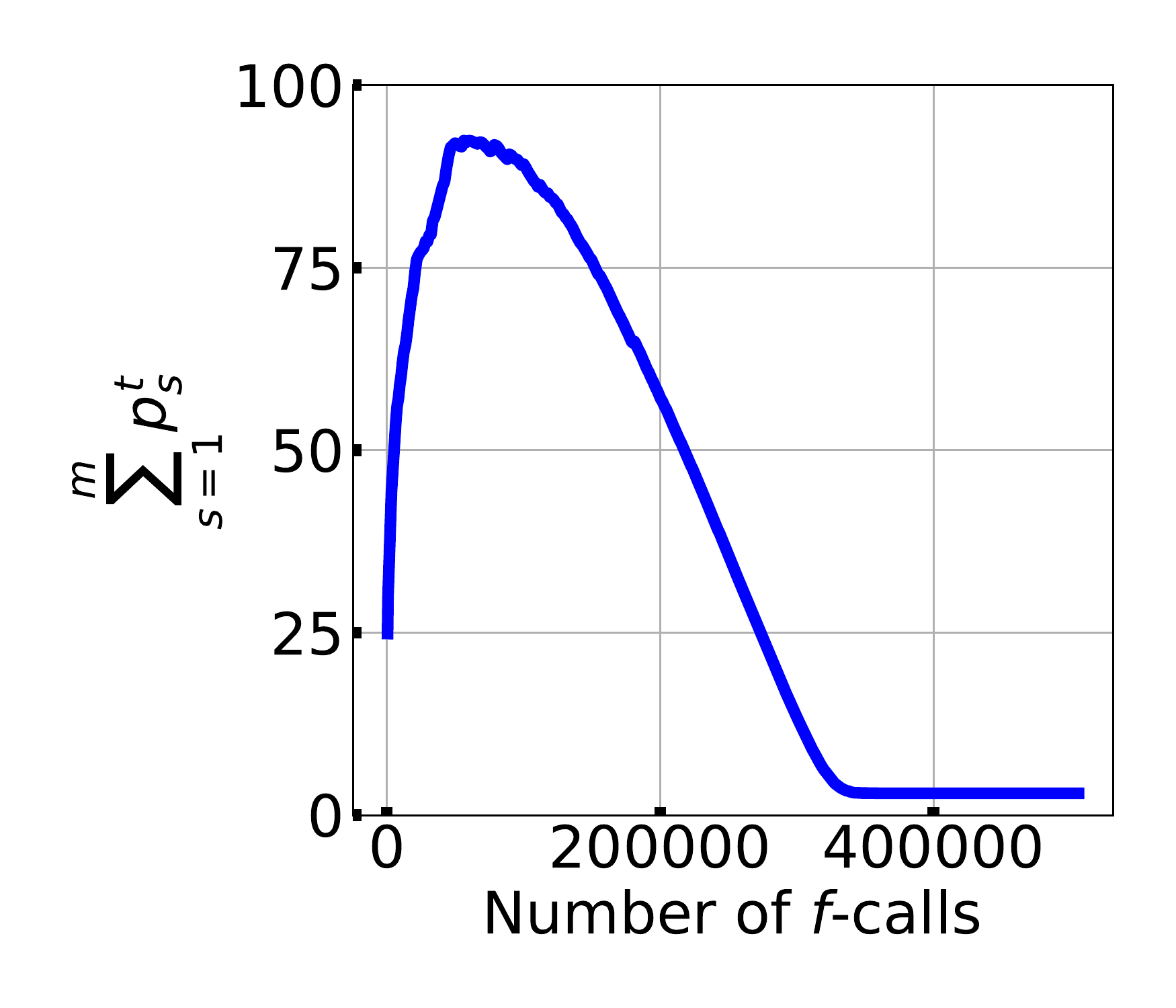}%
    \caption{$n=80$}%
  \end{subfigure}%
  \caption{History of $\sum_{s=1}^m p^t_s$ on a typical run on P4 ($L=2, m=200$)
  }%
\label{fig:Nsensitivity-typical}
\end{minipage}
\end{figure}

\subsection{Scalability to $n$}

To show the effect of $n$ and the robustness of \pbilcma{}, we conducted a scalability analysis of $n$. 
For this analysis, we applied \texttt{CMA-ES} and \pbilcma{} to P1--P5 in the following problem settings to analyze the scalability of $n$. We set, $n = \{5, 10, 20, 30, 40, 60, 80\}$, and $K = 2$ at $m = 200$ for $P1$ and $f_2$, and $m = 200$ (hence, $K = \{20, 10, 5, 4, 3, 2, 2\}$) for $f_3$,  $L = 2$ at $m = 200$ for $f_4$ and $m = 200$ for $f_5$.

The results are presented in \Cref{fig:Nsensitivity}. The results for P1--P3 and P5 showed a tendency similar to that observed on P4; hence, they are omitted.
As \Cref{fig:Nsensitivity} shows, both algorithms showed increased numbers of $f$-calls for convergence with increasing $n$. 
The efficiency of \pbilcma{} over \texttt{CMA-ES} in terms of the number of $f$-calls was at most a factor of $20$ for all the cases.  

The results show that the efficiency of \pbilcma{} over \texttt{CMA-ES} was higher at a higher $n$. 
This is because \pbilcma{} on the problem with $n=80$ spent sufficient $f$-calls to learn $p^t_s$ for each $s \in S$ when approaching $x^*$. 
We confirmed that \pbilcma{} on the problem with $n=80$ successfully determined the optimum solution while maintaining $\sum_s^m p^t_s \approx \abs{S_\mathrm{support}(x^*)}$.
\Cref{fig:Nsensitivity-typical} shows the history of $\sum_s^m p^t_s$ resulting from a typical run with $n = 5$ and $80$ on P4.
The expected number of sampled scenarios, i.e., $\E[\abs{A^t}] = \sum^m_{s=1}p^t_s$, was decreased to $\abs{S_\mathrm{support}(x^*)}=2$ when $n = 80$, whereas it was more than $\abs{S_\mathrm{support}(x^*)}$ ($\sum_s^m p^t_s \approx 25$ in the end) in the case of $n=5$. 

\subsection{Scalability to $m$}

The number of $f$-calls spent by \pbilcma{} depends on the number of support scenarios $\abs{S_\mathrm{support}(x^*)}$, whereas \texttt{CMA-ES} and \texttt{lq-CMA-ES} linearly increases the number of $f$-calls if the number of scenarios $m$ increases.
Therefore, if $\abs{S_\mathrm{support}(x^*)}$ is fixed, \pbilcma{} is expected to be more efficient than \texttt{CMA-ES} and \texttt{lq-CMA-ES} for a larger $m$.   
However, if $m$ is larger, \pbilcma{} will spend more $f$-calls to adapt $p^t_s$ for each $s \in S$ to $\ind{s \in S_\mathrm{support}(H_{\gamma}^t)}$. 
On the other hand, if the ratio $\abs{S_\mathrm{support}(x^*)}/m$ is fixed, the efficiency of \pbilcma{} is expected to be even at a higher $m$.  
 
To demonstrate the scalability to $m$ under a fixed $\abs{S_\mathrm{support}(x^*)}$, we applied \texttt{CMA-ES} , \texttt{lq-CMA-ES}, and \pbilcma{} to P1--P5.
We set, $n = 10$ and $m = \{20, 40, 80, 120, 160, 200, 240, 280, 320, 360, 400\}$ (hence, $K = \{2, 4, 8, 12, 16, \\ 20, 24, 28, 32, 36, 40\}$) for $f_3$, and $m = \{ 10, 20, 40, 80, 120, 160, 200, 240, 280, 320, 360, 400\}$ for other problems. The number $\abs{S_\mathrm{support}(x^*)}$ of the support scenarios was $20$ for P3 and $2$ for the others.
To demonstrate the scalability to $m$ under a fixed $\abs{S_\mathrm{support}(x^*)} / m$, 
we applied \texttt{CMA-ES}, \texttt{lq-CMA-ES}, and \pbilcma{} to P1 and P2, respectively.
We set, $n = 10$ and $m = \{40, 80, 120, 160, 200, 240, 280, 320, 360, 400\}$ for P1 and P2. We fixed the ratio $\abs{S_\mathrm{support}(x^*)}/m$ at $0.05$. 

\begin{figure}[t]
\centering
  \begin{minipage}{0.45\textwidth}
  \centering
  \begin{subfigure}{\hsize}%
    \includegraphics[width=\hsize]{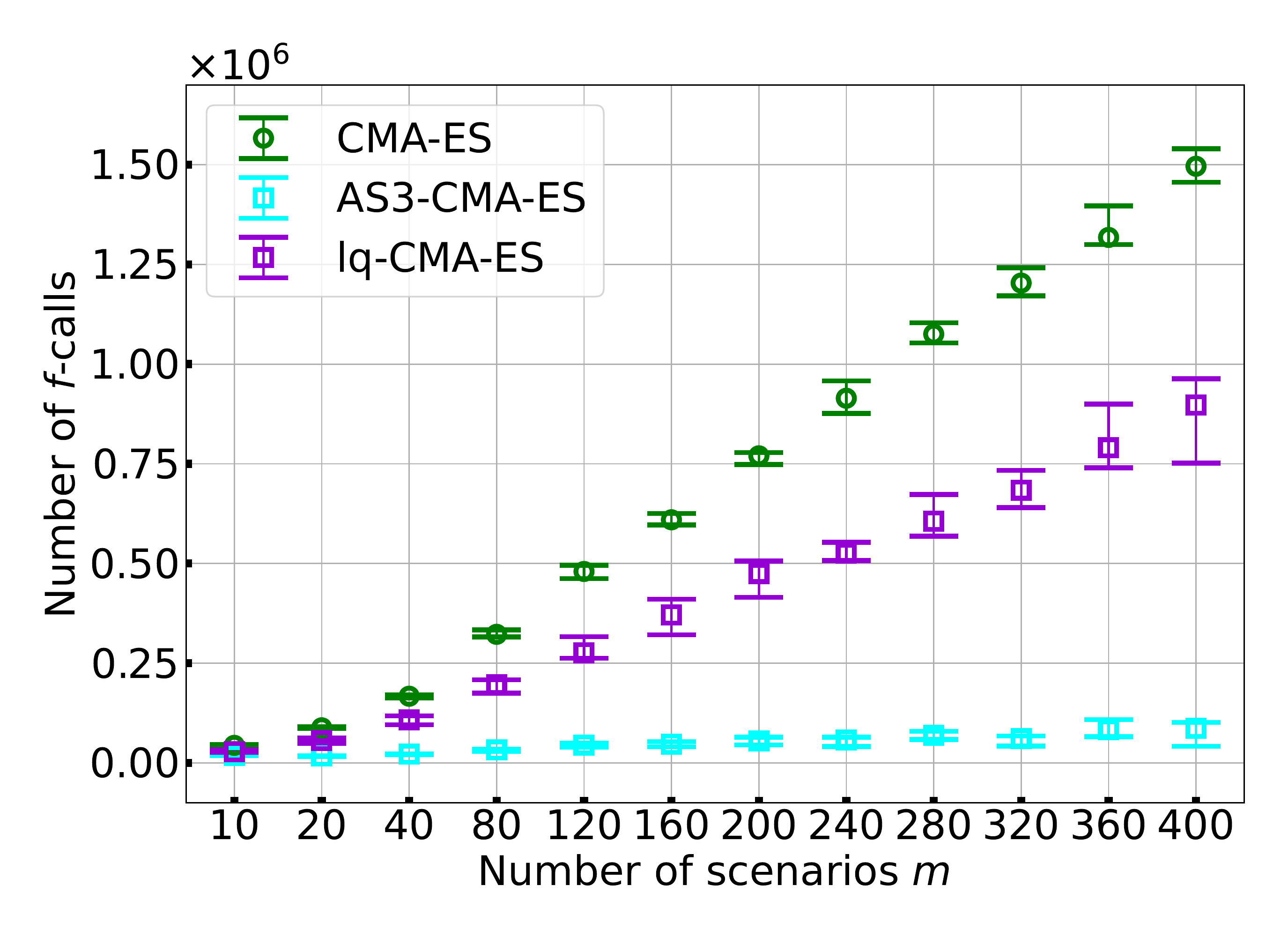}%
  \end{subfigure}%  
  \caption{Mean and standard deviation of the number of $f$-calls over 20 trials with various $m$ and fixed $\abs{S_\mathrm{support}(x^*)}$ on P4.}
  \label{fig:msensitivity}
\end{minipage}
\quad
\begin{minipage}{0.45\textwidth}
  \centering
    \includegraphics[width=\hsize]{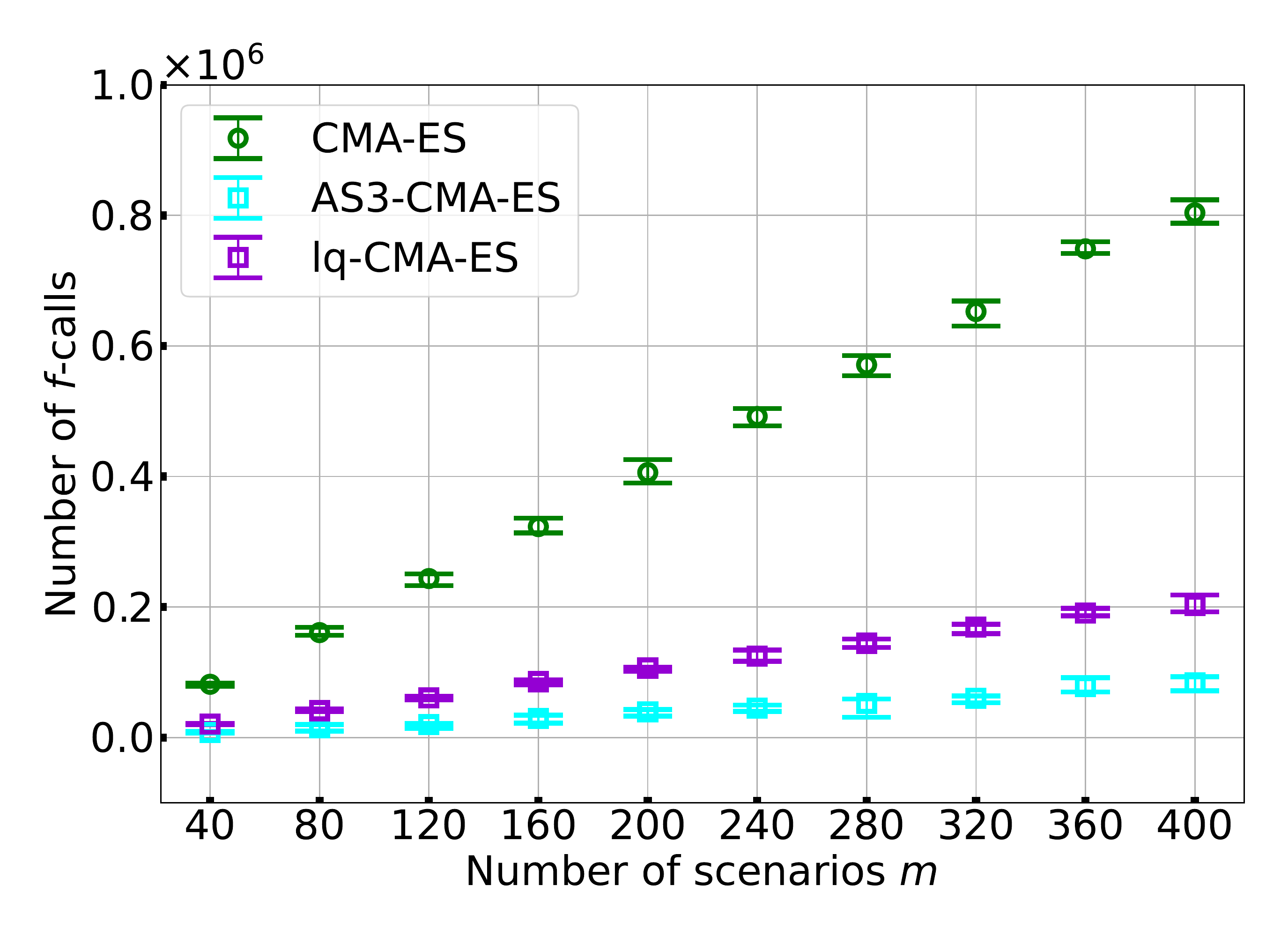}%
  \caption{Mean and standard deviation of the number of $f$-calls over 20 trials with various $m$ and fixed $\abs{S_\mathrm{support}(x^*)}/m$ on P2.}
  \label{fig:msensitivity2}
\end{minipage}
\end{figure}

The scalability to $m$ under a fixed $\abs{S_\mathrm{support}(x^*)}$ is shown in \Cref{fig:msensitivity}. The results for P2--P5 showed a tendency similar to that observed on P1; hence, they were omitted. The number of $f$-calls spent by \texttt{CMA-ES} and \texttt{lq-CMA-ES} increased linearly with increasing $m$, whereas \texttt{lq-CMA-ES} required fewer $f$-calls than \texttt{CMA-ES}. 
On the other hand, \pbilcma{} increased the number of $f$-calls; however, the increment in $f$-calls was less than that of \texttt{CMA-ES} and \texttt{lq-CMA-ES}. Although $S_\mathrm{support}(x^*)$ was fixed, \pbilcma{} spent more $f$-calls by increasing $m$. We considered that more $f$-calls were spent for the adaptation of $p^t_s$ for each $s \in S$ when $m$ was set at a higher value.   
\Cref{fig:msensitivity} show that the efficiency of \pbilcma{} over \texttt{CMA-ES} and \texttt{lq-CMA-ES} was improved with increasing $m$, if $S_\mathrm{support}(x^*)$ was fixed.

The scalability to $m$ under a fixed $\abs{S_\mathrm{support}(x^*)} / m$ is shown in \Cref{fig:msensitivity2}. The results for P1 showed a tendency similar to that observed for P2; hence, they were omitted. The efficiency of \pbilcma{} over \texttt{CMA-ES} was maintained until $m=400$. By contrast, the number of $f$-calls spent by \texttt{lq-CMA-ES} was close to that of \pbilcma{}. As shown in \Cref{fig:ssensitivity}, \texttt{lq-CMA-ES} showed a high efficiency for P2. We consider that it is easy for \texttt{lq-CMA-ES} to build a proper surrogate model for P2.

We observed that the number of $f$-calls spent by \pbilcma{} depends on the number of support scenarios $\abs{S_\mathrm{support}(x^*)}$ rather than the number of scenarios $m$, and its efficiency was maintained until $m=400$. The maximum ratio of the number of $f$-calls spent by \pbilcma{} and \texttt{CMA-ES} was approximately $10$ and that spent for \pbilcma{} and \texttt{lq-CMA-ES} was approximately $6$ for all cases in this experiment.

\section{Effect of $\lambda_s$ in \pbilcma{} with fixed $\lambda_s$}\label{sec:lambdas}

The hyperparameter $\lambda_s$, which is the number of subsampled scenarios, is expected to have the following effects on the performance of \pbilcma{} with fixed $\lambda_s$. A higher $\lambda_s$ is expected to require more $f$-calls. However, a low $\lambda_s$ such that $\lambda_s \ll \abs{S_\mathrm{support}(x^*)}$ will struggle to converge because approximating $F$ becomes difficult. In this study, we investigated the impact of $\lambda_s$.

We chose P1 and P4 for  analysis. P2 and P3 are the same in terms of $\lambda_s \geq \abs{S_\mathrm{x^*}}$ is one of the necessary conditions to obtain a successful convergence; hence, they are omitted. On P4, in contrast to P1--P3, it is possible for \pbilcma{} with fixed $\lambda_s$ to optimize the problems if $\lambda_s$ is set to more than $2$ or $3$.  

\begin{figure}[t]
  \centering
  \begin{subfigure}{0.5\hsize}%
    \centering%
    \includegraphics[width=\hsize]{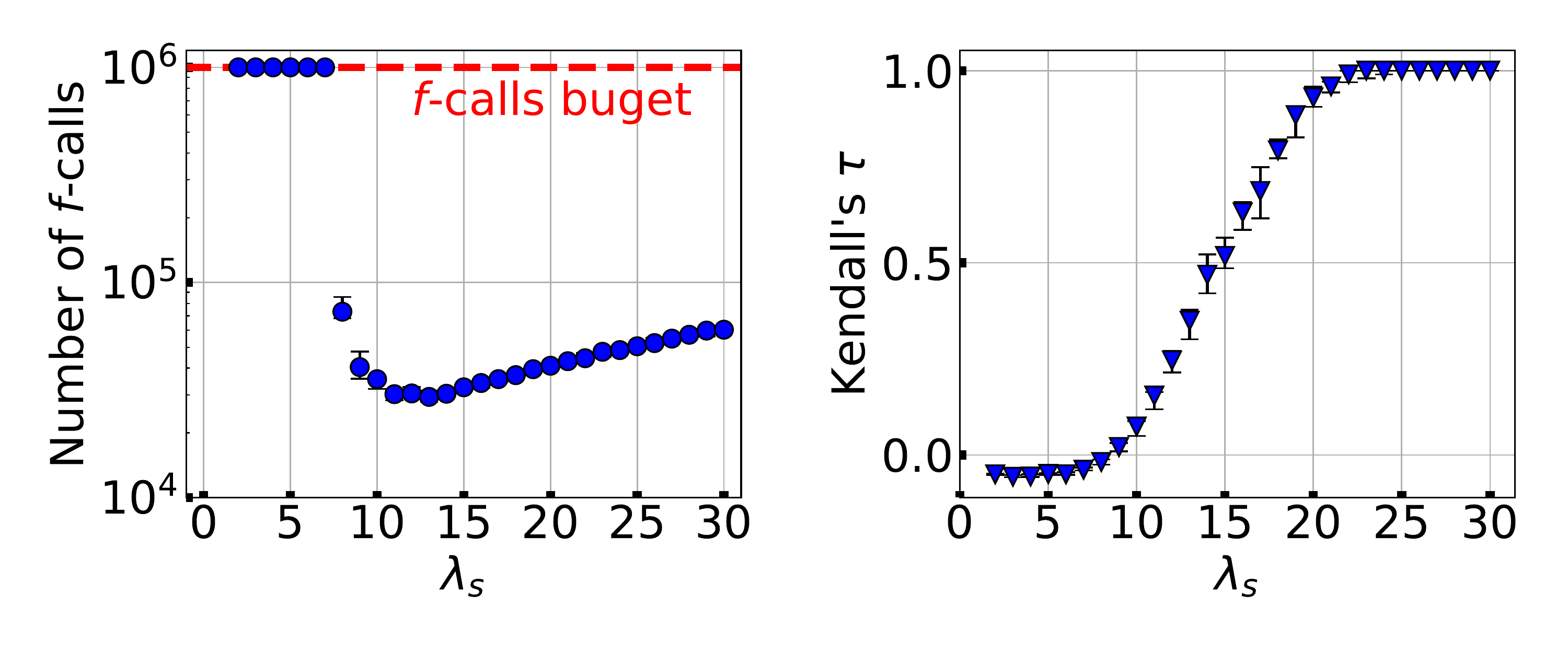}%
    \caption{P1 ($\abs{S_\mathrm{support}(x^*)} = 15$)}%
  \end{subfigure}%
  %\\  
  \begin{subfigure}{0.5\hsize}%
    \centering%    
    \includegraphics[width=\hsize]{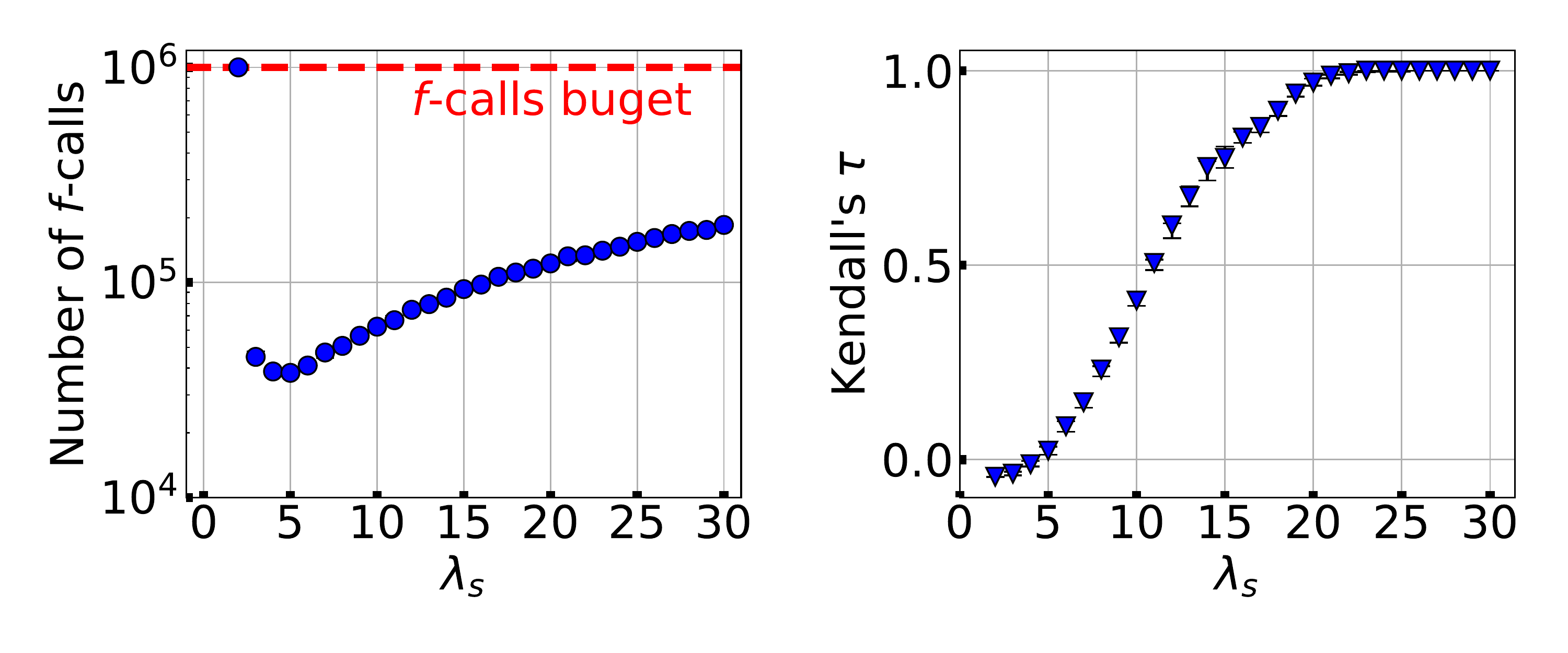}%
    \caption{P4 ($\abs{S_\mathrm{support}(x^*)} = 15$)}%
  \end{subfigure}%
  \caption{
Mean and standard deviation of the number of $f$-calls and Kendall's $\tau$ over 20 trials obtained from \pbilcma{} at fixed $\lambda_s$ for various $\lambda_s$. Kendall's $\tau$ values were averaged over the latter half of the search iterations. Cases with $f$-calls reaching $10^6$ were evaluated as optimization failures in this experiment.
  }
\label{fig:dis1-lambdas}
\end{figure}

In this analysis, we set $n=10$ and $m=30$ with $K = 15$ for $P1$ and $L = 15$ for $P4$. For the other experimental settings, we followed the setting described in \Cref{sec:common}. The results are shown in \Cref{fig:dis1-lambdas}. 

At P1, \pbilcma{} with fixed $\lambda_s$ converged at a lower number of $f$-calls by setting $\lambda_s \approx \abs{S_\mathrm{support}(x^*)}$. In addition, optimization was successful in some problems, even when $\lambda_s < \abs{S_\mathrm{support}(x^*)}$. 
This might be because the worst-case objective function $F$ was relatively well-approximated in the search neighborhood $H_\gamma^t$, as indicated by the relatively high values of Kendall's $\tau$. 
However, at the settings $\lambda_s \ll \abs{S_\mathrm{support}(x^*)}$, the optimization failed because $\lambda_s$ was too small to approximate $F$. 
At P4, $\lambda_s \geq 3$ was one of the necessary conditions to obtain a successful convergence, and $\lambda_s \approx 3$ had the highest performance, whereas $\lambda_s < 3$ led to failure. 

Consequently, the number of $f$-calls spent by \pbilcma{} with fixed $\lambda_s$ depended on the setting of $\lambda_s$. Among the cases obtaining successful convergence, the ratio of the number of $f$-calls was approximately $7$ at most in this experiment. However, if $\lambda_s$ was too small, the optimization was considered to have failed. 

\bibliography{./thebibliography.bib}

\end{document}